\documentclass{article} \usepackage{iclr2027_conference,times}

\usepackage{amsmath,amsfonts,bm}
\usepackage{xspace}

\makeatletter
\DeclareRobustCommand\onedot{\futurelet\@let@token\@onedot}
\def\@onedot{\ifx\@let@token.\else.\null\fi\xspace}

\makeatother

\def\eqref#1{equation~\ref{#1}}

\def\1{\bm{1}}

\def\vone{{\bm{1}}}

\def\vc{{\bm{c}}}
\def\vd{{\bm{d}}}

\def\vh{{\bm{h}}}

\def\vm{{\bm{m}}}
\def\vn{{\bm{n}}}

\def\vp{{\bm{p}}}
\def\vq{{\bm{q}}}

\def\vt{{\bm{t}}}

\def\vx{{\bm{x}}}
\def\vy{{\bm{y}}}
\def\vz{{\bm{z}}}

\def\evc{{c}}

\def\evm{{m}}

\def\evw{{w}}

\def\mF{{\bm{F}}}

\def\mR{{\bm{R}}}

\def\mX{{\bm{X}}}

\DeclareMathAlphabet{\mathsfit}{\encodingdefault}{\sfdefault}{m}{sl}
\SetMathAlphabet{\mathsfit}{bold}{\encodingdefault}{\sfdefault}{bx}{n}

\def\gD{{\mathcal{D}}}
\def\gE{{\mathcal{E}}}

\def\gG{{\mathcal{G}}}

\def\gI{{\mathcal{I}}}

\def\gK{{\mathcal{K}}}

\def\gP{{\mathcal{P}}}
\def\gQ{{\mathcal{Q}}}

\def\gV{{\mathcal{V}}}

\newcommand{\Ls}{\mathcal{L}}
\newcommand{\R}{\mathbb{R}}

\newcommand{\sigmoid}{\sigma}

\DeclareMathOperator*{\argmin}{arg\,min}

\DeclareMathOperator{\sign}{sign}

\usepackage{hyperref}
\usepackage{url}

\usepackage{algorithm}
\usepackage{algpseudocode}
\usepackage{lipsum}
\usepackage{booktabs}
\usepackage{multirow}
\usepackage[dvipsnames]{xcolor}
\usepackage{graphicx}
\usepackage{tikz}
\usepackage{caption}
\usepackage{subcaption}
\usepackage{booktabs}
\usepackage{tabularx}
\usepackage{siunitx}
\usepackage{pdflscape}
\usepackage{array}
\usepackage{amssymb}
\usepackage{pifont}
\usepackage{hyperref}
\usepackage{float}
\usepackage[table]{xcolor}
\usepackage{microtype}
\usepackage{placeins}
\usepackage{arydshln}
\usepackage{csquotes}

\hypersetup{
    colorlinks=true,
    urlcolor=magenta,
    linkcolor=RoyalBlue,
    citecolor=RoyalBlue
}

\usepackage[capitalise,noabbrev]{cleveref}
\crefname{section}{Section}{Sections}
\Crefname{section}{Section}{Sections}
\crefname{table}{Table}{Tables}
\Crefname{table}{Table}{Tables}
\crefname{appendix}{Appendix}{Appendices}
\Crefname{appendix}{Appendix}{Appendices}

\newcommand{\cmark}{\ding{51}}\newcommand{\xmark}{\ding{55}}

\definecolor{rankfirst}{HTML}{BFD8EB}
\definecolor{ranksecond}{HTML}{DAE9F4}
\definecolor{rankthird}{HTML}{EFF6FB}

\newcommand{\rankcell}[2]{\cellcolor{#1}#2}
\newcommand{\rankunderline}[1]{\underline{#1}}
\NewExpandableDocumentCommand{\rankfirst}{O{\rankcell} m}{#1{rankfirst}{\bfseries #2}}
\NewExpandableDocumentCommand{\ranksecond}{O{\rankcell} m}{#1{ranksecond}{\rankunderline{#2}}}
\NewExpandableDocumentCommand{\rankthird}{O{\rankcell} m}{#1{rankthird}{#2}}

\AddToHook{cmd/siunitx_cell_begin:w/before}{\renewcommand{\rankunderline}[1]{\underline{\tablenum[table-fixed-width=false]{#1}}}}

\newcommand{\litemidrule}{\arrayrulecolor{black!33}\midrule[\cmidrulewidth]\arrayrulecolor{black}}

\algrenewcommand{\algorithmiccomment}[1]{\hfill{\color{black!60}$\triangleright$ \textit{#1}}}

\title{\centering ARROW: Arbitrary Reconstruction\\
and Tracking of 4D Observations in the Wild}

\author{\parbox[t]{\dimexpr\textwidth-2\tabcolsep\relax}{\centering\bfseries
Ilya Fradlin \qquad Christian Schmidt \qquad Jens Piekenbrinck \\[4pt]
Karim Knaebel \qquad Gonzalo Martin Garcia \qquad Bastian Leibe \\[6pt]
\normalfont RWTH Aachen University \\[4pt]
{\small\url{https://vision.rwth-aachen.de/arrow}}
}
}

\newcommand{\method}{ARROW}
\colorlet{comparisonrule}{black!33}
\newlength{\comparisonrulewidth}
\newlength{\comparisonleftgap}
\newlength{\comparisonrightgap}
\newlength{\comparisonsep}
\newcommand{\comparisonstyle}{\setlength{\tabcolsep}{0pt}\renewcommand{\arraystretch}{0}\setlength{\comparisonrulewidth}{\cmidrulewidth}\setlength{\comparisonleftgap}{0.019\linewidth}\setlength{\comparisonrightgap}{0.008\linewidth}\setlength{\comparisonsep}{\dimexpr\comparisonleftgap+\comparisonrulewidth+\comparisonrightgap\relax}}
\newcolumntype{I}{c@{\hspace{\comparisonleftgap}{\color{comparisonrule}\vrule width\comparisonrulewidth}\hspace{\comparisonrightgap}}}
\newcommand{\comparisoninputheading}{\multicolumn{1}{@{}c@{\hspace{\comparisonsep}}}{\small Input}}
\newcommand{\comparisonours}{\small\textbf{\method{} (ours)}}
\newcommand{\comparisonrowsep}{\\\rule{0pt}{6pt}\\}

\iclrfinalcopy \begin{document}

\maketitle

\begin{figure}[H]
    \centering
    \includegraphics[width=\linewidth]{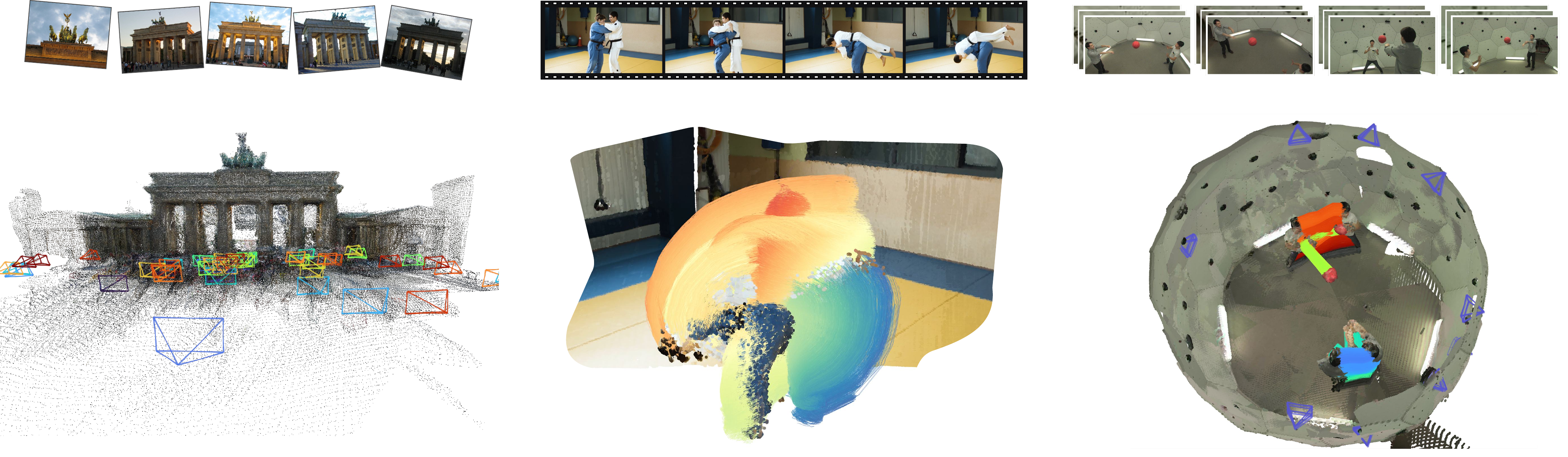}
    \par\vspace{2pt}
    \noindent\makebox[\linewidth]{\footnotesize \makebox[0.315\linewidth]{Multi-view 3D Reconstruction}\hfill
        \makebox[0.315\linewidth]{3D Tracking}\hfill
        \makebox[0.315\linewidth]{Multi-view 3D Tracking}}
    \caption{
        \textbf{\method{}} reconstructs 4D scenes over arbitrary sets of input images.
        It sets a new state of the art in 3D point tracking \& multi-view tracking and performs on par in 3D reconstruction.
    }
    \label{fig:teaser}
\end{figure}

\begin{abstract}

    Dynamic scenes may be captured by a moving camera, multiple video streams, or images taken at different times.
    These observations reveal complementary aspects of scene geometry and motion, yet bringing them together requires
    establishing correspondence across viewpoints, capture times, and visibility changes.
    We introduce \method{}, a feed-forward model that unifies 3D reconstruction and 3D point tracking from arbitrary image sets.
    At its core is a novel order-invariant querying approach, which allows the association of queries with observations across arbitrary inputs.
    We show that exposing the model to more diverse sets of inputs during training results in improved task performance.
    Moreover, the resulting model is capable of generalization to a wider range of tasks including multi-view tracking.
    Trained with this strategy, \method{} establishes a new state of the art in 3D tracking on WorldTrack and TAPVid-3D  
    and outperforms dedicated multi-view trackers on an adapted RGB-only MVTracker benchmark, 
    while remaining competitive across 3D reconstruction tasks.
    Code and weights are publicly available.

\end{abstract}

\section{Introduction}
\label{sec:introduction}
Reconstruction recovers a scene's 3D structure, while tracking maintains the identity of physical points and estimates their positions across observations.
Multi-view reconstruction has long considered images captured from diverse viewpoints, including unordered image collections~\citep{schoenberger2016colmap,schoenberger2016mvs,Wang_2024_DUSt3R}.
Point tracking, in contrast, has primarily focused on temporal correspondence within video~\citep{doersch2023tapir,karaev2024cotracker3,xiao2024spatialtracker}.
These are closely connected problems: geometry constrains correspondence, and correspondence relates structure across views and time, yet they have long been treated separately.
Recent models such as St4RTrack and D4RT demonstrate the power of unifying geometry and tracking within a single model~\citep{feng2025st4rtrack, zhang2026d4rt}.
D4RT, in particular, encodes a video once and answers independent queries specifying a source pixel, target time, and reference camera, allowing reconstruction and tracking to share a single representation space.
However, this unification retains the temporal scope of conventional tracking.
Training primarily on temporally local video introduces an inductive bias toward motion and appearance continuity. 
This in turn narrows the correspondence problems and supervision encountered during learning and potentially limits generalization.
Yet scene points can also be matched across unordered viewpoints without temporal continuity~\citep{leroy2024mast3r}.

To lift this restriction, we consider \emph{arbitrary observations} of the same scene, differing in viewpoint, capture time, or both, without requiring temporal continuity (\cref{fig:teaser}).
We exploit this broader observation space to diversify tracking supervision beyond continuous motion.
Observations spanning different cameras and distant capture times expose a wider range of appearances and can pose more challenging correspondence problems.
Motivated by positive transfer between related tasks~\citep{caruana1997multitask}, 
we aim to support these broader configurations through a shared reconstruction and tracking interface.

To this end, we take inspiration from D4RT's query formulation:
each query selects a pixel in a source observation (src) and requests its 3D position at a target observation's capture time (tgt),
expressed in a reference observation's coordinate system (ref).

\begin{figure}[t]
    \centering
\subcaptionbox{Order-dependent querying\label{fig:concept-order-dependent}}{\includegraphics[height=0.21\linewidth]{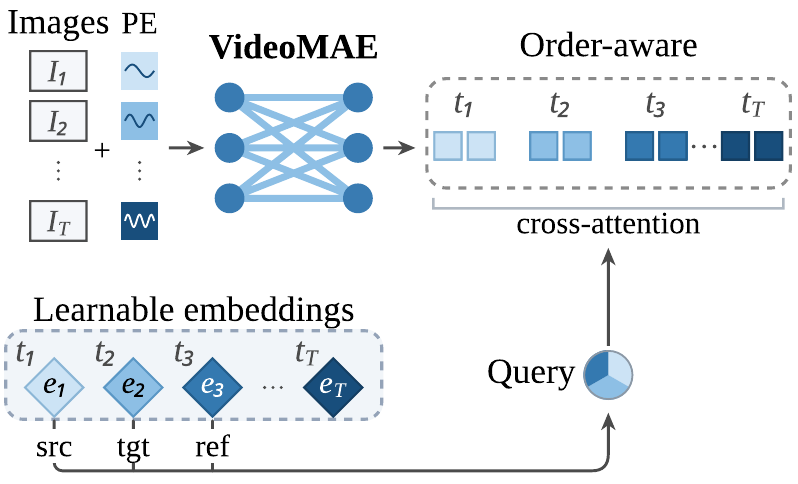}}
    \hfill
    \subcaptionbox{Feature-query mismatch\label{fig:concept-mismatch}}{\includegraphics[height=0.21\linewidth]{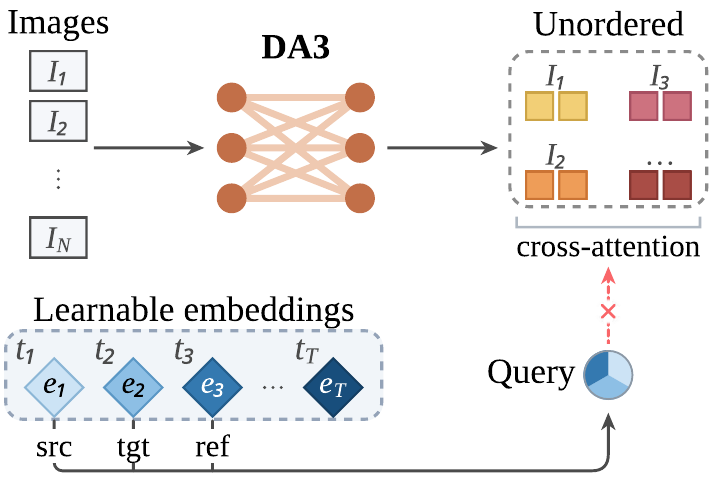}}
    \hfill
    \subcaptionbox{Order-invariant querying (ours)\label{fig:concept-order-invariant}}{\includegraphics[height=0.21\linewidth]{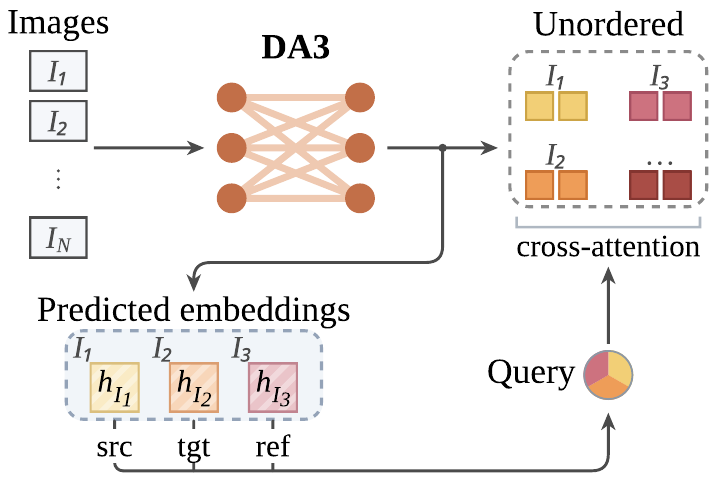}}
    \caption{
        \textbf{From indexed to input-dependent correspondence.}
        (a) Order-aware global context enables position-based associations.
(b) Order-invariant global context exposes no fixed order to exploit.
(c) Content-derived identity embeddings enable order-invariant associations.}
\label{fig:concept}
\end{figure}

However, extending this query interface to arbitrary observations reveals an architectural limitation.
Decoding from a shared representation requires associating queries with the intended observations. D4RT enables this association by explicitly imposing order on the global context via positional encodings (\cref{fig:concept-order-dependent}). Such order is natural for monocular video but does not apply to arbitrary observations: multi-view streams may share capture times, and static image collections have no inherent order.
Thus, while this query formulation trivially extends to general image collections, D4RT's architecture imposes unwarranted order.

We seek to retain this query flexibility without imposing order.
Geometry transformers such as DUSt3R, VGGT, and Depth Anything~3 (DA3) recover geometry from unposed image sets, learning relationships across diverse
viewpoints~\citep{Wang_2024_DUSt3R,wang2025vggt,lin2026depthanything3}. Such a prior is well-suited to relating observations that are not temporally adjacent.
Replacing the encoder with an order-free image-set representation, however, removes the sequence-position cues that associate fixed timestep query embeddings with the intended observations (\cref{fig:concept-mismatch}).
Adding positional encoding (PE) could reestablish the required association, but maintains the order dependency.
Instead, we propose to establish the correspondence between queries and the global context through content-derived identities.
We utilize DA3 to generate an order-free global context from the input images and predict per-frame embeddings.
Specifically, we repurpose DA3's per-image camera tokens as \emph{identity tokens} (ID~tokens) serving as input-derived embeddings unique to each frame.
These ID~tokens are used to indicate the queried source, target, and reference observations, replacing the role of the learned temporal embedding.
Crucially, these ID~tokens are order-invariant, as they are derived from the input images regardless of the frame position.
Unlike the learned temporal embeddings, the ID~tokens and the global context share the same embedding space and therefore can establish a meaningful order-invariant association (\cref{fig:concept-order-invariant}).

Based on this strategy, we introduce \method{}, a unified model for reconstruction and tracking across arbitrary unposed RGB observations.
Each query independently attends to the full encoded context, allowing all available observations to inform its prediction.
The same decoder supports the desired granularity an application needs, from sparse trajectories for interaction or control to dense geometry and motion for reconstruction.
Content-derived identities natively accommodate multiple camera streams and remove the need for a fixed bank of learned timestep identities.

We exploit this architectural change through \emph{non-sequential sampling} across camera views and beyond local temporal neighborhoods.
We jointly train on static image collections, dynamic videos, and multi-view sequences.
Thus, learning across arbitrary observations broadens supervision as well as the model's supported input settings.

Our cumulative ablations show the benefits of both the architectural change and the broader training strategy (\cref{sec:ablations}).
Specifically, content-derived addressing improves tracking and reconstruction, while exposing the model to a more challenging training distribution further improves performance.
The resulting model sets a new state of the art in 3D tracking on WorldTrack and TAPVid-3D, and achieves competitive performance in 3D reconstruction and camera pose estimation.
\method{} also outperforms dedicated multi-view trackers on the adapted RGB-only MVTracker benchmark.
These results show that broadening training beyond temporally local observations can improve temporal tracking itself, while supporting correspondence across a wider range of scene observations.

We summarize our contributions as follows:
\begin{itemize}
    \item We introduce \method{} (\cref{sec:method}), an open-source model trained on publicly available datasets, setting the state of the art on multiple benchmarks (\cref{sec:experiments}).
    \item At its core is our proposed order-invariant querying approach (\cref{fig:concept-order-invariant}), aligning the embedding space of the queries and the global context.
\item Extensive ablations showing the benefits of diversified tracking supervision (\cref{sec:ablations}).

\end{itemize}

\section{Related Work}
\label{sec:related}

\paragraph{3D geometry estimation.}
Classical structure-from-motion and multi-view stereo recover cameras and the geometry of a
fixed world through feature matching, triangulation, and optimization~\citep{schoenberger2016colmap,schoenberger2016mvs}.
Feed-forward models replaced this pipeline with direct point-map regression from unposed
images~\citep{Wang_2024_DUSt3R,leroy2024mast3r}, and subsequent geometry transformers extended
prediction from pairs to image sets~\citep{wang2025vggt,yang2025fast3r,wang2026pi3,keetha2026mapanything,lin2026depthanything3}.
Applying them to moving scenes requires distinguishing camera motion from changes in scene content,
addressed through dynamic-scene fine-tuning, joint depth and pose optimization,
or motion-aware masking~\citep{zhang2025monst3r,li2025megasam,hu2025vggt4d,zhou2026page4d}.
Recovering geometry at each observation, however, does not establish how the same physical
objects move between observations.

\paragraph{3D point tracking.}
A parallel line of work pursues this correspondence directly.
Image-plane trackers follow point identities through video~\citep{doersch2023tapir,karaev2024cotracker3}
but cannot resolve depth or separate camera from scene motion, motivating trackers that consume
depth or camera poses from an external estimator~\citep{xiao2024spatialtracker,ngo2024delta,ngo2025deltav2,zhang2025tapip3d}.

Feed-forward models remove this dependency by jointly predicting geometry and motion or correspondence, either from image
pairs~\citep{sucar2025dpm,feng2025st4rtrack,qian2026flow4r} or over a video~\citep{xiao2025spatialtrackerv2,lu2026track4world}.
Other works represent scene motion through compact parameterizations.
Shape of Motion~\citep{wang2025shapeofmotion} optimizes shared motion bases per scene,
while Trace Anything~\citep{liu2025traceanything} and SM4RT~\citep{lin2026sm4rt}
directly predict per-pixel B-spline trajectories and shared motion bases, respectively.
These representations support temporal interpolation but model motion across a video timeline,
irrespective of the requested point states.

\paragraph{Multi-view tracking.}
The trackers above follow points through a single camera stream. Multi-view trackers
instead combine simultaneous views to reduce depth ambiguity and recover points occluded
in individual views, but assume supplied camera calibration and synchronized
videos~\citep{rajic2025mvtracker,koo2026mvtap,galoaa2026lapa}. Without calibrated capture,
they therefore rely on an external reconstructor for cameras and, in MVTracker's case,
depth (\cref{tab:mvtrack-paper-ready}).
Recently OmniX~\citep{jiang2026omnix} seeks to jointly recover geometry and dense 3D trajectories
from multiple videos and image--video combinations in a feed-forward pass, but it retains explicit
timestamp, video-index, and local-frame encodings and does not expose sparse point queries.
The recent TAPVid-MV benchmark reflects growing interest in this setting, yet finds that
dedicated multi-view trackers do not consistently outperform monocular methods,
highlighting substantial room to better exploit additional views~\citep{koppula2026tapvidmv}.
\method{} seamlessly handles dense and sparse multi-view tracking without external geometric inputs thanks to its content-derived query mechanism.

\paragraph{Conditional and query-based tracking.}
Sparse decoding offers finer control by predicting only the requested point
states from a shared scene representation. L4P~\citep{badki2026l4p} shares a video backbone across geometry
and tracking but retains task-specific output heads, whereas D4RT~\citep{zhang2026d4rt} employs cross-attention
to independently decode queries specified by a source pixel, a target time, and a reference camera.
Its input side, however, remains video-oriented: a video-pretrained encoder, training on
temporally sampled clips, and query identities drawn from learned timestep embeddings.

A growing body of recent and concurrent work adapts geometry transformers to 4D reconstruction and tracking,
reporting promising results~\citep{karhade2025any4d,sucar2026vdpm,luo20264rc,huang2026fastspatial,chen2026uniquery4r,jeon2026point4d,jiang2026omnix}.
We discuss these methods in detail in~\cref{sup:extended_related}.
In short, each of these approaches recovers part of D4RT's decoding flexibility, yet none combines
\emph{(i)} a single model accepting static image collections, dynamic videos, and multi-view sequences,
\emph{(ii)} sparse decoding with independently selectable source, target, and reference, and
\emph{(iii)} a flexible representation free of explicit temporal or camera indexing.
\method{} realizes all three with one shared encoder and a pointwise decoder.

\section{Method}
\begin{figure}[t]
    \centering
    \includegraphics[width=\linewidth]{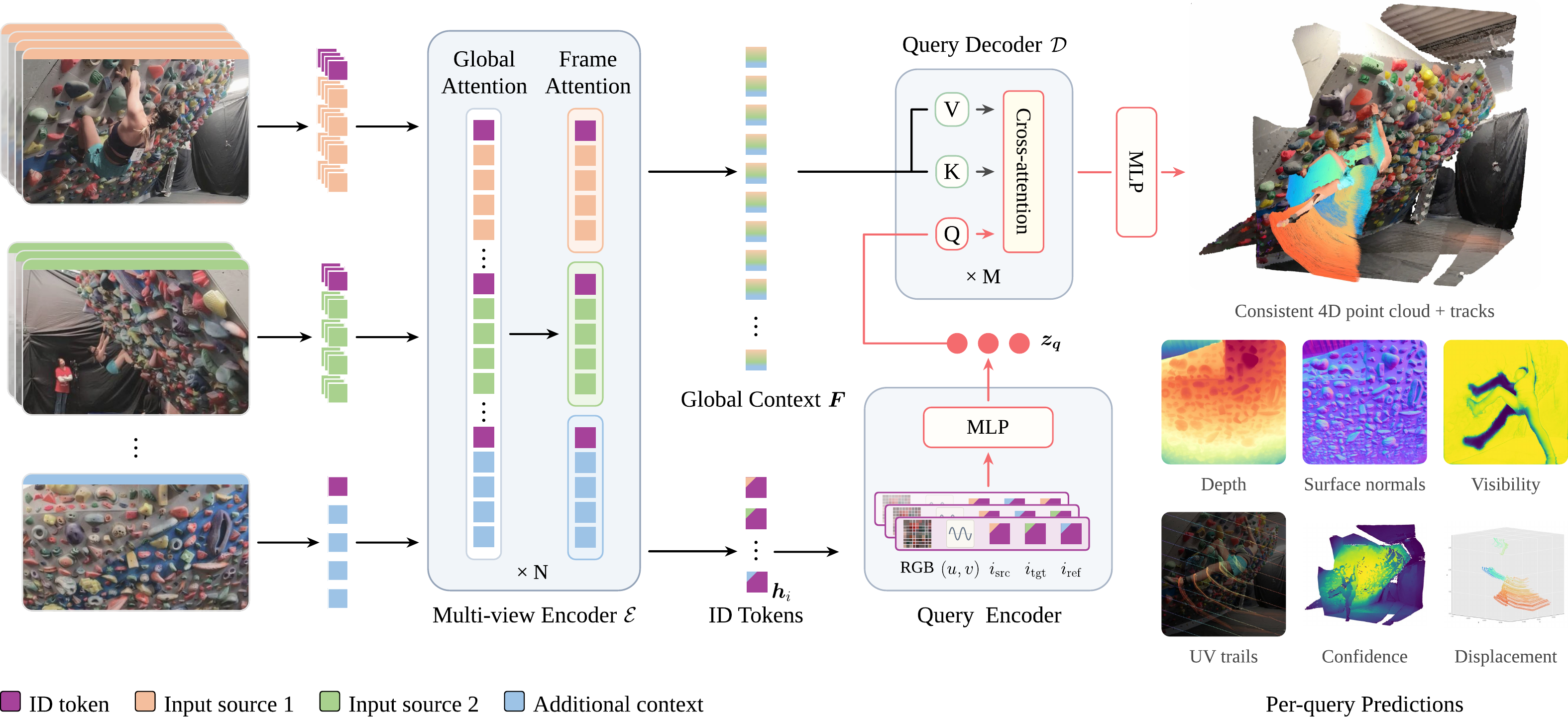}
    \caption{
        \textbf{\method{} architecture.}
The multi-view encoder $\gE$ tokenizes each
        frame independently and employs global attention to
        produce a shared global context $\mF$ together with an ID~token $\vh_i$ for each frame.
The query encoder combines a source-centered RGB crop, the Fourier-encoded $(u,v)$ coordinate, and the ID~tokens of the queried source, target, and reference observations into a latent query token $\vz_{\vq}$.
Each latent query token independently cross-attends to the global context, so a single
        encoding serves arbitrarily many queries and yields a consistent 4D representation of the scene.
    }
    \label{fig:architecture}
\end{figure}

\label{sec:method}

We consider an unordered set of RGB images $\gI=\{I_i\}_{i=1}^{S}$ constituting observations of a 4D scene.
Each observation $I_i\in\R^{H\times W\times 3}$ captures the scene from a particular viewpoint at a particular time.
Unlike a frame index in a video, $i$ only identifies an observation and implies neither an ordering nor temporal separation.
Distinct observations may share the same capture time.

\paragraph{Query formulation.} Inspired by the query interface of D4RT~\citep{zhang2026d4rt},
we define a query as a tuple of $(u,v)$ coordinates and indices specifying source, target, and reference observations:
\begin{equation}
    \vq=(u,v,i_{\mathrm{src}},i_{\mathrm{tgt}},i_{\mathrm{ref}}),
    \qquad (u,v)\in[0,1]^2,
    \qquad {i_\mathrm{src}},i_{\mathrm{tgt}},i_{\mathrm{ref}}\in \{1,\ldots,S\}.
    \label{eq:query}
\end{equation}
The pixel coordinates $(u,v)$ and the source observation~$I_{i_{\mathrm{src}}}$ describe a physical point at the capture time of~$I_{i_{\mathrm{src}}}$.
The desired output is the 3D position of that point at the time specified by the target observation~$I_{i_{\mathrm{tgt}}}$ and in the coordinate system specified by the reference observation~$I_{i_{\mathrm{ref}}}$.
This formulation enables the inference of a diverse set of outputs such as tracks, point clouds, and camera parameters \citep{zhang2026d4rt}.
Compared to D4RT, we explicitly support arbitrary input image sets.

\subsection{\method{} Architecture}
\Cref{fig:architecture} provides an overview of \method{}, which consists of a multi-view encoder, a query encoder, and a sparse query decoder.
The multi-view encoder receives an image set and outputs a global scene representation and a set of ID~tokens.
The query encoder translates a query's $(u,v)$ coordinates and observation indices into a latent query token, using the ID~tokens to encode observation identity.
The query decoder then predicts the per-query outputs via cross-attention between the query token and the global scene representation.
The network is trained end-to-end via per-query losses, without explicit supervision of observation identity.

\paragraph{Multi-view encoder.}
Our multi-view encoder $\gE$ is initialized from DA3~\citep{lin2026depthanything3}, a ViT~\citep{dosovitskiy2021vit}
that alternates frame-wise and global self-attention and is pretrained on
multi-view image sets to predict per-frame point maps and camera poses.
We repurpose the camera tokens as ID~tokens allowing the decoder to match the latent query token to the patch tokens of the source, target, and reference observations.
The ID~tokens are neither interpreted as camera estimates nor tied to the later pose-recovery procedure.

Formally, the multi-view encoder $\gE$ produces a global context $\mF$ and one
ID~token $\vh_i$ per observation:
\begin{equation}
    \bigl(\mF,\{\vh_i\}_{i=1}^{S}\bigr)=\gE(\gI),
    \qquad
    \mF\in\R^{N\times C},\quad \vh_i\in\R^{C},
    \label{eq:image_set_encoder}
\end{equation}
where $N$ is the total number of patch tokens across the image set and $C$ is
the feature dimension.
We add no temporal encoding to the encoder or context, since the input is treated as an
order-agnostic set.

\paragraph{Query encoder.}
The query encoder combines local appearance, $(u,v)$ coordinates, and the ID~tokens of the queried observations into a latent query token $\vz_{\vq}\in\R^C$:
\begin{equation}
    \vz_{\vq}=
    \phi_{\mathrm{rgb}}(I_{i_{\mathrm{src}}},u,v)
    +\phi_{\mathrm{uv}}(u,v)
    +\phi_{\mathrm{rel}}\!\left(
    [\vh_{i_{\mathrm{src}}};
        \vh_{i_{\mathrm{tgt}}};
        \vh_{i_{\mathrm{ref}}}]
    \right).
    \label{eq:query_composition}
\end{equation}
Here, $\phi_{\mathrm{rgb}}$ embeds a small source-centered RGB crop,
$\phi_{\mathrm{uv}}$ embeds the Fourier-encoded normalized $(u,v)$ coordinates, and
$\phi_{\mathrm{rel}}$ projects the concatenation of the ID~tokens to the feature dimension of the decoder.
All three $\phi$ are implemented as two-layer MLPs.
We normalize pixel centers as $u=(x+1/2)/W$ and $v=(y+1/2)/H$.
Empirically, this relative-position parameterization supports rectangular inputs with varying aspect ratios.

\paragraph{Sparse decoder.}
Each latent query token independently cross-attends to $\mF$, with no self-attention between
query tokens. The lightweight cross-attention decoder $\gD$ predicts:
\begin{equation}
    \hat{\vy}_{\vq}=\gD(\vz_{\vq},\mF)
    =\left(
    \hat{\vx}^{i_{\mathrm{ref}}}_{i_{\mathrm{tgt}}},
    \hat{c},
    (\hat{u}_{i_{\mathrm{tgt}}},\hat{v}_{i_{\mathrm{tgt}}}),
    \hat{o}_{i_{\mathrm{tgt}}},
    \hat{\vn}^{i_{\mathrm{ref}}}_{i_{\mathrm{tgt}}},
    \hat{\vd}^{i_{\mathrm{ref}}}_{i_{\mathrm{src}}\rightarrow i_{\mathrm{tgt}}}
    \right).
    \label{eq:query_predictions}
\end{equation}
A superscript denotes the reference frame in which a 3D quantity is expressed and a
subscript the target time; indices are suppressed in the text for brevity.
The query decoder outputs a 3D position $\hat{\vx}\in\R^3$, a confidence $\hat{c}\in\R_{>1}$, image coordinates $(\hat u,\hat v)\in\R^2$, a visibility logit $\hat o\in\R$, a unit surface normal
$\hat{\vn}\in\R^3$,
and the source-to-target displacement $\hat{\vd}\in\R^3$.
All quantities are predicted at the time of capture of the target observation.
The 3D position, surface normals, and displacement are predicted in the coordinate system of the reference observation.
The point coordinates and displacement are normalized as described in~\cref{sup:implementation}.
UV predictions are unconstrained in normalized image coordinates, and the
visibility probability is $\sigmoid(\hat o)$.
We recover camera poses from the predicted 3D points, as described in~\cref{sup:camera_recovery}.
Since queries are independent from each other, the encoded context can be reused for sparse queries and dense decoding.

\subsection{Training}

\paragraph{Image and query sampling.}
We use three complementary data regimes.
For \emph{static scenes}, known geometry provides trivial correspondence: the physical 3D point remains fixed while its appearance and camera coordinates change.
For \emph{dynamic scenes with tracks}, annotated tracks provide cross-time correspondence and motion supervision, while static background provides a crucial anchor for the scene.
We sample both static and dynamic regions and choose observations from a candidate temporal window up to ten times larger than the requested set, with multi-view datasets potentially also contributing different cameras from the same scene.
Finally, for \emph{dynamic scenes without tracks}, such as nuScenes~\citep{caesar2020nuscenes}, we restrict queries to the same target and source time step ($i_{\mathrm{tgt}}=i_{\mathrm{src}}$), avoiding incorrect motion supervision while retaining geometric supervision.
Dataset-level image policies and point-level edge, boundary, and motion sampling are detailed in \cref{sup:sampling}.

\paragraph{Losses.}
We optimize a multi-task objective, applying each term only where its target is valid:
\begin{equation}
    \Ls=
    \lambda_{\mathrm{3D}}\Ls_{\mathrm{3D}}
    +\lambda_{\mathrm{conf}}\Ls_{\mathrm{conf}}
    +\lambda_{\mathrm{uv}}\Ls_{\mathrm{uv}}
    +\lambda_{\mathrm{vis}}\Ls_{\mathrm{vis}}
    +\lambda_{\mathrm{normal}}\Ls_{\mathrm{normal}}
    +\lambda_{\mathrm{disp}}\Ls_{\mathrm{disp}}.
    \label{eq:training_objective}
\end{equation}
The primary term $\Ls_{\mathrm{3D}}$ is a scale-invariant, confidence-weighted
$\ell_1$ loss on the predicted 3D positions, and
$\Ls_{\mathrm{conf}}=-\operatorname{mean}\log\hat c$ regularizes the confidence.
The auxiliary terms supervise the remaining outputs: $\Ls_{\mathrm{uv}}$ and
$\Ls_{\mathrm{disp}}$ are $\ell_1$ losses on the target image coordinates and
displacement, $\Ls_{\mathrm{vis}}$ is a binary cross-entropy on visibility, and
$\Ls_{\mathrm{normal}}$ is a cosine distance on surface normals.
Every source, target, and reference combination is thus supervised by the
same objective, without a dedicated tracking loss or a shared world frame.
The coefficients $\lambda$ balance the loss terms; their values, exact formulations, and normalization are detailed in \cref{sup:implementation}.

\begin{figure}[t]
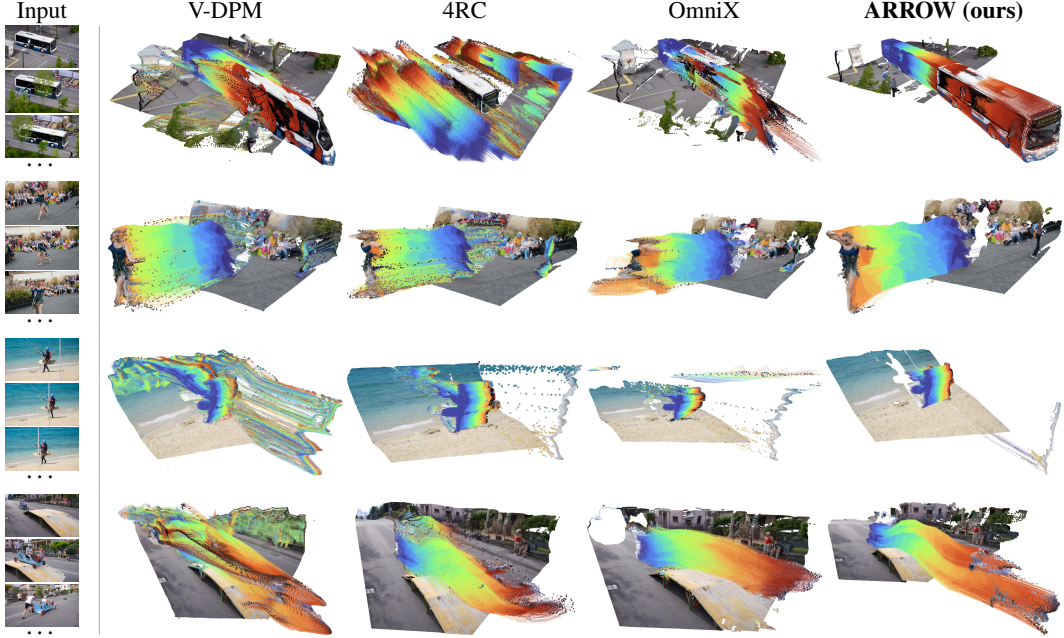

    \centering
    \begingroup
    \comparisonstyle
    \newlength{\qualrgbw}\setlength{\qualrgbw}{0.07\linewidth}
    \newlength{\qualrgbh}\setlength{\qualrgbh}{0.04\linewidth}
    \newlength{\qualh}\setlength{\qualh}{\dimexpr3\qualrgbh+2pt\relax}
    \newlength{\qualw}\setlength{\qualw}{\dimexpr(\linewidth-\qualrgbw-6pt-\comparisonsep)/4\relax}
    \newcommand{\qualpanel}[2]{\vbox to \qualh{\vfil
            \hbox to \qualw{\hfil\includegraphics[width=\qualw,height=\qualh,keepaspectratio]{fig/qual_comparison/#1/#2}\hfil}\vfil}}
    \newcommand{\qualrgb}[2]{\includegraphics[width=\qualrgbw,height=\qualrgbh]{fig/qual_comparison/#1/rgb#2}}
    \newcommand{\comparisonrow}[1]{\vbox{\offinterlineskip
            \hbox{\qualrgb{#1}{1}}\kern1pt
            \hbox{\qualrgb{#1}{2}}\kern1pt
            \hbox{\qualrgb{#1}{3}}\kern2pt
            \hbox to \qualrgbw{\hfil\ldots\hfil}}&
        \qualpanel{#1}{vdpm} &
        \qualpanel{#1}{4rc} &
        \qualpanel{#1}{omnix} &
        \qualpanel{#1}{arrow}}
    \begin{tabular}{@{}Ic@{\hspace{2pt}}c@{\hspace{2pt}}c@{\hspace{2pt}}c@{}}
        \comparisoninputheading & \small V-DPM & \small 4RC & \small OmniX & \comparisonours \\[3pt]
        \comparisonrow{bus}\comparisonrowsep
        \comparisonrow{dance_twirl}\comparisonrowsep
\comparisonrow{kite_walk}\comparisonrowsep
        \comparisonrow{soapbox}\end{tabular}\par
    \endgroup
    \caption{
        \textbf{4D reconstruction and tracking.}
        V-DPM, 4RC, OmniX, and \method{} on four DAVIS~\citep{perazzi2016davis} sequences.
        Input frames are on the left; each panel shows its 3D reconstruction with trajectory overlays.
        \method{} shows clean reconstructions and consistent long-term tracks.
    }
    \label{fig:qual_comparison}
\end{figure}

\section{Experiments}
\label{sec:experiments}
We train for $200\mathrm{k}$ steps using DA3-Giant, with ablations using $100\mathrm{k}$ steps, DA3-Large, and six decoder layers (\cref{sup:implementation}).
\Cref{tab:training-data,tab:sampling-rules} detail the training mixture and dataset-specific sampling policies.
We evaluate single- and multi-view 3D tracking, camera pose, reconstruction, and video depth.
\Cref{fig:qual_comparison} shows qualitative 4D reconstruction and tracking results, while \cref{fig:feature_matching_uv} shows point correspondences across challenging image pairs.

\paragraph{Monocular 4D reconstruction and tracking.}
\label{sec:world_tracking}
WorldTrack~\citep{feng2025st4rtrack} benchmarks 3D point tracking over
$64$-frame sequences from Aria Digital Twin~(ADT)~\citep{pan2023adt}, Dynamic Replica~\citep{karaev2023dynamicstereo},
PointOdyssey~\citep{zheng2023pointodyssey}, and Panoptic Studio~\citep{joo2015pstudio}, with queries initialized
in the first frame. We express trajectories in the first camera's coordinate
system and apply global median scale alignment per scene.
\Cref{sup:worldtrack} details the evaluation thresholds and
dynamic-track selection. \method{} achieves the highest APD and lowest EPE across all four datasets,
for both overall and dynamic-only tracking
(\cref{tab:worldtrack-paper-ready}).

\begin{table}[t]
    \centering
    \scriptsize
    \setlength{\tabcolsep}{2pt}
    \renewcommand{\arraystretch}{1.2}
    \sisetup{mode=text,reset-text-series=false}
\caption{
        \textbf{WorldTrack 3D point tracking.}
        We report the average percentage of
        points within distance thresholds (APD, \%) and mean 3D end-point
        error (EPE, meters). All and Dyn.\ denote overall and dynamic-only scores.
        Results use global median scale alignment.
        \method{} sets a new state of the art.
    }
    \label{tab:worldtrack-paper-ready}
    \begin{tabularx}{\linewidth}{@{}>{\raggedright\arraybackslash}X*{4}{*{2}{S[table-format=2.2]}*{2}{S[table-format=1.4]}}@{}}
        \toprule
        & \multicolumn{4}{c}{\textbf{Aria Digital Twin}} & \multicolumn{4}{c}{\textbf{Dynamic Replica}} & \multicolumn{4}{c}{\textbf{PointOdyssey}} & \multicolumn{4}{c}{\textbf{Panoptic Studio}} \\
        \cmidrule(lr){2-5}
        \cmidrule(lr){6-9}
        \cmidrule(lr){10-13}
        \cmidrule(lr){14-17}
        \textbf{Method} & \multicolumn{2}{c}{APD \ensuremath{\uparrow}} & \multicolumn{2}{c}{EPE \ensuremath{\downarrow}} & \multicolumn{2}{c}{APD \ensuremath{\uparrow}} & \multicolumn{2}{c}{EPE \ensuremath{\downarrow}} & \multicolumn{2}{c}{APD \ensuremath{\uparrow}} & \multicolumn{2}{c}{EPE \ensuremath{\downarrow}} & \multicolumn{2}{c}{APD \ensuremath{\uparrow}} & \multicolumn{2}{c}{EPE \ensuremath{\downarrow}} \\
        \cmidrule(lr){2-3}
        \cmidrule(lr){4-5}
        \cmidrule(lr){6-7}
        \cmidrule(lr){8-9}
        \cmidrule(lr){10-11}
        \cmidrule(lr){12-13}
        \cmidrule(lr){14-15}
        \cmidrule(lr){16-17}
        & {All} & {Dyn.} & {All} & {Dyn.} & {All} & {Dyn.} & {All} & {Dyn.} & {All} & {Dyn.} & {All} & {Dyn.} & {All} & {Dyn.} & {All} & {Dyn.} \\
        \midrule
        SpatialTrackerV2 & \ranksecond{91.75} & 83.18 & \ranksecond{0.0964} & 0.0832 & 69.90 & 67.18 & 0.2986 & 0.3348 & 67.36 & 68.03 & 0.4299 & 0.4789 & 73.42 & 73.42 & 0.2397 & 0.2397 \\
        St4RTrack & 72.89 & 77.98 & 0.2958 & 0.1087 & 74.29 & 76.85 & 0.2600 & 0.1973 & 67.45 & 69.89 & 0.3144 & 0.2722 & 70.79 & 70.79 & 0.2480 & 0.2480 \\
        Trace Anything & 75.71 & 76.07 & 0.2517 & 0.1875 & 60.64 & 60.58 & 0.5751 & 0.4730 & 39.51 & 43.70 & 1.0546 & 0.8906 & 71.17 & 71.17 & 0.2788 & 0.2788 \\
        V-DPM & 85.35 & 73.27 & 0.1763 & 0.1455 & 77.28 & 80.13 & 0.2306 & 0.1904 & 80.47 & 82.59 & \rankthird{0.1886} & \rankthird{0.1672} & 77.11 & 77.11 & 0.1881 & 0.1881 \\
        Any4D & 57.88 & 69.59 & 0.4149 & 0.1420 & 79.45 & 69.74 & 0.2140 & 0.2976 & 66.87 & 67.86 & 0.3498 & 0.3574 & 63.98 & 63.98 & 0.2978 & 0.2978 \\
        4RC & 83.19 & 71.83 & 0.1864 & 0.1427 & \ranksecond{84.41} & 76.27 & \ranksecond{0.1750} & 0.2515 & \rankthird{83.15} & \rankthird{83.83} & 0.2175 & 0.2351 & 68.19 & 68.19 & 0.2657 & 0.2657 \\
        Track4World & 79.84 & \rankthird{88.48} & 0.2173 & \rankthird{0.0571} & 71.49 & 70.24 & 0.2709 & 0.2740 & 66.78 & 68.19 & 0.3505 & 0.3151 & 73.16 & 73.16 & 0.2265 & 0.2265 \\
        OmniX & \rankthird{91.05} & 86.58 & \rankthird{0.1029} & 0.1108 & 79.91 & 72.64 & 0.2138 & 0.3071 & 78.68 & 78.47 & 0.2545 & 0.2779 & 77.23 & 77.23 & 0.1994 & 0.1994 \\
        OpenD4RT\textsuperscript{*} & 71.92 & 72.68 & 0.2814 & 0.1627 & 76.59 & \ranksecond{82.18} & 0.2438 & \rankthird{0.1754} & 67.61 & 70.09 & 0.3220 & 0.2969 & \rankthird{78.98} & \rankthird{78.98} & 0.1794 & 0.1794 \\
        SM4RT\textsuperscript{\textdagger} & 86.86 & 80.55 & 0.1515 & 0.0957 & \rankthird{82.77} & 73.70 & \rankthird{0.1848} & 0.2654 & 81.05 & 80.69 & 0.2345 & 0.2614 & 78.76 & 78.76 & \rankthird{0.1788} & \rankthird{0.1788} \\
        Point4D\textsuperscript{\textdagger} & 62.73 & 77.38 & 0.4462 & 0.1578 & 77.77 & 70.72 & 0.2356 & 0.3004 & 69.84 & 72.55 & 0.3604 & 0.3580 & 72.92 & 72.92 & 0.2354 & 0.2354 \\
        UniQuery4R\textsuperscript{\textdagger} & 87.18 & \ranksecond{90.11} & 0.1387 & \ranksecond{0.0501} & 79.25 & \rankthird{81.95} & 0.2146 & \ranksecond{0.1633} & \ranksecond{83.84} & \ranksecond{85.47} & \ranksecond{0.1607} & \ranksecond{0.1510} & \ranksecond{84.76} & \ranksecond{84.76} & \ranksecond{0.1339} & \ranksecond{0.1339} \\
        \midrule
        \textbf{\method{}} & \rankfirst{93.55} & \rankfirst{90.63} & \rankfirst{0.0818} & \rankfirst{0.0481} & \rankfirst{88.00} & \rankfirst{91.47} & \rankfirst{0.1426} & \rankfirst{0.0896} & \rankfirst{85.26} & \rankfirst{86.72} & \rankfirst{0.1532} & \rankfirst{0.1417} & \rankfirst{85.76} & \rankfirst{85.76} & \rankfirst{0.1332} & \rankfirst{0.1332} \\
        \bottomrule
    \end{tabularx}
    \par
    \begin{minipage}{\linewidth}
        \textsuperscript{*} OpenD4RT is an unofficial implementation of D4RT~\citep{zhang2026d4rt}: \url{https://github.com/Lijiaxin0111/Open-d4rt}.
        \par
        \textsuperscript{\textdagger} Concurrent work first released on arXiv between July and September 2026.
    \end{minipage}
\end{table}

\begin{table}[t]
    \centering
    \tiny
    \setlength{\tabcolsep}{2pt}
    \renewcommand{\arraystretch}{1.2}
    \caption{\textbf{TAPVid-3D evaluation in camera and world coordinates.} Without
        ground-truth intrinsics. L1 is measured after normalizing predictions
        and ground truth by their median distances to the reference camera.
Baseline results are taken from D4RT~\citep{zhang2026d4rt}.
        UD2 denotes UniDepthV2~\citep{piccinelli2025unidepthv2}.
        \method{} leads in average AJ, APD, and normalized L1.
    }
\label{tab:tapvid3d-paper-ready}
    \begin{tabularx}{\linewidth}{@{}>{\raggedright\arraybackslash}X*{18}{S[table-format=1.3,mode=text,reset-text-series=false]}@{}}
        \toprule
                & \multicolumn{12}{c}{\textbf{Camera-coordinate 3D tracking}} & \multicolumn{6}{c}{\textbf{World-coordinate 3D tracking}}                                                                                                                                                                                                                                                                                                                                                                                                                                                                                                                                                                                                                                                                                                                                 \\
        \cmidrule(lr){2-13}
        \cmidrule(lr){14-19}
        \textbf{Method} & \multicolumn{3}{c}{\textbf{DriveTrack}}                     & \multicolumn{3}{c}{\textbf{Aria Digital Twin}}                          & \multicolumn{3}{c}{\textbf{Panoptic Studio}}      & \multicolumn{3}{c}{\textbf{Avg.}}         & \multicolumn{2}{c}{\textbf{DriveTrack}}   & \multicolumn{2}{c}{\textbf{ADT}}          & \multicolumn{2}{c}{\textbf{Avg.}}                                                                                                                                                                                                                                                                                                                                                                                                                                                                                                             \\
        \cmidrule(lr){2-4}
        \cmidrule(lr){5-7}
        \cmidrule(lr){8-10}
        \cmidrule(lr){11-13}
        \cmidrule(lr){14-15}
        \cmidrule(lr){16-17}
        \cmidrule(lr){18-19}
                                                   & {\mbox{AJ \ensuremath{\uparrow}}}                                    & {\mbox{APD \ensuremath{\uparrow}}}                               & {\mbox{OA \ensuremath{\uparrow}}}                  & {\mbox{AJ \ensuremath{\uparrow}}}                  & {\mbox{APD \ensuremath{\uparrow}}}               & {\mbox{OA \ensuremath{\uparrow}}}                  & {\mbox{AJ \ensuremath{\uparrow}}}                  & {\mbox{APD \ensuremath{\uparrow}}}               & {\mbox{OA \ensuremath{\uparrow}}}                  & {\mbox{AJ \ensuremath{\uparrow}}}                  & {\mbox{APD \ensuremath{\uparrow}}}               & {\mbox{OA \ensuremath{\uparrow}}}                  & {\mbox{APD \ensuremath{\uparrow}}}               & {\mbox{L1 \ensuremath{\downarrow}}}                & {\mbox{APD \ensuremath{\uparrow}}}               & {\mbox{L1 \ensuremath{\downarrow}}}                & {\mbox{APD \ensuremath{\uparrow}}}               & {\mbox{L1 \ensuremath{\downarrow}}}                \\
        \midrule
        St4RTrack               & {---} & {---} & {---} & {---} & {---} & {---} & {---} & {---} & {---} & {---} & {---} & {---} & 0.020                                     & 0.499                                     & 0.062                                     & 0.839                                     & 0.041                                     & 0.669                                     \\
        CoTracker3 \; + \textit{UD2} & 0.038                                                       & 0.062                                                     & \rankthird{0.856}             & 0.102                                     & 0.146                                     & \rankfirst{0.937}    & \rankthird{0.119}             & \rankthird{0.176}             & \rankthird{0.862}             & 0.086                                     & 0.128                                     & \ranksecond{0.885} & {---} & {---} & {---} & {---} & {---} & {---} \\
        CoTracker3 \; + \textit{VGGT}       & \rankthird{0.129}                               & \rankthird{0.189}                             & \rankthird{0.856}             & 0.132                                     & 0.190                                     & \rankfirst{0.937}    & 0.045                                     & 0.070                                     & \rankthird{0.862}             & 0.102                                     & 0.150                                     & \ranksecond{0.885} & \rankthird{0.205}             & 0.170                                     & 0.192                                     & 0.736                                     & 0.199                                     & 0.453                                     \\
        SpatialTrackerV2        & 0.064                                                       & 0.100                                                     & \ranksecond{0.865} & \ranksecond{0.260} & \ranksecond{0.342} & \ranksecond{0.936} & 0.097                                     & 0.152                                     & 0.805                                     & \rankthird{0.140}             & \rankthird{0.198}             & \rankthird{0.869}             & 0.101                                     & \rankthird{0.139}             & \ranksecond{0.336} & \rankthird{0.238}             & \rankthird{0.219}             & \rankthird{0.189}             \\
        D4RT                    & \rankfirst{0.257}                      & \ranksecond{0.345}                 & \rankfirst{0.875}    & \rankthird{0.240}             & \rankthird{0.319}             & \rankthird{0.926}             & \ranksecond{0.138} & \ranksecond{0.186} & \rankfirst{0.897}    & \ranksecond{0.212} & \ranksecond{0.283} & \rankfirst{0.899}    & \rankfirst{0.373}    & \rankfirst{0.020}    & \rankthird{0.319}             & \ranksecond{0.096} & \ranksecond{0.346} & \ranksecond{0.058} \\
        \midrule
        \textbf{\method{}}                         & \ranksecond{0.241}                   & \rankfirst{0.350}                    & 0.816                                     & \rankfirst{0.360}    & \rankfirst{0.473}    & 0.902                                     & \rankfirst{0.141}    & \rankfirst{0.201}    & \ranksecond{0.870} & \rankfirst{0.248}    & \rankfirst{0.341}    & 0.863                                     & \ranksecond{0.327} & \ranksecond{0.027} & \rankfirst{0.442}    & \rankfirst{0.054}    & \rankfirst{0.385}    & \rankfirst{0.041}    \\
        \bottomrule
    \end{tabularx}
\end{table}

TAPVid-3D~\citep{koppula2024tapvid3d} evaluates bidirectional tracking from arbitrary query frames over sequences of up to $300$ frames taken from DriveTrack~\citep{balasingam2024drivetrack}, ADT, and Panoptic Studio.
Long sequences and arbitrary source frames require windowed methods to propagate tracks bidirectionally and align them across windows.
\method{} instead queries every target against the full encoded sequence, choosing the target camera or first camera as reference (cf.\ \cref{sup:longtracking}).

Following D4RT~\citep{zhang2026d4rt}, we evaluate on the mini-val split without ground-truth intrinsics in camera coordinates and a shared world frame (the first input camera).
Both settings use the same per-sequence median-scale alignment formula as WorldTrack.
We report APD, average Jaccard~(AJ) and occlusion accuracy~(OA) in \cref{tab:tapvid3d-paper-ready}.
\method{} achieves the best average camera-space AJ and APD, and world-space APD and L1, among the compared methods.
It remains competitive in OA, slightly trailing the baselines.
We observe image-coordinate errors of a few pixels and hypothesize that training across varying aspect ratios might require longer optimization for precise UV localization.

\paragraph{Multi-view 4D reconstruction and tracking.}
\label{sec:multiview_tracking}
We adapt the MVTracker benchmark~\citep{rajic2025mvtracker} to an RGB-only input setting for 3D tracking on Multi-View Kubric~\citep{greff2022kubric}, DexYCB~\citep{chao2021dexycb}, and Panoptic Studio.
The original protocol supplies depth, camera calibration, and 3D queries; our adaptation uses RGB images from four static cameras over $24$ synchronized timesteps and 2D queries in visible source views.
This requires joint scene reconstruction and tracking throughout the clip, before and after each query's source time.
Trajectories are expressed in the first input camera's coordinate system and aligned by a single scale factor per scene.
\Cref{tab:mvtrack-paper-ready} reports APD and EPE on ground-truth-visible positions, alongside AJ and OA.

\method{} leads on 11 of the 12 dataset--metric combinations, with DexYCB EPE as the sole exception. 
MVTracker shows similarly elevated EPE with DA3 but not with VGGT or VGGT-$\Omega$~\citep{wang2026vggtomega}, suggesting a mismatch with DA3's pretraining distribution, and illustrates that the prior remains influential for tracking accuracy.
Nevertheless, \method{} achieves the highest DexYCB APD and AJ, which are less sensitive than EPE to the magnitude of large errors.

\begin{table}[t]
    \centering
    \tiny
    \setlength{\tabcolsep}{3pt}
    \renewcommand{\arraystretch}{1.2}
    \sisetup{mode=text,reset-text-series=false}
    \caption{
        \textbf{RGB-only multi-view 3D tracking.}
        We report APD and EPE on ground-truth-visible positions, as well as OA and AJ.
        $+$ denotes methods requiring an external reconstruction model.
    }
\label{tab:mvtrack-paper-ready}
    \begin{tabularx}{\linewidth}{@{}>{\raggedright\arraybackslash}X*{4}{S[table-format=2.1]S[table-format=1.4]*{2}{S[table-format=2.1]}}@{}}
        \toprule
        \multirow{2}{*}[-0.5\dimexpr\aboverulesep+\cmidrulewidth+\belowrulesep\relax]{\textbf{Method}} & \multicolumn{4}{c}{\textbf{Panoptic Studio}} & \multicolumn{4}{c}{\textbf{DexYCB}} & \multicolumn{4}{c}{\textbf{Multi-View Kubric}} & \multicolumn{4}{c}{\textbf{Avg.}} \\
        \cmidrule(lr){2-5}
        \cmidrule(lr){6-9}
        \cmidrule(lr){10-13}
        \cmidrule(lr){14-17}
         & {\mbox{APD \ensuremath{\uparrow}}} & {\mbox{EPE \ensuremath{\downarrow}}} & {\mbox{OA \ensuremath{\uparrow}}} & {\mbox{AJ \ensuremath{\uparrow}}} & {\mbox{APD \ensuremath{\uparrow}}} & {\mbox{EPE \ensuremath{\downarrow}}} & {\mbox{OA \ensuremath{\uparrow}}} & {\mbox{AJ \ensuremath{\uparrow}}} & {\mbox{APD \ensuremath{\uparrow}}} & {\mbox{EPE \ensuremath{\downarrow}}} & {\mbox{OA \ensuremath{\uparrow}}} & {\mbox{AJ \ensuremath{\uparrow}}} & {\mbox{APD \ensuremath{\uparrow}}} & {\mbox{EPE \ensuremath{\downarrow}}} & {\mbox{OA \ensuremath{\uparrow}}} & {\mbox{AJ \ensuremath{\uparrow}}} \\
        \midrule
        MVTracker \; + \textit{VGGT} & 54.6 & 0.2369 & 69.5 & 37.0 & 39.7 & \rankthird{0.1063} & 82.6 & 30.0 & 31.7 & 0.3070 & 72.4 & 21.2 & 42.0 & 0.2167 & 74.8 & 29.4 \\
        MVTracker \; + \textit{DA3} & \rankthird{59.9} & 0.2054 & 69.1 & \ranksecond{39.1} & 32.4 & 0.1732 & 80.8 & 23.0 & \ranksecond{43.6} & \ranksecond{0.1595} & \rankthird{76.4} & \ranksecond{30.4} & \rankthird{45.3} & \rankthird{0.1793} & \rankthird{75.4} & \rankthird{30.8} \\
MVTracker \; + \textit{VGGT-\ensuremath{\Omega}}\textsuperscript{*} & 49.1 & 0.2923 & 63.6 & 28.5 & \rankthird{49.3} & \ranksecond{0.1057} & 78.3 & \ranksecond{39.9} & 28.4 & 0.2279 & 72.5 & 18.0 & 42.3 & 0.2087 & 71.5 & 28.8 \\
LAPA \; + \textit{VGGT-\ensuremath{\Omega}}\textsuperscript{*} & 41.4 & 0.3668 & \rankthird{85.8} & 31.0 & 31.3 & 0.1543 & \rankthird{85.2} & 24.4 & 13.6 & 0.3954 & 51.6 & 7.0 & 28.8 & 0.3055 & 74.2 & 20.8 \\
TAPIP3D \; + \textit{VGGT-\ensuremath{\Omega}}\textsuperscript{*} & 50.1 & 0.2835 & \ranksecond{87.6} & \rankthird{38.8} & \ranksecond{50.2} & \rankfirst{0.0991} & \ranksecond{88.6} & \rankthird{38.6} & \rankthird{32.4} & \rankthird{0.2002} & \ranksecond{83.5} & \rankthird{21.8} & 44.3 & 0.1943 & \ranksecond{86.6} & \ranksecond{33.1} \\
        \litemidrule
        4RC\textsuperscript{\textdagger} & 56.9 & \rankthird{0.1973} & {---} & {---} & 42.9 & 0.1151 & {---} & {---} & 32.1 & 0.2333 & {---} & {---} & 44.0 & 0.1819 & {---} & {---} \\
        OmniX\textsuperscript{\textdagger} & \ranksecond{68.4} & \ranksecond{0.1417} & {---} & {---} & 45.6 & 0.1215 & {---} & {---} & 30.0 & 0.2470 & {---} & {---} & \ranksecond{48.0} & \ranksecond{0.1701} & {---} & {---} \\
        \midrule
        \textbf{\method{}} & \rankfirst{79.2} & \rankfirst{0.0929} & \rankfirst{89.9} & \rankfirst{66.9} & \rankfirst{54.8} & 0.1705 & \rankfirst{89.4} & \rankfirst{44.2} & \rankfirst{49.0} & \rankfirst{0.1121} & \rankfirst{86.2} & \rankfirst{36.5} & \rankfirst{61.0} & \rankfirst{0.1252} & \rankfirst{88.5} & \rankfirst{49.2} \\
        \bottomrule
    \end{tabularx}
    \par
    \begin{minipage}{\linewidth}
        \textsuperscript{*} August 2026 checkpoint.\quad \textsuperscript{\textdagger} No meaningful visibility prediction; AJ/OA are unavailable.
    \end{minipage}
\end{table}

\begin{table}[t]
    \centering
    \scriptsize
    \setlength{\tabcolsep}{2pt}
    \renewcommand{\arraystretch}{1.2}
    \sisetup{mode=text,reset-text-series=false}
    \caption{
        \textbf{Pose estimation and 3D reconstruction.}
        We compare against dedicated 3D reconstruction models (top) and 4D reconstruction and tracking approaches (bottom).
        Although \method{} is not solely trained for pose prediction or 3D reconstruction, it still achieves competitive results.
    }
\label{tab:3d-reconstruction-paper-ready}
    \begin{tabularx}{\linewidth}{@{}>{\raggedright\arraybackslash}X*{2}{*{2}{S[table-format=1.4]}S[table-format=1.3]}*{2}{S[table-format=1.4]}S[table-format=2.3]*{2}{S[table-format=1.4]S[table-format=1.3]}@{}}
        \toprule
                                   & \multicolumn{9}{c}{\textbf{Camera pose}}   & \multicolumn{4}{c}{\textbf{3D reconstruction}}                                                                                                                                                                                                                                                                                                                                                                                                                                                                                                                                                                                  \\
        \cmidrule(lr){2-10}
        \cmidrule(lr){11-14}
        \textbf{Method} & \multicolumn{3}{c}{\textbf{Bonn}}          & \multicolumn{3}{c}{\textbf{Sintel}}                     & \multicolumn{3}{c}{\textbf{Panoptic Studio}\textsuperscript{$\dagger$}} & \multicolumn{2}{c}{\textbf{NRGBD}}         & \multicolumn{2}{c}{\textbf{7-Scenes}}                                                                                                                                                                                                                                                                                                                                                                                                                         \\
        \cmidrule(lr){2-4}
        \cmidrule(lr){5-7}
        \cmidrule(lr){8-10}
        \cmidrule(lr){11-12}
        \cmidrule(lr){13-14}
        & {\mbox{ATE \ensuremath{\downarrow}}} & {\mbox{RPE-T \ensuremath{\downarrow}}} & {\mbox{RPE-R \ensuremath{\downarrow}}} & {\mbox{ATE \ensuremath{\downarrow}}} & {\mbox{RPE-T \ensuremath{\downarrow}}} & {\mbox{RPE-R \ensuremath{\downarrow}}} & {\mbox{ATE \ensuremath{\downarrow}}} & {\mbox{RPE-T \ensuremath{\downarrow}}} & {\mbox{RPE-R \ensuremath{\downarrow}}} & {\mbox{Chamfer \ensuremath{\downarrow}}} & {\mbox{NC \ensuremath{\uparrow}}} & {\mbox{Chamfer \ensuremath{\downarrow}}} & {\mbox{NC \ensuremath{\uparrow}}} \\
        \midrule
        VGGT                                                          & 0.0151                                     & 0.0066                                                  & 0.223                                                 & 0.1657                                     & 0.0512                                                  & 0.484                                                 & 0.4310                                     & 0.9849                                                  & 16.914                                                & 0.0429                                     & 0.910                                     & 0.0538                                     & \rankthird{0.754}             \\
        $\pi^3$                                                           & 0.0114                                     & 0.0063                                                  & 0.225                                                 & \ranksecond{0.0735} & 0.0417                                                  & \ranksecond{0.263}             & 0.1506                                     & 0.3755                                                  & 5.663                                                 & 0.0300                                     & 0.910                                     & 0.0721                                     & 0.736                                     \\
VGGT-\ensuremath{\Omega}\textsuperscript{*}                                      & \rankfirst{0.0081}    & 0.0041                                                  & \rankthird{0.213}                         & 0.0777                                     & \rankfirst{0.0305}                 & \rankfirst{0.195}                & 0.3195                                     & 0.6475                                                  & 8.947                                                 & 0.0354                                     & 0.895                                     & \rankfirst{0.0411}    & \rankthird{0.754}             \\
        DA3                                                           & \rankthird{0.0088}             & 0.0047                                                  & \ranksecond{0.212}             & 0.1287                                     & 0.0531                                                  & 0.357                                                 & 0.2084                                     & 0.5030                                                  & 6.037                                                 & \ranksecond{0.0287} & \ranksecond{0.927} & 0.0598                                     & 0.752                                     \\
        \litemidrule
        V-DPM                                                         & 0.0118                                     & 0.0064                                                  & 0.219                                                 & 0.1284                                     & 0.0544                                                  & 0.461                                                 & 0.2146                                     & 0.3906                                                  & 3.797                                                 & 0.0397                                     & 0.887                                     & 0.0739                                     & 0.730                                     \\
        Any4D                                                         & 0.0277                                     & 0.0155                                                  & 0.472                                                 & 0.2228                                     & 0.1132                                                  & 2.120                                                 & 0.5788                                     & 1.3593                                                  & 30.033                                                & 0.0691                                     & 0.845                                     & 0.0969                                     & 0.696                                     \\
        4RC                                                           & \rankfirst{0.0081}    & \ranksecond{0.0037}              & \rankfirst{0.211}                & 0.1292                                     & 0.0527                                                  & 0.396                                                 & \rankthird{0.1414}             & \rankthird{0.2785}                          & \rankthird{1.942}                         & 0.0374                                     & \rankthird{0.911}             & 0.0616                                     & \rankthird{0.754}             \\
        OpenD4RT                                                      & 0.0289                                     & 0.0062                                                  & 0.331                                                 & 0.2328                                     & 0.1256                                                  & 1.833                                                 & 1.5576                                     & 2.8762                                                  & 78.012                                                & 0.4077                                     & 0.592                                     & 0.2503                                     & 0.556                                     \\
        OmniX                                                         & \ranksecond{0.0084} & \rankthird{0.0040}                          & 0.214                                                 & \rankthird{0.0770}             & \rankthird{0.0352}                          & \rankthird{0.291}                         & \ranksecond{0.0786} & \ranksecond{0.1459}              & \ranksecond{1.824}             & \rankthird{0.0294}             & \ranksecond{0.927} & \ranksecond{0.0481} & \rankfirst{0.769}    \\
        SM4RT                                                         & 0.0168                                     & 0.0075                                                  & 0.240                                                 & 0.1120                                     & 0.0445                                                  & 0.422                                                 & 0.1483                                     & 0.3366                                                  & 3.281                                                 & 0.1082                                     & 0.776                                     & 0.0831                                     & 0.691                                     \\
        \midrule
        \textbf{\method{}}                                            & \ranksecond{0.0084} & \rankfirst{0.0036}                 & \rankfirst{0.211}                & \rankfirst{0.0613}    & \ranksecond{0.0337}              & 0.453                                                 & \rankfirst{0.0714}    & \rankfirst{0.1301}                 & \rankfirst{1.406}                & \rankfirst{0.0263}    & \rankfirst{0.933}    & \rankthird{0.0487}             & \ranksecond{0.762} \\
        \bottomrule
    \end{tabularx}
    \par
    \begin{minipage}{\linewidth}
        \textsuperscript{*} August 2026 checkpoint.
        \quad
        \textsuperscript{$\dagger$} {Multi-camera setup capturing dynamic scenes simultaneously from multiple viewpoints.}
    \end{minipage}
\end{table}

\paragraph{3D reconstruction.}
Beyond tracking, we evaluate camera pose estimation and scene reconstruction.
For pose estimation, we evaluate on Bonn~\citep{palazzolo2019bonn}, Sintel~\citep{butler2012sintel}, and Panoptic Studio.
For reconstruction, we evaluate on NRGBD~\citep{azinovic2022nrgbd} and 7-Scenes~\citep{shotton2013sevenscenes}.
We recover camera poses by aligning a sparse set of predicted 3D points across coordinate frames, without a dedicated pose head (\cref{sup:camera_recovery}).
Pose accuracy therefore also reflects the consistency of the predicted geometry.
Results are shown in \cref{tab:3d-reconstruction-paper-ready}.

\method{} achieves the lowest ATE on Sintel and leads across all three pose
metrics on Panoptic Studio, followed by OmniX. Both methods train with multi-view
dynamic observations, suggesting that such training benefits multi-camera pose
recovery in dynamic scenes. On Bonn, \method{} achieves the lowest relative
translation error and ties for the lowest relative rotation error.

For scene reconstruction, \method{} achieves the lowest Chamfer distance and
highest normal consistency on NRGBD and remains competitive on 7-Scenes.
Video depth results (\cref{tab:video-depth-paper-ready}) further demonstrate
its geometric accuracy: \method{} leads on both Bonn metrics, achieves the
lowest AbsRel on KITTI~\citep{uhrig2017kitti_depth}, and attains the highest $\delta_1$ on 7-Scenes.
\Cref{fig:video_depth} in the appendix provides a qualitative comparison of
video depth predictions.

\begin{table}[t]
    \centering
    \begin{minipage}[t]{0.48\linewidth}
        \scriptsize
\setlength{\tabcolsep}{2pt}
\renewcommand{\arraystretch}{1.2}
\sisetup{mode=text,reset-text-series=false}
\centering
\captionof{table}{
    \textbf{Video depth.}
    We report AbsRel and $\delta_1$ results after per-sequence scale-only, depth-relative L1 alignment.
}
\label{tab:video-depth-paper-ready}
\begin{tabularx}{\linewidth}{@{}>{\raggedright\arraybackslash}X*{6}{S[table-format=1.3]}@{}}
    \toprule
    \multirow{2}{*}[-0.5\dimexpr\aboverulesep+\cmidrulewidth+\belowrulesep\relax]{\textbf{Method}}                           & \multicolumn{2}{c}{\textbf{Bonn}}         & \multicolumn{2}{c}{\textbf{KITTI}}        & \multicolumn{2}{c}{\textbf{7-Scenes}}                                                                                                                                      \\
    \cmidrule(lr){2-3}
    \cmidrule(lr){4-5}
    \cmidrule(lr){6-7}
    & {\mbox{AbsRel \ensuremath{\downarrow}}} & {\mbox{$\delta_1$ \ensuremath{\uparrow}}} & {\mbox{AbsRel \ensuremath{\downarrow}}} & {\mbox{$\delta_1$ \ensuremath{\uparrow}}} & {\mbox{AbsRel \ensuremath{\downarrow}}} & {\mbox{$\delta_1$ \ensuremath{\uparrow}}} \\
    \midrule
    VGGT                                                          & 0.053                                     & 0.975                                     & 0.056                                     & 0.961                                     & 0.212                                     & 0.896                                     \\
    $\pi^3$                                                           & \ranksecond{0.042} & \ranksecond{0.986} & \ranksecond{0.037} & \ranksecond{0.986} & 0.219                                     & 0.876                                     \\
VGGT-\ensuremath{\Omega}\textsuperscript{*}                                      & 0.115                                     & 0.937                                     & \rankthird{0.038}             & \rankfirst{0.988}    & \rankthird{0.210}             & 0.910                                     \\
    DA3                                                           & \ranksecond{0.042} & 0.982                                     & 0.051                                     & 0.976                                     & \ranksecond{0.209} & \rankthird{0.919}             \\
    \litemidrule
    V-DPM                                                         & 0.049                                     & \rankthird{0.985}             & 0.068                                     & 0.943                                     & \rankthird{0.210}             & \ranksecond{0.921} \\
    Any4D                                                         & 0.052                                     & 0.979                                     & 0.066                                     & 0.966                                     & 0.229                                     & 0.866                                     \\
    4RC                                                           & \rankthird{0.043}             & 0.983                                     & 0.053                                     & 0.953                                     & \rankfirst{0.204}    & 0.903                                     \\
    OpenD4RT                                                      & 0.071                                     & 0.952                                     & 0.112                                     & 0.854                                     & 0.223                                     & 0.913                                     \\
    OmniX                                                         & \ranksecond{0.042} & 0.984                                     & 0.079                                     & 0.929                                     & 0.214                                     & 0.870                                     \\
    SM4RT                                                         & 0.051                                     & 0.979                                     & \ranksecond{0.037} & 0.977                                     & 0.211                                     & 0.904                                     \\
    \midrule
    \textbf{\method{}}                                            & \rankfirst{0.041}    & \rankfirst{0.987}    & \rankfirst{0.036}    & \rankthird{0.983}             & 0.213                                     & \rankfirst{0.924}    \\
    \bottomrule
\end{tabularx}
\par
\begin{minipage}{\linewidth}
    \textsuperscript{*} August 2026 checkpoint. 
\end{minipage}

    \end{minipage}\hfill
    \begin{minipage}[t]{0.48\linewidth}
        \input{fig/feature_matching_uv}
    \end{minipage}
\end{table}

\begin{table}[t]
    \renewcommand{\cmark}{\textcolor{black!50}{\ding{51}}}
    \newcommand{\dartquery}{\raisebox{-0.25em}{\includegraphics[height=1em]{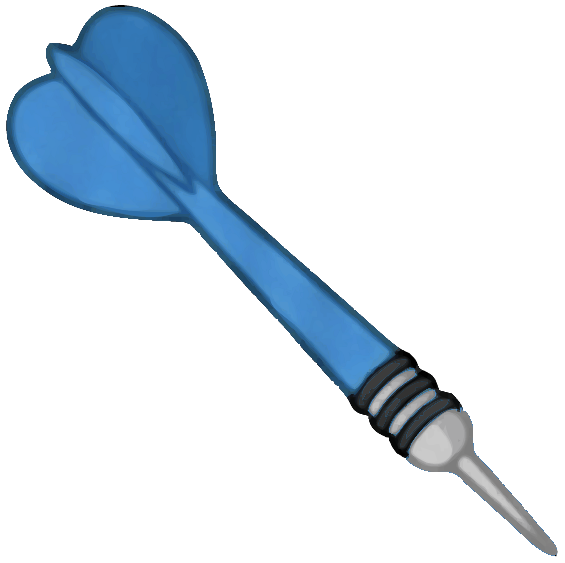}}}
    \newcommand{\arrowquery}{\raisebox{-0.25em}{\includegraphics[height=1em]{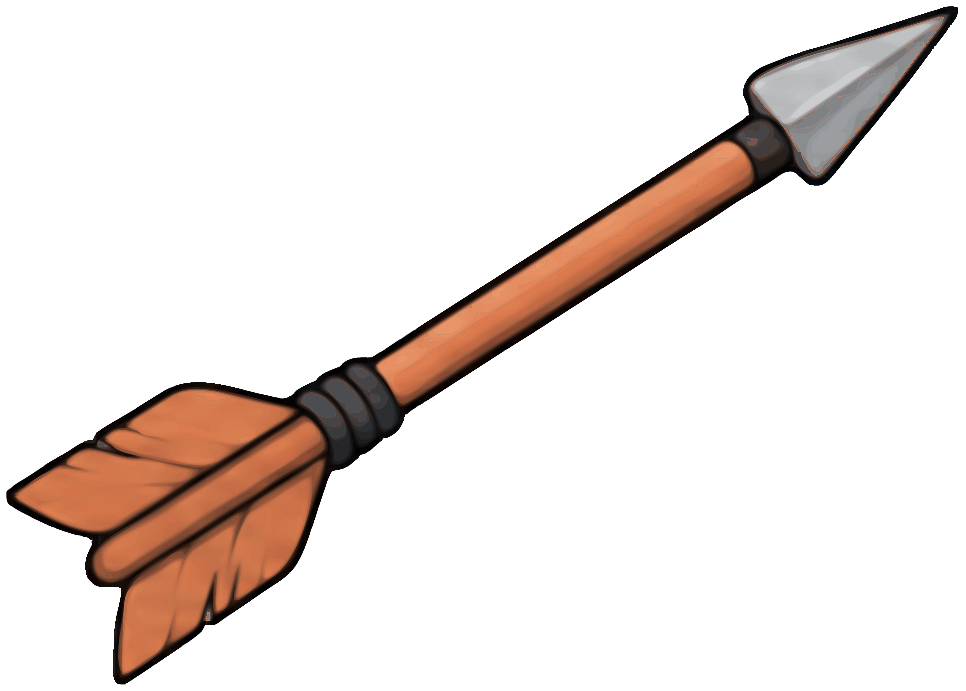}}}
    \centering
    \tiny
    \setlength{\tabcolsep}{3pt}
    \renewcommand{\arraystretch}{1.2}
    \sisetup{mode=text,reset-text-series=false,retain-explicit-plus=true,table-format=-2.2}
    \renewcommand{\tabularxcolumn}[1]{S[table-column-width=#1]}
    \setlength{\bigstrutjot}{\dimexpr\aboverulesep+\cmidrulewidth+\belowrulesep\relax}
    \newcommand{\ablationheader}[1]{\multirow{3}[b2]{*}{\rotatebox{90}{\bfseries\shortstack{#1}}}}
\caption{
        \textbf{Cumulative ablations.} S5 is the $100\mathrm{k}$-step DA3-Large ablation model and reports absolute scores;
        other rows report signed differences from S5. APD, AUC, and F1 differences are in percentage points.
        \dartquery{}: order-dependent querying (D4RT); \arrowquery{}: order-invariant querying (\method{}).
        \enquote{video} comprises the datasets in \cref{tab:training-data} excluding Hypersim, ScanNet++, BlendedMVS, and Structured3D.
}
    \label{tab:ablation-paper-main}
    \begin{tabularx}{\linewidth}{@{}*{6}{c}*{2}{X>{\sisetup{table-format=+1.4}}X}*{4}{X}@{}}
        \toprule
        \ablationheader{Stage} & \ablationheader{Query\\Approach} & \ablationheader{PE} & \ablationheader{Sequential\\Sampling} & \ablationheader{Data} & \ablationheader{Eval} & \multicolumn{4}{c}{\textbf{Tracking}} & \multicolumn{4}{c}{\textbf{Reconstruction}} \\
        \cmidrule(lr){7-10}
        \cmidrule(lr){11-14}
        & & & & & & \multicolumn{2}{c}{\textbf{Monocular}} & \multicolumn{2}{c}{\textbf{Multi-view}} & \multicolumn{2}{c}{\textbf{Monocular}} & \multicolumn{2}{c}{\textbf{Multi-view}} \\
        \cmidrule(lr){7-8}
        \cmidrule(lr){9-10}
        \cmidrule(lr){11-12}
        \cmidrule(lr){13-14}
         &  &  &  &  &  & {\mbox{APD \ensuremath{\uparrow}}} & {\mbox{EPE \ensuremath{\downarrow}}} & {\mbox{APD \ensuremath{\uparrow}}} & {\mbox{EPE \ensuremath{\downarrow}}} & {\mbox{AUC@30 \ensuremath{\uparrow}}} & {\mbox{F1 \ensuremath{\uparrow}}} & {\mbox{AUC@30 \ensuremath{\uparrow}}} & {\mbox{F1 \ensuremath{\uparrow}}} \\
        \midrule
        \multirow{2}{*}{S1}                                                                               & \dartquery                                                                           & \cmark                                                                        & \cmark                                                                           & video                                                                            & windowed                                                                            & \cellcolor[HTML]{F0BFC4}-8.41             & \cellcolor[HTML]{F0BFC4}+0.0822             & \cellcolor[HTML]{F1C4C7}-34.23            & \cellcolor[HTML]{F3CCCC}+0.4416             & \cellcolor[HTML]{F1C3C6}-19.32            & \cellcolor[HTML]{F0BFC4}-31.36            & \cellcolor[HTML]{F0C0C4}-61.61            & \cellcolor[HTML]{F0BFC4}-26.72            \\
                                              & \dartquery                                                                           & \cmark                                                                        & \cmark                                                                           & video                                                                            & interpolated\textsuperscript{\textdagger}                                                                            & \cellcolor[HTML]{F7DDD7}-3.25             & \cellcolor[HTML]{F7DCD7}+0.0322             & \cellcolor[HTML]{F0BFC4}-38.68            & \cellcolor[HTML]{F0BFC4}+0.6257             & \cellcolor[HTML]{F7DED8}-7.62             & \cellcolor[HTML]{F0BFC4}-31.36            & \cellcolor[HTML]{F0BFC4}-62.63            & \cellcolor[HTML]{F0BFC4}-26.72            \\
        \multirow{2}{*}{S2}                                                                               & \arrowquery                                                                           & \cmark                                                                        & \cmark                                                                           & video                                                                            & windowed                                                                            & \cellcolor[HTML]{F2C6C8}-7.06             & \cellcolor[HTML]{F2C8CA}+0.0651             & \cellcolor[HTML]{F2C8CA}-30.37            & \cellcolor[HTML]{F4CFCE}+0.3995             & \cellcolor[HTML]{F0BFC4}-21.14            & \cellcolor[HTML]{F5D3D1}-17.86            & \cellcolor[HTML]{F0C1C5}-59.81            & \cellcolor[HTML]{F0BFC4}-26.53            \\
                                              & \arrowquery                                                                           & \cmark                                                                        & \cmark                                                                           & video                                                                            & interpolated\textsuperscript{\textdagger}                                                                            & \cellcolor[HTML]{F9E4DC}-2.18             & \cellcolor[HTML]{F9E4DC}+0.0213             & \cellcolor[HTML]{F3CCCC}-27.23            & \cellcolor[HTML]{F5D3D1}+0.3567             & \cellcolor[HTML]{F7DBD6}-8.71             & \cellcolor[HTML]{F5D3D1}-17.86            & \cellcolor[HTML]{F1C3C6}-56.94            & \cellcolor[HTML]{F0BFC4}-26.53            \\
        S3                                                                               & \arrowquery                                                                           & \xmark                                                          & \cmark                                                                           & video                                                                            & full                                                                           & \cellcolor[HTML]{F9E6DD}-2.00             & \cellcolor[HTML]{F9E4DC}+0.0217             & \cellcolor[HTML]{F3CACB}-28.89            & \cellcolor[HTML]{F4CECE}+0.4142             & \cellcolor[HTML]{F7DED8}-7.55             & \cellcolor[HTML]{F4D1CF}-18.97            & \cellcolor[HTML]{F4D0CF}-38.41            & \cellcolor[HTML]{F2C9CB}-20.33            \\
        S4                                                                               & \arrowquery                                                                           & \xmark                                                          & \xmark                                                             & video                                                                            & full                                                                           & \cellcolor[HTML]{FAEAE0}\rankunderline{-1.52} & \cellcolor[HTML]{FAEAE0}\rankunderline{+0.0144} & \cellcolor[HTML]{FFFEED}\rankunderline{-0.04} & \cellcolor[HTML]{FEFAEA}\rankunderline{+0.0108} & \cellcolor[HTML]{FDF7E8}\rankunderline{-0.91} & \cellcolor[HTML]{FDF7E8}\rankunderline{-1.16} & \cellcolor[HTML]{FEFBEA}\rankunderline{-0.88} & \cellcolor[HTML]{FAE9DF}\rankunderline{-5.03} \\
        \midrule
        \textbf{S5} & \arrowquery & \xmark & \xmark & all & full & \bfseries 80.01 & \bfseries 0.1983 & \bfseries 48.49 & \bfseries 0.1526 & \bfseries 51.10 & \bfseries 76.59 & \bfseries 63.61 & \bfseries 52.02 \\
        \bottomrule
    \end{tabularx}
    \par
    \begin{minipage}{\linewidth}
        \textsuperscript{\textdagger} Indicates repeated evaluation with full context and interpolated temporal encodings.
    \end{minipage}
\end{table}

\paragraph{Cumulative Ablations.}
\label{sec:ablations}
We evaluate cumulative progression from a D4RT-inspired baseline to the \method{} design across tracking, pose estimation, and reconstruction (\cref{tab:ablation-paper-main}).
For compact comparisons, we average results across datasets within each task and input
setting: monocular tracking uses WorldTrack, monocular pose uses Bonn and Sintel, and monocular reconstruction uses NRGBD.
Multi-view results average Multi-View Kubric, DexYCB, and Panoptic Studio.

The baseline (\textbf{S1}) adopts D4RT's ordered-video formulation, replacing VideoMAEv2~\citep{wang2023videomaev2} with our pretrained 3D backbone, i.e., a direct backbone swap.
Sinusoidal sequence-position embeddings provide ordering information to the encoder's camera tokens, while learned discrete embeddings identify each query's source, target, and reference images.
For inputs exceeding the $24$-image training limit, S1 and S2 use either windowed inference (Win; anchored image-block pairs for multi-view tracking) or full-context inference with temporal encodings interpolated over the training range (Int, $\dagger$).
S3, S4, and S5 process the full context directly (Full).
Scene-reconstruction F1 uses full context with at most $24$ images and is shared between paired Win/Int rows; pose AUC uses each row's inference mode.
Replacing discrete query embeddings with identity-based conditioning (\textbf{S2}) improves monocular and multi-view tracking under both inference modes.
The larger multi-view gain with interpolation supports content-derived observation identities for querying beyond the training context length.
Removing the encoder's temporal embeddings (\textbf{S3}) leaves monocular tracking nearly unchanged relative to S2 with interpolation, which also uses full context.
Effects on other tasks are mixed.
Overall, this design remains competitive without explicit temporal encoding.
Non-sequential sampling (\textbf{S4}) introduces cross-view and non-contiguous training observations and yields the largest multi-view tracking gain: $28.85$ percentage points in APD relative to S3.
Monocular tracking, pose estimation, and reconstruction also improve.
Adding four static multi-view datasets (\textbf{S5}) expands the mixture from $15$ to $19$ datasets and improves every reported metric, supporting broader training data for both monocular and multi-view inputs.

Additional design comparisons and $24$-frame results appear in \cref{sup:additional_ablations}.
\cref{sup:longtracking} compares full-context and windowed inference on longer sequences.

\section{Conclusion}
\label{sec:conclusion}

In this work, we introduced \method{}, a feed-forward model capable of reconstructing and tracking across arbitrary observations.
This capability is enabled by our order-invariant querying mechanism, which uses content-derived ID~tokens, replacing order-dependent learnable embeddings.
Our ablations show that broadening training beyond monocular video, together with the proposed query mechanism, improves generalization and task performance.
Trained with this recipe, \method{} sets a new state of the art in multiple tracking benchmarks, while maintaining competitive results in 3D reconstruction.
We further discuss the limitations of \method{} in~\cref{sec:limitations}. To facilitate reproducibility and support future research, we make our code and model weights publicly available at \mbox{\url{https://vision.rwth-aachen.de/arrow}}.

\FloatBarrier

\subsubsection*{Acknowledgments}
We acknowledge funding by BMFTR project \enquote{WestAI} (grant no.\ \texttt{16IS22094D}) and the EU project \enquote{JUPITER AI Factory} (grant no.\ \texttt{101250682}).
Computations were performed using resources granted by RWTH Aachen under projects \texttt{rwth1742}, \texttt{rwth1788}, and \texttt{rwth1968} and by the Gauss Centre for Supercomputing e.V.\ 
through the John von Neumann Institute for Computing on the GCS Supercomputer JUWELS at the Jülich Supercomputing Centre.

\bibliography{iclr2027_conference}

@inproceedings{butler2012sintel,
    title         = {A Naturalistic Open Source Movie for Optical Flow Evaluation},
    author        = {Butler, Daniel J. and Wulff, Jonas and Stanley, Garrett B. and Black, Michael J.},
    year          = 2012,
    booktitle     = {ECCV},
}

@inproceedings{schoenberger2016colmap,
    title         = {Structure-From-Motion Revisited},
    author        = {Sch{\"o}nberger, Johannes Lutz and Frahm, Jan-Michael},
    year          = 2016,
    booktitle     = {CVPR},
}

@inproceedings{uhrig2017kitti_depth,
    title         = {Sparsity Invariant {CNN}s},
    author        = {Uhrig, Jonas and Schneider, Nick and Schneider, Lukas and Franke, Uwe and Brox, Thomas and Geiger, Andreas},
    year          = 2017,
    booktitle     = {3DV},
}

@inproceedings{huang2018mvssynth,
    title         = {{DeepMVS}: Learning Multi-View Stereopsis},
    author        = {Huang, Po-Han and Matzen, Kevin and Kopf, Johannes and Ahuja, Narendra and Huang, Jia-Bin},
    year          = 2018,
    booktitle     = {CVPR},
}

@inproceedings{bao2022beit,
    title         = {{BE}iT: {BERT} Pre-Training of Image Transformers},
    author        = {Bao, Hangbo and Dong, Li and Piao, Songhao and Wei, Furu},
    year          = 2022,
    booktitle     = {ICLR},
}

@inproceedings{loshchilov2019adamw,
    title         = {Decoupled Weight Decay Regularization},
    author        = {Loshchilov, Ilya and Hutter, Frank},
    year          = 2019,
    booktitle     = {ICLR},
}

@inproceedings{fonder2019midair,
    title         = {Mid-Air: A Multi-Modal Dataset for Extremely Low Altitude Drone Flights},
    author        = {Fonder, Michael and Van Droogenbroeck, Marc},
    year          = 2019,
    booktitle     = {CVPR Workshops},
}

@inproceedings{caesar2020nuscenes,
    title         = {{nuScenes}: A Multimodal Dataset for Autonomous Driving},
    author        = {Caesar, Holger and Bankiti, Varun and Lang, Alex H. and Vora, Sourabh and Liong, Venice Erin and Xu, Qiang and Krishnan, Anush and Pan, Yu and Baldan, Giancarlo and Beijbom, Oscar},
    year          = 2020,
    booktitle     = {CVPR},
}

@inproceedings{zheng2020structured3d,
    title         = {{Structured3D}: A Large Photo-realistic Dataset for Structured {3D} Modeling},
    author        = {Zheng, Jia and Zhang, Junfei and Li, Jing and Tang, Rui and Gao, Shenghua and Zhou, Zihan},
    year          = 2020,
    booktitle     = {ECCV},
}

@inproceedings{sun2020waymo,
    title         = {Scalability in Perception for Autonomous Driving: {Waymo} Open Dataset},
    author        = {Sun, Pei and Kretzschmar, Henrik and Dotiwalla, Xerxes and Chouard, Aurelien and Patnaik, Vijaysai and Tsui, Paul and Guo, James and Zhou, Yin and Chai, Yuning and Caine, Benjamin and Vasudevan, Vijay and Han, Wei and Ngiam, Jiquan and Zhao, Hang and Timofeev, Aleksei and Ettinger, Scott and Krivokon, Maxim and Gao, Amy and Joshi, Aditya and Zhang, Yu and Shlens, Jonathon and Chen, Zhifeng and Anguelov, Dragomir},
    year          = 2020,
    booktitle     = {CVPR},
}

@inproceedings{yao2020blendedmvs,
    title         = {{BlendedMVS}: A Large-Scale Dataset for Generalized Multi-View Stereo Networks},
    author        = {Yao, Yao and Luo, Zixin and Li, Shiwei and Zhang, Jingyang and Ren, Yufan and Zhou, Lei and Fang, Tian and Quan, Long},
    year          = 2020,
    booktitle     = {CVPR},
}

@article{wang2020gtasfm,
    title         = {Flow-Motion and Depth Network for Monocular Stereo and Beyond},
    author        = {Wang, Kaixuan and Shen, Shaojie},
    year          = 2020,
    journal       = {IEEE RAL},
}

@inproceedings{dosovitskiy2021vit,
    title         = {An Image is Worth {16x16} Words: Transformers for Image Recognition at Scale},
    author        = {Dosovitskiy, Alexey and Beyer, Lucas and Kolesnikov, Alexander and Weissenborn, Dirk and Zhai, Xiaohua and Unterthiner, Thomas and Dehghani, Mostafa and Minderer, Matthias and Heigold, Georg and Gelly, Sylvain and Uszkoreit, Jakob and Houlsby, Neil},
    year          = 2021,
    booktitle     = {ICLR},
}

@inproceedings{roberts2021hypersim,
    title         = {{Hypersim}: A Photorealistic Synthetic Dataset for Holistic Indoor Scene Understanding},
    author        = {Roberts, Mike and Ramapuram, Jason and Ranjan, Anurag and Kumar, Atulit and Bautista, Miguel Angel and Paczan, Nathan and Webb, Russ and Susskind, Joshua M.},
    year          = 2021,
    booktitle     = {ICCV},
}

@inproceedings{yeshwanthliu2023scannetpp,
    title         = {{ScanNet++}: A High-Fidelity Dataset of {3D} Indoor Scenes},
    author        = {Yeshwanth, Chandan and Liu, Yueh-Cheng and Nie{\ss}ner, Matthias and Dai, Angela},
    year          = 2023,
    booktitle     = {ICCV},
}

@inproceedings{leroy2024mast3r,
    title         = {Grounding Image Matching in {3D} with {MASt3R}},
    author        = {Leroy, Vincent and Cabon, Yohann and Revaud, J{\'e}r{\^o}me},
    year          = 2024,
    booktitle     = {ECCV},
}

@inproceedings{wang2025vggt,
    title         = {{VGGT}: Visual Geometry Grounded Transformer},
    author        = {Wang, Jianyuan and Chen, Minghao and Karaev, Nikita and Vedaldi, Andrea and Rupprecht, Christian and Novotny, David},
    year          = 2025,
    booktitle     = {CVPR},
}

@inproceedings{yang2025fast3r,
    title         = {{Fast3R}: Towards {3D} Reconstruction of 1000+ Images in One Forward Pass},
    author        = {Yang, Jianing and Sax, Alexander and Liang, Kevin J. and Henaff, Mikael and Tang, Hao and Cao, Ang and Chai, Joyce and Meier, Franziska and Feiszli, Matt},
    year          = 2025,
    booktitle     = {CVPR},
}

@article{piccinelli2025unidepthv2,
    title         = {{UniDepthV2}: Universal Monocular Metric Depth Estimation Made Simpler},
    author        = {Piccinelli, Luigi and Sakaridis, Christos and Yang, Yung-Hsu and Segu, Mattia and Li, Siyuan and Abbeloos, Wim and Van Gool, Luc},
    year          = 2026,
    journal       = {IEEE TPAMI},
}

@inproceedings{yu2026infinidepth,
    title         = {{InfiniDepth}: Arbitrary-Resolution and Fine-Grained Depth Estimation with Neural Implicit Fields},
    author        = {Yu, Hao and Lin, Haotong and Wang, Jiawei and Li, Jiaxin and Wang, Yida and Zhang, Xueyang and Wang, Yue and Zhou, Xiaowei and Hu, Ruizhen and Peng, Sida},
    year          = 2026,
    booktitle     = {CVPR},
}

@inproceedings{lin2026depthanything3,
    title         = {{Depth Anything 3}: Recovering the Visual Space from Any Views},
    author        = {Lin, Haotong and Chen, Sili and Liew, Jun Hao and Chen, Donny Y. and Li, Zhenyu and Shi, Guang and Feng, Jiashi and Kang, Bingyi},
    year          = 2026,
    booktitle     = {ICLR},
}

@inproceedings{keetha2026mapanything,
    title         = {{MapAnything}: Universal Feed-Forward Metric {3D} Reconstruction},
    author        = {Keetha, Nikhil and M{\"u}ller, Norman and Sch{\"o}nberger, Johannes and Porzi, Lorenzo and Zhang, Yuchen and Fischer, Tobias and Knapitsch, Arno and Zauss, Duncan and Weber, Ethan and Antunes, Nelson and Luiten, Jonathon and Lopez-Antequera, Manuel and Rota Bul{\`o}, Samuel and Richardt, Christian and Ramanan, Deva and Scherer, Sebastian and Kontschieder, Peter},
    year          = 2026,
    booktitle     = {3DV},
}

@inproceedings{wang2026pi3,
    title         = {{$\pi^3$}: Permutation-Equivariant Visual Geometry Learning},
    author        = {Wang, Yifan and Zhou, Jianjun and Zhu, Haoyi and Chang, Wenzheng and Zhou, Yang and Li, Zizun and Chen, Junyi and Pang, Jiangmiao and Shen, Chunhua and He, Tong},
    year          = 2026,
    booktitle     = {ICLR},
}

@inproceedings{zhang2026d4rt,
    title         = {Efficiently Reconstructing Dynamic Scenes One {D4RT} at a Time},
    author        = {Zhang, Chuhan and Le Moing, Guillaume and Koppula, Skanda and Rocco, Ignacio and Momeni, Liliane and Xie, Junyu and Sun, Shuyang and Sukthankar, Rahul and Barral, Jo{\"e}lle K. and Hadsell, Raia and Ghahramani, Zoubin and Zisserman, Andrew and Zhang, Junlin and Sajjadi, Mehdi S. M.},
    year          = 2026,
    booktitle     = {CVPR},
}

@inproceedings{Wang_2024_DUSt3R,
    title         = {{DUSt3R}: Geometric {3D} Vision Made Easy},
    author        = {Wang, Shuzhe and Leroy, Vincent and Cabon, Yohann and Chidlovskii, Boris and Revaud, Jerome},
    year          = 2024,
    booktitle     = {CVPR},
}

@inproceedings{feng2025st4rtrack,
    title         = {{St4RTrack}: Simultaneous {4D} Reconstruction and Tracking in the World},
    author        = {Feng, Haiwen and Zhang, Junyi and Wang, Qianqian and Ye, Yufei and Yu, Pengcheng and Black, Michael J. and Darrell, Trevor and Kanazawa, Angjoo},
    year          = 2025,
    booktitle     = {ICCV},
}

@inproceedings{sucar2025dpm,
    title         = {Dynamic Point Maps: A Versatile Representation for Dynamic {3D} Reconstruction},
    author        = {Sucar, Edgar and Lai, Zihang and Insafutdinov, Eldar and Vedaldi, Andrea},
    year          = 2025,
    booktitle     = {ICCV},
}

@inproceedings{sucar2026vdpm,
    title         = {{V-DPM}: {4D} Video Reconstruction with Dynamic Point Maps},
    author        = {Sucar, Edgar and Insafutdinov, Eldar and Lai, Zihang and Vedaldi, Andrea},
    year          = 2026,
    booktitle     = {CVPR},
}

@article{liu2025traceanything,
    title         = {Trace Anything: Representing Any Video in {4D} via Trajectory Fields},
    author        = {Liu, Xinhang and Xiao, Yuxi and Chen, Donny Y. and Feng, Jiashi and Tai, Yu-Wing and Tang, Chi-Keung and Kang, Bingyi},
    year          = 2025,
    journal       = {arXiv preprint arXiv:2510.13802},
}

@inproceedings{karhade2025any4d,
    title         = {{Any4D}: Unified Feed-Forward Metric {4D} Reconstruction},
    author        = {Karhade, Jay and Keetha, Nikhil and Zhang, Yuchen and Gupta, Tanisha and Sharma, Akash and Scherer, Sebastian and Ramanan, Deva},
    year          = 2026,
    booktitle     = {CVPR},
}

@article{lin2026sm4rt,
    title         = {{SM4RT}: Learning Structured Motion Geometry for {4D} Reconstruction},
    author        = {Lin, Shing Ho J. and Zheng, Wenzhao and Zhuo, Dong and Wu, Yuqi and Zhou, Jie and Lu, Jiwen},
    year          = 2026,
    journal       = {arXiv preprint arXiv:2607.22534},
}

@inproceedings{ngo2024delta,
    title         = {{DELTA}: Dense Efficient Long-range {3D} Tracking for any video},
    author        = {Ngo, Tuan Duc and Zhuang, Peiye and Gan, Chuang and Kalogerakis, Evangelos and Tulyakov, Sergey and Lee, Hsin-Ying and Wang, Chaoyang},
    year          = 2025,
    booktitle     = {ICLR},
}

@inproceedings{xiao2025spatialtrackerv2,
    title         = {{SpatialTrackerV2}: Advancing {3D} Point Tracking with Explicit Camera Motion},
    author        = {Xiao, Yuxi and Wang, Jianyuan and Xue, Nan and Karaev, Nikita and Makarov, Yuri and Kang, Bingyi and Zhu, Xing and Bao, Hujun and Shen, Yujun and Zhou, Xiaowei},
    year          = 2025,
    booktitle     = {ICCV},
}

@inproceedings{zhang2025tapip3d,
    title         = {{TAPIP3D}: Tracking Any Point in Persistent {3D} Geometry},
    author        = {Zhang, Bowei and Ke, Lei and Harley, Adam W. and Fragkiadaki, Katerina},
    year          = 2025,
    booktitle     = {NeurIPS},
}

@inproceedings{rajic2025mvtracker,
    title         = {Multi-View {3D} Point Tracking},
    author        = {Raji{\v{c}}, Frano and Xu, Haofei and Mihajlovic, Marko and Li, Siyuan and Demir, Irem and G{\"u}ndo{\u{g}}du, Emircan and Ke, Lei and Prokudin, Sergey and Pollefeys, Marc and Tang, Siyu},
    year          = 2025,
    booktitle     = {ICCV},
}

@inproceedings{jiang2026omnix,
    title         = {{OmniX}: Any-view and Any-time {4D} Reconstruction via Feed-forward Trajectory Fields},
    author        = {Jiang, Yanqin and Wang, Tengfei and Wang, Zhengwei and Cao, Chenjie and Wu, Junta and Luo, Wenhan and Hu, Weiming and Gao, Jin and Guo, Chunchao},
    year          = 2026,
    booktitle     = {ECCV},
}

@inproceedings{luo20264rc,
    title         = {{4RC}: {4D} Reconstruction via Conditional Querying Anytime and Anywhere},
    author        = {Luo, Yihang and Zhou, Shangchen and Lan, Yushi and Pan, Xingang and Loy, Chen Change},
    year          = 2026,
    booktitle     = {ICML},
}

@article{chen2026uniquery4r,
    title         = {{UniQuery4R}: Unified {4D} Scene Reconstruction from a Single Query},
    author        = {Chen, Tiancheng and Tang, Sheng and Jin, Wenhua and Zhang, Weiqi and Fang, Juntong and Zhou, Junsheng and Li, Zesong},
    year          = 2026,
    journal       = {arXiv preprint arXiv:2608.17283},
}

@inproceedings{karaev2024cotracker3,
    title         = {{CoTracker3}: Simpler and Better Point Tracking by Pseudo-Labelling Real Videos},
    author        = {Karaev, Nikita and Makarov, Yuri and Wang, Jianyuan and Neverova, Natalia and Vedaldi, Andrea and Rupprecht, Christian},
    year          = 2025,
    booktitle     = {ICCV},
}

@article{ngo2025deltav2,
    title         = {{DELTAv2}: Accelerating Dense {3D} Tracking},
    author        = {Ngo, Tuan Duc and Mirzaei, Ashkan and Qian, Guocheng and Liang, Hanwen and Gan, Chuang and Kalogerakis, Evangelos and Wonka, Peter and Wang, Chaoyang},
    year          = 2025,
    journal       = {arXiv preprint arXiv:2508.01170},
}

@inproceedings{huang2026fastspatial,
    title         = {Fast Spatial Tracking with Visual Geometry Transformer},
    author        = {Huang, Chengjie and Wu, Guile and Bai, Dongfeng and Liu, Bingbing},
    year          = 2026,
    booktitle     = {CVPR},
}

@article{qian2026flow4r,
    title         = {{Flow4R}: Unifying {4D} Reconstruction and Tracking with Scene Flow},
    author        = {Qian, Shenhan and Zhang, Ganlin and Wu, Shangzhe and Cremers, Daniel},
    year          = 2026,
    journal       = {arXiv preprint arXiv:2602.14021},
}

@article{lu2026track4world,
    title         = {{Track4World}: Feedforward World-centric Dense {3D} Tracking of All Pixels},
    author        = {Lu, Jiahao and Xu, Jiayi and Hu, Wenbo and Zhu, Ruijie and Zhao, Chengfeng and Yeung, Sai-Kit and Shan, Ying and Liu, Yuan},
    year          = 2026,
    journal       = {arXiv preprint arXiv:2603.02573},
}

@inproceedings{koo2026mvtap,
    title         = {{MV-TAP}: Tracking Any Point in Multi-View Videos},
    author        = {Koo, Jahyeok and Kim, In{\`e}s Hyeonsu and Kim, Mungyeom and Park, Junghyun and Park, Seohyeon and Kim, Jaeyeong and Yi, Jung and Cho, Seokju and Kim, Seungryong},
    year          = 2026,
    booktitle     = {CVPR},
}

@article{koppula2026tapvidmv,
    title         = {{TAPVid-MV}: A Benchmark for Tracking Any Point in {3D} Across Multiple Views},
    author        = {Koppula, Skanda and Rajic, Frano and Rahman, Abdullah Faiz Ur and Yang, Yi and Rocco, Ignacio and Thakwani, Jeet and Kabra, Rishabh and Zisserman, Andrew and Carreira, Joao and Tang, Siyu and Doersch, Carl and Brostow, Gabriel},
    year          = 2026,
    journal       = {arXiv preprint arXiv:2609.01899},
}

@inproceedings{koppula2024tapvid3d,
    title         = {{TAPVid-3D}: A Benchmark for Tracking Any Point in {3D}},
    author        = {Koppula, Skanda and Rocco, Ignacio and Yang, Yi and Heyward, Joe and Carreira, Jo{\~a}o and Zisserman, Andrew and Brostow, Gabriel and Doersch, Carl},
    year          = 2024,
    booktitle     = {NeurIPS},
}

@inproceedings{wang2026vggtomega,
    title         = {{VGGT-$\Omega$}},
    author        = {Wang, Jianyuan and Chen, Minghao and Zhang, Shangzhan and Karaev, Nikita and Sch{\"o}nberger, Johannes and Labatut, Patrick and Bojanowski, Piotr and Novotny, David and Vedaldi, Andrea and Rupprecht, Christian},
    year          = 2026,
    booktitle     = {CVPR},
}

@inproceedings{zhou2026page4d,
    title         = {{PAGE-4D}: Disentangled pose and geometry estimation for vggt-4d perception},
    author        = {Zhou, Kaichen and Wang, Yuhan and Chen, Grace and Chang, Xinhai and Beaudouin, Gaspard and Zhan, Fangneng and Liang, Paul Pu and Wang, Mengyu},
    year          = 2026,
    booktitle     = {ICLR},
}

@article{hu2025vggt4d,
    title         = {{VGGT4D}: Mining Motion Cues in Visual Geometry Transformers for {4D} Scene Reconstruction},
    author        = {Hu, Yu and Cheng, Chong and Yu, Sicheng and Guo, Xiaoyang and Wang, Hao},
    year          = 2025,
    journal       = {arXiv preprint arXiv:2511.19971},
}

@article{lee2026mvtrack4gen,
    title         = {{MVTrack4Gen}: Multi-View Point Tracking as Geometric Supervision for {4D} Video Generation},
    author        = {Lee, JoungBin and Jung, Jaewoo and Lee, Jongmin and Kim, Tongmin and Kim, Hyunsung and Narihira, Takuya and Fukuda, Kazumi and Koo, Jahyeok and Han, Jisang and Mitsufuji, Yuki and Kim, Seungryong},
    year          = 2026,
    journal       = {arXiv preprint arXiv:2606.26087},
}

@inproceedings{zhang2025monst3r,
    title         = {{MonST3R}: A Simple Approach for Estimating Geometry in the Presence of Motion},
    author        = {Zhang, Junyi and Herrmann, Charles and Hur, Junhwa and Jampani, Varun and Darrell, Trevor and Cole, Forrester and Sun, Deqing and Yang, Ming-Hsuan},
    year          = 2025,
    booktitle     = {ICLR},
}

@inproceedings{li2025megasam,
    title         = {{MegaSaM}: Accurate, Fast and Robust Structure and Motion from Casual Dynamic Videos},
    author        = {Li, Zhengqi and Tucker, Richard and Cole, Forrester and Wang, Qianqian and Jin, Linyi and Ye, Vickie and Kanazawa, Angjoo and Holynski, Aleksander and Snavely, Noah},
    year          = 2025,
    booktitle     = {CVPR},
}

@inproceedings{doersch2023tapir,
    title         = {{TAPIR}: Tracking Any Point with Per-Frame Initialization and Temporal Refinement},
    author        = {Doersch, Carl and Yang, Yi and Vecerik, Mel and Gokay, Dilara and Gupta, Ankush and Aytar, Yusuf and Carreira, Joao and Zisserman, Andrew},
    year          = 2023,
    booktitle     = {ICCV},
}

@inproceedings{xiao2024spatialtracker,
    title         = {{SpatialTracker}: Tracking Any {2D} Pixels in {3D} Space},
    author        = {Xiao, Yuxi and Wang, Qianqian and Zhang, Shangzhan and Xue, Nan and Peng, Sida and Shen, Yujun and Zhou, Xiaowei},
    year          = 2024,
    booktitle     = {CVPR},
}

@inproceedings{wang2025shapeofmotion,
    title         = {Shape of Motion: {4D} Reconstruction from a Single Video},
    author        = {Wang, Qianqian and Ye, Vickie and Gao, Hang and Zeng, Weijia and Austin, Jake and Li, Zhengqi and Kanazawa, Angjoo},
    year          = 2025,
    booktitle     = {ICCV},
}

@inproceedings{badki2026l4p,
    title         = {{L4P}: Towards Unified Low-Level {4D} Vision Perception},
    author        = {Badki, Abhishek and Su, Hang and Wen, Bowen and Gallo, Orazio},
    year          = 2026,
    booktitle     = {3DV},
}

@article{jeon2026point4d,
    title         = {{Point4D}: Long-range {4D} Motion Reconstruction},
    author        = {Jeon, Minsik and Karhade, Jay and Ramanan, Deva and Tulsiani, Shubham},
    year          = 2026,
    journal       = {arXiv preprint arXiv:2609.09145},
}

@inproceedings{schoenberger2016mvs,
    title         = {Pixelwise View Selection for Unstructured Multi-View Stereo},
    author        = {Sch{\"o}nberger, Johannes Lutz and Zheng, Enliang and Pollefeys, Marc and Frahm, Jan-Michael},
    year          = 2016,
    booktitle     = {ECCV},
}

@inproceedings{galoaa2026lapa,
    title         = {Look Around and Pay Attention: Multi-camera Point Tracking Reimagined with Transformers},
    author        = {Galoaa, Bishoy and Bai, Xiangyu and Moezzi, Shayda and Nandi, Utsav and Rangoju, Sai Siddhartha Vivek Dhir and Amraee, Somaieh and Ostadabbas, Sarah},
    year          = 2026,
    booktitle     = {3DV},
}

@article{caruana1997multitask,
    title         = {Multitask Learning},
    author        = {Caruana, Rich},
    year          = 1997,
    journal       = {Machine Learning},
}

@inproceedings{jiang2026syn4d,
    title         = {{Syn4D}: A Multiview Synthetic {4D} Dataset},
    author        = {Jiang, Zeren and Lan, Yushi and Luo, Yihang and Deng, Yufan and Lai, Zihang and Sucar, Edgar and Rupprecht, Christian and Laina, Iro and Larlus, Diane and Zheng, Chuanxia and Vedaldi, Andrea},
    year          = 2026,
    booktitle     = {ECCV},
}

@inproceedings{karaev2023dynamicstereo,
    title         = {{DynamicStereo}: Consistent Dynamic Depth From Stereo Videos},
    author        = {Karaev, Nikita and Rocco, Ignacio and Graham, Benjamin and Neverova, Natalia and Vedaldi, Andrea and Rupprecht, Christian},
    year          = 2023,
    booktitle     = {CVPR},
}

@inproceedings{jin2025stereo4d,
    title         = {{Stereo4D}: Learning How Things Move in {3D} from Internet Stereo Videos},
    author        = {Jin, Linyi and Tucker, Richard and Li, Zhengqi and Fouhey, David and Snavely, Noah and Holynski, Aleksander},
    year          = 2025,
    booktitle     = {CVPR},
}

@inproceedings{zheng2023pointodyssey,
    title         = {{PointOdyssey}: A Large-Scale Synthetic Dataset for Long-Term Point Tracking},
    author        = {Zheng, Yang and Harley, Adam W. and Shen, Bokui and Wetzstein, Gordon and Guibas, Leonidas J.},
    year          = 2023,
    booktitle     = {ICCV},
}

@inproceedings{vanhoorick2024gcd,
    title         = {Generative Camera Dolly: Extreme Monocular Dynamic Novel View Synthesis},
    author        = {Van Hoorick, Basile and Wu, Rundi and Ozguroglu, Ege and Sargent, Kyle and Liu, Ruoshi and Tokmakov, Pavel and Dave, Achal and Zheng, Changxi and Vondrick, Carl},
    year          = 2024,
    booktitle     = {ECCV},
}

@inproceedings{greff2022kubric,
    title         = {{Kubric}: A Scalable Dataset Generator},
    author        = {Greff, Klaus and Belletti, Francois and Beyer, Lucas and Doersch, Carl and Du, Yilun and Duckworth, Daniel and Fleet, David J. and Gnanapragasam, Dan and Golemo, Florian and Herrmann, Charles and Kipf, Thomas and Kundu, Abhijit and Lagun, Dmitry and Laradji, Issam and Liu, Hsueh-Ti and Meyer, Henning and Miao, Yishu and Nowrouzezahrai, Derek and Oztireli, Cengiz and Pot, Etienne and Radwan, Noha and Rebain, Daniel and Sabour, Sara and Sajjadi, Mehdi S. M. and Sela, Matan and Sitzmann, Vincent and Stone, Austin and Sun, Deqing and Vora, Suhani and Wang, Ziyu and Wu, Tianhao and Yi, Kwang Moo and Zhong, Fangcheng and Tagliasacchi, Andrea},
    year          = 2022,
    booktitle     = {CVPR},
}

@inproceedings{balasingam2024drivetrack,
    title         = {{DriveTrack}: A Benchmark for Long-Range Point Tracking in Real-World Videos},
    author        = {Balasingam, Arjun and Chandler, Joseph and Li, Chenning and Zhang, Zhoutong and Balakrishnan, Hari},
    year          = 2024,
    booktitle     = {CVPR},
}

@inproceedings{xia2024wildrgbd,
    title         = {{RGBD} Objects in the Wild: Scaling Real-World {3D} Object Learning from {RGB-D} Videos},
    author        = {Xia, Hongchi and Fu, Yang and Liu, Sifei and Wang, Xiaolong},
    year          = 2024,
    booktitle     = {CVPR},
}

@inproceedings{avetisyan2024scenescript,
    title         = {{SceneScript}: Reconstructing Scenes With An Autoregressive Structured Language Model},
    author        = {Avetisyan, Armen and Xie, Christopher and Howard-Jenkins, Henry and Yang, Tsun-Yi and Aroudj, Samir and Patra, Suvam and Zhang, Fuyang and Frost, Duncan and Holland, Luke and Orme, Campbell and Engel, Jakob and Miller, Edward and Newcombe, Richard and Balntas, Vasileios},
    year          = 2024,
    booktitle     = {ECCV},
}

@inproceedings{perazzi2016davis,
    title         = {A Benchmark Dataset and Evaluation Methodology for Video Object Segmentation},
    author        = {Perazzi, Federico and Pont-Tuset, Jordi and McWilliams, Brian and Van Gool, Luc and Gross, Markus and Sorkine-Hornung, Alexander},
    year          = 2016,
    booktitle     = {CVPR},
}

@inproceedings{pan2023adt,
    title         = {{Aria} Digital Twin: A New Benchmark Dataset for Egocentric {3D} Machine Perception},
    author        = {Pan, Xiaqing and Charron, Nicholas and Yang, Yongqian and Peters, Scott and Whelan, Thomas and Kong, Chen and Parkhi, Omkar and Newcombe, Richard and Ren, Yuheng Carl},
    year          = 2023,
    booktitle     = {ICCV},
}

@inproceedings{joo2015pstudio,
    title         = {Panoptic Studio: A Massively Multiview System for Social Motion Capture},
    author        = {Joo, Hanbyul and Liu, Hao and Tan, Lei and Gui, Lin and Nabbe, Bart and Matthews, Iain and Kanade, Takeo and Nobuhara, Shohei and Sheikh, Yaser},
    year          = 2015,
    booktitle     = {ICCV},
}

@inproceedings{chao2021dexycb,
    title         = {{DexYCB}: A Benchmark for Capturing Hand Grasping of Objects},
    author        = {Chao, Yu-Wei and Yang, Wei and Xiang, Yu and Molchanov, Pavlo and Handa, Ankur and Tremblay, Jonathan and Narang, Yashraj S and Van Wyk, Karl and Iqbal, Umar and Birchfield, Stan and Kautz, Jan and Fox, Dieter},
    year          = 2021,
    booktitle     = {CVPR},
}

@inproceedings{palazzolo2019bonn,
    title         = {{ReFusion}: {3D} Reconstruction in Dynamic Environments for {RGB-D} Cameras Exploiting Residuals},
    author        = {Palazzolo, Emanuele and Behley, Jens and Lottes, Philipp and Giguere, Philippe and Stachniss, Cyrill},
    year          = 2019,
    booktitle     = {IROS},
}

@inproceedings{azinovic2022nrgbd,
    title         = {Neural {RGB-D} Surface Reconstruction},
    author        = {Azinovi{\'c}, Dejan and Martin-Brualla, Ricardo and Goldman, Dan B and Nie{\ss}ner, Matthias and Thies, Justus},
    year          = 2022,
    booktitle     = {CVPR},
}

@inproceedings{shotton2013sevenscenes,
    title         = {Scene Coordinate Regression Forests for Camera Relocalization in {RGB-D} Images},
    author        = {Shotton, Jamie and Glocker, Ben and Zach, Christopher and Izadi, Shahram and Criminisi, Antonio and Fitzgibbon, Andrew},
    year          = 2013,
    booktitle     = {CVPR},
}

@inproceedings{knaebel2026surge,
    title         = {{SurGe}: Improved Surface Geometry in Point Maps},
    author        = {Knaebel, Karim and Martin Garcia, Gonzalo and Schmidt, Christian and Fradlin, Ilya and Nunes, Lucas and de Geus, Daan and Leibe, Bastian},
    year          = 2026,
    booktitle     = {NeurIPS},
}

@inproceedings{kong2026finedetailmonoculargeometryestimation,
    title         = {Fine-Detail Monocular Geometry Estimation with Self-Guided Sparse Volumetric Refinement},
    author        = {Lingyu Kong and Ruicheng Li and Ruicheng Wang and Sicheng Xu and Chengtang Yao and Jianfeng Xiang and Jiaolong Yang},
    year          = 2026,
    booktitle     = {NeurIPS},
}

@inproceedings{Grauman_2024_CVPR,
    title         = {Ego-Exo4D: Understanding Skilled Human Activity from First- and Third-Person Perspectives},
    author        = {Grauman, Kristen and Westbury, Andrew and Torresani, Lorenzo and Kitani, Kris and Malik, Jitendra and Afouras, Triantafyllos and Ashutosh, Kumar and Baiyya, Vijay and Bansal, Siddhant and Boote, Bikram and Byrne, Eugene and Chavis, Zach and Chen, Joya and Cheng, Feng and Chu, Fu-Jen and Crane, Sean and Dasgupta, Avijit and Dong, Jing and Escobar, Maria and Forigua, Cristhian and Gebreselasie, Abrham and Haresh, Sanjay and Huang, Jing and Islam, Md Mohaiminul and Jain, Suyog and Khirodkar, Rawal and Kukreja, Devansh and Liang, Kevin J and Liu, Jia-Wei and Majumder, Sagnik and Mao, Yongsen and Martin, Miguel and Mavroudi, Effrosyni and Nagarajan, Tushar and Ragusa, Francesco and Ramakrishnan, Santhosh Kumar and Seminara, Luigi and Somayazulu, Arjun and Song, Yale and Su, Shan and Xue, Zihui and Zhang, Edward and Zhang, Jinxu and Castillo, Angela and Chen, Changan and Fu, Xinzhu and Furuta, Ryosuke and Gonzalez, Cristina and Gupta, Prince and Hu, Jiabo and Huang, Yifei and Huang, Yiming and Khoo, Weslie and Kumar, Anush and Kuo, Robert and Lakhavani, Sach and Liu, Miao and Luo, Mi and Luo, Zhengyi and Meredith, Brighid and Miller, Austin and Oguntola, Oluwatumininu and Pan, Xiaqing and Peng, Penny and Pramanick, Shraman and Ramazanova, Merey and Ryan, Fiona and Shan, Wei and Somasundaram, Kiran and Song, Chenan and Southerland, Audrey and Tateno, Masatoshi and Wang, Huiyu and Wang, Yuchen and Yagi, Takuma and Yan, Mingfei and Yang, Xitong and Yu, Zecheng and Zha, Shengxin Cindy and Zhao, Chen and Zhao, Ziwei and Zhu, Zhifan and Zhuo, Jeff and Arbelaez, Pablo and Bertasius, Gedas and Damen, Dima and Engel, Jakob and Farinella, Giovanni Maria and Furnari, Antonino and Ghanem, Bernard and Hoffman, Judy and Jawahar, C.V. and Newcombe, Richard and Park, Hyun Soo and Rehg, James M. and Sato, Yoichi and Savva, Manolis and Shi, Jianbo and Shou, Mike Zheng and Wray, Michael},
    year          = 2024,
    booktitle     = {CVPR},
}

@inproceedings{wang2023videomaev2,
    title         = {VideoMAE V2: Scaling Video Masked Autoencoders With Dual Masking},
    author        = {Wang, Limin and Huang, Bingkun and Zhao, Zhiyu and Tong, Zhan and He, Yinan and Wang, Yi and Wang, Yali and Qiao, Yu},
    year          = 2023,
    booktitle     = {CVPR},
}
\bibliographystyle{iclr2027_conference}

\newpage
\appendix
\addtocontents{toc}{\protect\setcounter{tocdepth}{2}}
\crefalias{section}{appendix}
\crefalias{subsection}{appendix}

\section*{Supplementary Material}
This supplementary material is structured as follows:

\setcounter{tocdepth}{-1}\makeatletter\@starttoc{toc}\makeatother

\section{Extended Related Work}
\label{sup:extended_related}

This section complements \cref{sec:related} with a closer
comparison to prior and concurrent reconstruction and tracking methods,
providing additional details on their strengths and limitations.

\paragraph{Dense 4D prediction.}

Geometry transformers provide a starting point for 3D tracking through their
learned multi-view priors, but require adaptations to specify motion between observations.
Any4D~\citep{karhade2025any4d} builds on MapAnything to predict dense geometry
and scene flow but remains limited to motion relative to a single frame.
V-DPM~\citep{sucar2026vdpm} encodes a video once, then repeats time-conditioned
frame and global attention in its decoder for each requested target time.
4RC~\citep{luo20264rc} also reuses a shared video encoding, but conditions dense
decoding on a selected source--target pair. These designs differ in temporal
conditioning while retaining dense map outputs.

\paragraph{Query composition and conditioning.}
Moving to point-level outputs, Fast Spatial Tracking~\citep{huang2026fastspatial} initializes
queries from sampled multi-scale source features and exchanges information between global and frame-level
branches: the former attends across the video, while the latter attends within each frame.
UniQuery4R~\citep{chen2026uniquery4r} similarly samples multi-scale source features, and
implicitly conditions decoding on a selected source--target pair, similar to 4RC.
Feature selection avoids fixed-length temporal embeddings, but 3D outputs lack the flexibility
to express states in arbitrary viewpoints.
Instead, \method{} forms queries from Fourier-encoded image coordinates,
and content-derived source, target, and reference identities.
Our ablations support this design over replacing appearance and coordinate encodings with interpolated
encoder features (\cref{tab:ablation-paper-design}).

Long-range tracking adds the challenge of preserving point identity
across occlusions and large viewpoint changes.
Point4D~\citep{jeon2026point4d} addresses this by querying directly in 3D.
It propagates predicted 3D endpoints across video chunks, avoiding image
reprojection at handoffs and allowing queries during occlusion.
Its formulation nevertheless remains centered on monocular video: target-time tokens
are initialized from sinusoidal encodings of normalized frame positions, explicitly
injecting order before contextual refinement.
In this work we focus on retaining tracking performance within a single context.
The content-derived identities of \method{} accommodate variable context lengths,
while non-sequential training exposes the model to the large viewpoint changes
encountered in long-range tracking.
\Cref{fig:longtracking} shows that retaining a longer context complements
improved handoffs, for both 2D and 3D query handoffs.

\paragraph{Multi-view reconstruction and tracking.}
Beyond the scope of single-view tracking, multi-view capture offers a more comprehensive
understanding of the 3D scene~\citep[e.g.,][]{Grauman_2024_CVPR} and can improve
tracking by providing additional viewpoints that reduce occlusions.
Multi-view reconstructions also provide a stronger foundation for consistent
novel-view video generation~\citep{lee2026mvtrack4gen}.
One line of work assumes supplied camera calibration
and synchronized videos: MVTracker~\citep{rajic2025mvtracker} fuses features across views
into a shared 3D cloud using supplied or estimated depth, MV-TAP~\citep{koo2026mvtap} uses
camera geometry for multi-view 2D tracking, and LAPA~\citep{galoaa2026lapa}
combines per-view tracks and appearance through camera-guided attention
to recover 3D trajectories. MV-TAP and LAPA do not require depth inputs. A second line recovers
geometry and motion together.
OmniX~\citep{jiang2026omnix} couples its DA3 geometry backbone with dedicated
dynamic-token selection, transformation-basis prediction, and deformable sampling
to produce dense trajectories. This specialized motion representation entails
predicting a scene-wide trajectory field.

\section{Training Details}
\label{sup:training}

\subsection{Implementation Details}
\label{sup:implementation}

\paragraph{Architecture.}
We initialize the encoder with DA3-Giant weights.
Its final-layer features provide context
to an eight-layer pointwise decoder with hidden dimension $1280$ and $16$
attention heads, matching the decoder dimensions used by OpenD4RT.
Each query embeds a $9\times9$ source-centered RGB crop and Fourier features
of $2(u,v)-1\in[-1,1]^2$. For identity conditioning, we apply a shared LayerNorm
to each ID~token, concatenate them in source--target--reference order,
and project with a two-layer MLP with GELU.

\paragraph{Training inputs and augmentation.}
We sample sets of $2$--$24$ images and
adjust the number of sets per batch to accommodate at most $24$ images
per GPU. We sample an aspect ratio uniformly from $[0.75,3.0]$ and set the
longer image dimension to $504$ pixels.
The query budget is $1024$ per input image.
We apply various augmentations, including perspective warping, color jitter, JPEG compression, and blur induced by downsampling
and upsampling.

\paragraph{Optimization.}
We train for $200\mathrm{k}$ steps on $32$ NVIDIA GH200 GPUs using AdamW~\citep{loshchilov2019adamw}, with weight decay $0.03$ for
the backbone and $0.003$ for the decoder. Biases and
one-dimensional parameters are excluded from weight decay, and gradients
are clipped to a global norm of $20$.
The query modules use a peak learning rate of $\sqrt{2}\times10^{-4}$.
We apply layer-wise learning-rate decay~\citep{bao2022beit} to the backbone: for block
$\ell\in\{0,\ldots,39\}$, the peak learning rate is
\begin{equation}
    \eta_{\ell}^{\mathrm{peak}}
    =5\times10^{-5}\,0.96^{39-\ell}.
    \label{eq:backbone_layer_decay}
\end{equation}
This ranges from approximately $1\times10^{-5}$ in the first block to
$5\times10^{-5}$ in the final block, allowing larger updates in later layers.
The backbone is frozen for the first $4000$ steps, followed by $1000$ steps
of learning-rate warmup. Query modules warm up over the first $1000$ steps.
Both schedules subsequently follow cosine decay.
We maintain an exponential moving average of the model parameters,
with decay warmed toward $0.9999$ over $60\mathrm{k}$ updates.
To reduce activation memory, we enable gradient checkpointing except in the
backbone's global-attention blocks, where we retain activations to avoid
recomputing attention across the full patch set.

\paragraph{Loss formulation.}
Our loss formulation largely follows D4RT~\citep{zhang2026d4rt}, as detailed below.
The loss weights are $\lambda_{\mathrm{3D}}=1$,
$\lambda_{\mathrm{conf}}=0.2$, $\lambda_{\mathrm{uv}}=0.1$,
$\lambda_{\mathrm{vis}}=0.1$, $\lambda_{\mathrm{normal}}=0.5$, and
$\lambda_{\mathrm{disp}}=0.1$.
For each sample, let $\gV$ index valid point queries and let
$\mX=[\vx_i]_{i\in\gV}$ contain their targets in the respective reference
cameras. We normalize targets and predictions independently by their mean
Euclidean distance from their respective reference-camera origins and apply
an elementwise signed logarithm:
\begin{equation}
    \mu(\mX)=\frac{1}{|\gV|}\sum_{i\in\gV}\|\vx_i\|_2,
    \qquad
    \psi(\vx;\mX)=\sign(\vx)\odot
    \log\!\left(\vone+\frac{|\vx|}{\mu(\mX)}\right).
\end{equation}
Independent normalization removes global scale ambiguity, while the logarithm
reduces the influence of distant points. We use mean distance rather than mean depth
as queries may use arbitrary reference cameras.
The point and confidence terms in
\cref{eq:training_objective} are
\begin{equation}
    \Ls_{\mathrm{3D}} = \frac{1}{|\gV|}\sum_{i\in\gV}
    \hat c_i\|\psi(\hat{\vx}_i;\hat{\mX})-\psi(\vx_i;\mX)\|_1,
    \qquad
    \Ls_{\mathrm{conf}} = -\frac{1}{|\gV|}\sum_{i\in\gV}\log\hat c_i.
\end{equation}
For $\Ls_{\mathrm{uv}}$ and $\Ls_{\mathrm{vis}}$, we supervise the source point's
coordinates and visibility in the target image, respectively. Changing the
reference observation leaves both supervision targets unchanged.

For displacement, $\vd_i$ is the source-to-target world vector rotated into
the reference camera and divided by the scene scale, defined as the mean
distance of valid scene points from the first input camera. The head directly
predicts its radial logarithm, which compresses motion magnitude while
preserving direction:
\begin{equation}
    \rho(\vd)=\frac{\vd}{\|\vd\|_2}
    \log\!\left(1+\frac{\|\vd\|_2}{\epsilon}\right),
    \qquad \epsilon=10^{-4},
\end{equation}
with $\rho(\mathbf{0})=\mathbf{0}$. We average
$\|\hat{\vd}_i-\rho(\vd_i)\|_1$ over valid displacement targets to obtain
$\Ls_{\mathrm{disp}}$.

\subsection{Training Data}
\label{sup:samplingdata}

The mixture contains approximately 78K scenes and 22.8M camera frames.
We draw a dataset according to its mixture weight, then a scene uniformly
within its training index. The weights balance annotation coverage and scene
diversity, allocating $78\%$ to dynamic data and $22\%$ to static data.

\begin{table}[t]
    \centering
    \caption{\textbf{\method{} training mixture.}
\enquote{Multi-camera} denotes datasets with multiple video streams per scene.
    \enquote{Weight} denotes the probability to sample a scene from this dataset during training.
    }
    \label{tab:training-data}
    \scriptsize
    \setlength{\tabcolsep}{3pt}
    \renewcommand{\arraystretch}{1.2}
    \sisetup{mode=text,reset-text-series=false,group-separator={,},group-minimum-digits=4}
    \begin{tabularx}{\linewidth}{@{}>{\raggedright\arraybackslash}Xl*{3}{c}S[table-format=5]rS[table-format=3.2]@{}}
        \toprule
        \textbf{Dataset} & \textbf{Scene type} & \textbf{Tracks} & \begin{tabular}[c]{@{}c@{}}\textbf{Motion}\\\textbf{mask}\end{tabular} & \begin{tabular}[c]{@{}c@{}}\textbf{Multi-}\\\textbf{camera}\end{tabular} & {\textbf{\# Scenes}} & {\textbf{\# Frames}} & {\mbox{\textbf{Weight}}} \\
        \midrule
        Syn4D~\citep{jiang2026syn4d} & dynamic & \cmark & \cmark & \cmark & 504 & 1.2M & 13.50 \\
        Dynamic Replica~\citep{karaev2023dynamicstereo} & dynamic & \cmark & \cmark & \cmark & 503 & 151K & 12.50 \\
        Stereo4D~\citep{jin2025stereo4d} & dynamic & \cmark & \xmark & \xmark & 31420 & 6M & 11.50 \\
        PointOdyssey~\citep{zheng2023pointodyssey} & dynamic & \cmark & \cmark & \xmark & 123 & 238K & 10.50 \\
        Kubric-4D~\citep{vanhoorick2024gcd} & dynamic & \cmark & \cmark & \cmark & 2992 & 2.9M & 8.00 \\
        CoTracker3-Kubric~\citep{karaev2024cotracker3} & dynamic & \cmark & \cmark & \xmark & 5837 & 700K & 6.00 \\
        ParallelDomain-4D~\citep{vanhoorick2024gcd} & dynamic & \cmark & \xmark & \cmark & 1269 & 1.2M & 5.00 \\
        Waymo DriveTrack~\citep{sun2020waymo,balasingam2024drivetrack} & dynamic & \cmark & \xmark & \cmark & 664 & 658K & 5.00 \\
        Kubric MOVi-F~\citep{greff2022kubric} & dynamic & \cmark & \cmark & \xmark & 5883 & 141K & 4.00 \\
        nuScenes~\citep{caesar2020nuscenes} & dynamic & \xmark & \xmark & \cmark & 850 & 205K & 2.00 \\
MidAir~\citep{fonder2019midair} & static & \xmark & \xmark & \xmark & 138 & 61K & 1.75 \\
        GTA-SfM~\citep{wang2020gtasfm} & static & \xmark & \xmark & \xmark & 200 & 18K & 1.75 \\
        MVS-Synth~\citep{huang2018mvssynth} & static & \xmark & \xmark & \xmark & 120 & 12K & 1.50 \\
        WildRGB-D~\citep{xia2024wildrgbd} & static & \xmark & \xmark & \xmark & 22151 & 7.7M & 1.50 \\
        Aria Synthetic Environments~\citep{avetisyan2024scenescript} & static & \xmark & \xmark & \xmark & 1000 & 568K & 2.25 \\
        Hypersim~\citep{roberts2021hypersim} & static & \xmark & \xmark & \xmark & 411 & 67K & 4.00 \\
        ScanNet++~\citep{yeshwanthliu2023scannetpp} & static & \xmark & \xmark & \xmark & 902 & 897K & 3.50 \\
        BlendedMVS~\citep{yao2020blendedmvs} & static & \xmark & \xmark & \xmark & 493 & 113K & 3.25 \\
        Structured3D~\citep{zheng2020structured3d} & static & \xmark & \xmark & \xmark & 2653 & 63K & 2.50 \\
        \midrule
        \textbf{Total} &  &  &  &  & \bfseries 78113 & \bfseries 22.8M & \bfseries 100.00 \\
        \bottomrule
    \end{tabularx}
\end{table}

\begin{table}[t]
    \centering
    \caption{
        \textbf{Dataset-specific sampling.}
        We sample frames based on temporal locality, covisibility, or randomly from all available observations.
        \enquote{Mixed} indicates sampling from multiple camera streams.
        Where 3D tracks are available, we use them to construct a large percentage of all queries.
        The remaining query budget is divided into queries constructed from motion boundaries (B), dynamic regions (D), or based on geometric clues (G).
        Geometry queries are based on depth and normal discontinuities or uniformly sampled over the image plane.
}
    \label{tab:sampling-rules}
    \scriptsize
    \setlength{\tabcolsep}{3pt}
    \renewcommand{\arraystretch}{1.2}
    \sisetup{mode=text,reset-text-series=false}
    \begin{tabularx}{\linewidth}{@{}>{\raggedright\arraybackslash}XlcS[table-format=3]*{2}{S[table-format=2]}S[table-format=3]l@{}}
        \toprule
        \multirow{2}{*}[-0.5\dimexpr\aboverulesep+\cmidrulewidth+\belowrulesep\relax]{\textbf{Dataset}} & \multicolumn{2}{c}{\textbf{Image sampling}} & \multicolumn{5}{c}{\textbf{Query sampling}} \\
        \cmidrule(lr){2-3}\cmidrule(lr){4-8}
        & \textbf{Policy} & \textbf{Local window} & {\mbox{\textbf{Track (\%)}}} & {\mbox{\textbf{B (\%)}}} & {\mbox{\textbf{D (\%)}}} & {\mbox{\textbf{G (\%)}}} & \textbf{Geometry} \\
        \midrule
        Syn4D & Mixed & $\times10$ & 80 & 25 & 40 & 35 & Edges \\
        Dynamic Replica & Local & $\times10$ & 80 & 25 & 40 & 35 & Edges \\
        Stereo4D & Local & $\times4$ & 100 & {---} & {---} & {---} & --- \\
        PointOdyssey, standard & Local & $\times10$ & 70 & 25 & 40 & 35 & Edges \\
        \quad flat background & Local & $\times10$ & 80 & 35 & 55 & 10 & Uniform \\
        \quad moving background & Local & $\times10$ & 80 & 35 & 55 & 10 & Uniform \\
        Kubric-4D & Mixed & $\times5$ & 80 & 35 & 55 & 10 & Uniform \\
        CoTracker3-Kubric & Local & $\times4$ & 80 & 35 & 55 & 10 & Uniform \\
        ParallelDomain-4D & Mixed & $\times5$ & 60 & 0 & 0 & 100 & Edges \\
        Waymo DriveTrack & Single camera & $\times3$ & 70 & 0 & 0 & 100 & Uniform \\
        Kubric MOVi-F & Random & --- & 80 & 35 & 55 & 10 & Uniform \\
        nuScenes & Single camera & $\times3$ & 0 & 0 & 0 & 100 & Edges \\
MidAir & Covisibility & --- & 0 & 0 & 0 & 100 & Edges \\
        GTA-SfM & Covisibility & --- & 0 & 0 & 0 & 100 & Edges \\
        MVS-Synth & Covisibility & --- & 0 & 0 & 0 & 100 & Edges \\
        WildRGB-D & Random & --- & 0 & 0 & 0 & 100 & Edges \\
        Hypersim & Covisibility & --- & 0 & 0 & 0 & 100 & Edges \\
        ScanNet++ & Covisibility & --- & 0 & 0 & 0 & 100 & Edges \\
        BlendedMVS & Covisibility & --- & 0 & 0 & 0 & 100 & Edges \\
        Structured3D & Covisibility & --- & 0 & 0 & 0 & 100 & Edges \\
        Aria Synthetic Environments & Local & $\times2$ & 0 & 0 & 0 & 100 & Edges \\
        \bottomrule
    \end{tabularx}
\end{table}

\FloatBarrier

\subsection{Sampling Strategy}
\label{sup:sampling}

The relational query interface separates the observations that provide context
from the locations and annotations used for supervision. We first select an
image set, then sample pixels or track identities, assign the roles in
\cref{eq:query}, and construct the available targets. This lets static
multi-view geometry, dynamic trajectories, and scenes without tracks
contribute to one training objective.

\paragraph{Image selection.}
For a set of $S\in\{2,\ldots,24\}$ images, local sampling selects observations
without replacement from a contiguous candidate window of up to $kS$ frames.
\Cref{tab:sampling-rules} reports the maximum local-window multiplier:
$\times10$, for example, draws integer $k$ uniformly from $1$ to $10$,
subject to sequence length. Multi-view datasets
mix single-camera sets with observations drawn across cameras in a temporal
window; the resulting sets need not contain every camera at every timestep.
Static datasets use covisibility where available, with minimum overlap $0.2$,
or local/random sampling as specified in the table.

\begin{figure}[t]
    \centering
    \begingroup
    \setlength{\tabcolsep}{0pt}
    \newcommand{\samplingimage}[1]{\includegraphics[width=0.32\linewidth]{#1}}
    \begin{tabular}{@{}c@{\hspace{0.02\linewidth}}c@{\hspace{0.02\linewidth}}c@{}}
        \small RGB & \small Derived edges & \small Queries \\[3pt]
        \samplingimage{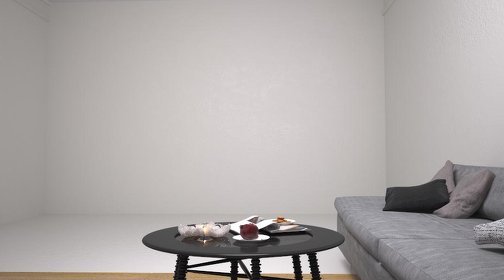}& \samplingimage{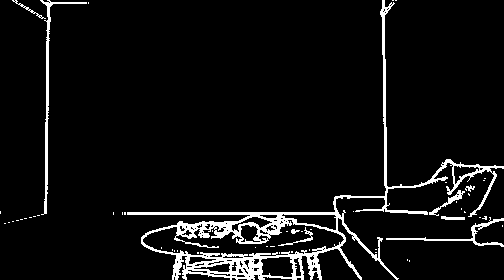}& \samplingimage{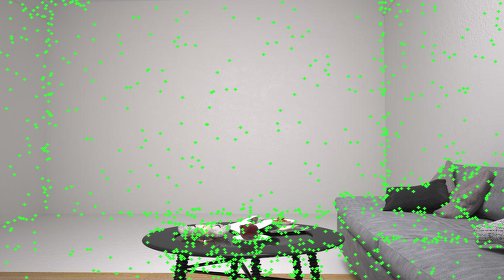}\\\noalign{\vskip4pt}
        \samplingimage{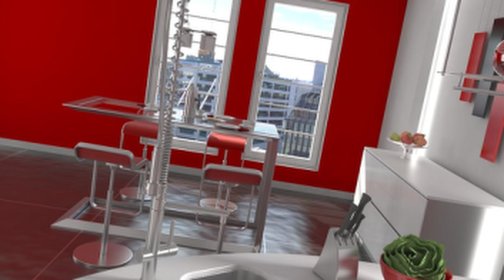}& \samplingimage{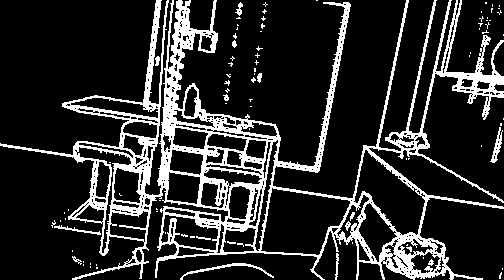}& \samplingimage{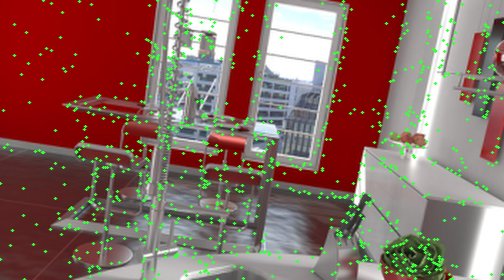}\end{tabular}
    \endgroup
    \caption{
        \textbf{Geometric query sampling.}
        On static scenes, we oversample queries near depth and normal discontinuities.
        From left to right: the RGB of a sample frame; the detected edges; the sampled source $(u,v)$ coordinates.
        The fraction of queries constructed this way varies by dataset (see~\cref{tab:sampling-rules}).
    }
    \label{fig:sampling-geometry}
\end{figure}

\paragraph{Depth and normal edges.}
Beyond a correct global layout, faithful 3D reconstruction requires accurate
local geometry, including sharp surface transitions and thin structures \citep{knaebel2026surge,kong2026finedetailmonoculargeometryestimation}.
These regions are notoriously difficult to recover, yet they cover only a few pixels and
thus receive few queries under uniform sampling. We therefore oversample
queries near depth and normal discontinuities (\cref{fig:sampling-geometry}).
We combine relative depth jumps above $5\%$ in a $3\times3$ neighborhood with
normal changes above $25^\circ$. Sampling normals are estimated from depth on
a normalized view plane, independently of the normal annotations used by the
loss. The normal detector compares four-connected neighbors, rejects
grazing-angle estimates above $80^\circ$, and requires changes to persist over
a three-pixel radius. For the combined edge set $\gP_{\mathrm{edge}}$ and
eligible pixel locations $\vp$,
we construct distance-based weights
\begin{equation}
    w(\vp)=1+10\exp\!\left(-d(\vp,\gP_{\mathrm{edge}})/2\right),
    \label{eq:sampling_weights}
\end{equation}
where $d$ is measured in pixels. Up to $30\%$ of the geometry budget is sampled
uniformly within three pixels of an edge; the remainder is sampled without
replacement from the unselected pixels with probability proportional to $w$.
The positive base weight preserves coverage away from boundaries, and sampling
becomes uniform when no edges are found.

Normal edges expose changes in surface orientation that need not produce a
large depth jump. Qualitatively, omitting them led to more rounded
reconstructions at sharp transitions, consistent with insufficient supervision
of the abrupt surface change.
As shown in \cref{fig:qual_reconstruction}, \method{} accurately recovers such structures. 
We attribute this capability to the pixel-level decoding and the dedicated edge supervision.

\begin{figure}[t]
    \centering
    \begin{subfigure}{\linewidth}
    \phantomsubcaption\label{fig:sampling-motion}
    \noindent\begin{minipage}[c]{0.025\linewidth}
        \centering\footnotesize (\thesubfigure)
    \end{minipage}\hspace{0.015\linewidth}\begin{minipage}[c]{0.96\linewidth}
    \begingroup
    \setlength{\tabcolsep}{0pt}
    \begin{tabular}{@{}c@{\hspace{0.02\linewidth}}c@{\hspace{0.02\linewidth}}c@{}}
        \multicolumn{3}{c}{\small Queries oversampled near motion boundaries} \\[3pt]
\includegraphics[width=0.32\linewidth,trim=0 14bp 0 0,clip]{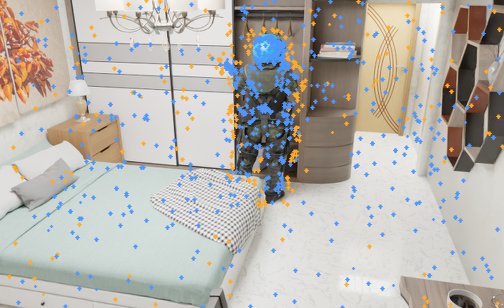}& \includegraphics[width=0.32\linewidth]{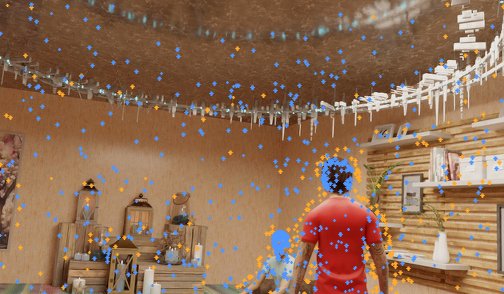}& \includegraphics[width=0.32\linewidth]{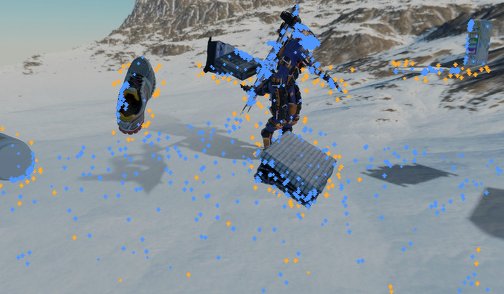}\end{tabular}
    \endgroup
    \end{minipage}
    \end{subfigure}
    \par\vspace{12pt}
    \begin{subfigure}{\linewidth}
    \phantomsubcaption\label{fig:sampling-static-side}
    \noindent\begin{minipage}[c]{0.025\linewidth}
        \centering\footnotesize (\thesubfigure)
    \end{minipage}\hspace{0.015\linewidth}\begin{minipage}[c]{0.96\linewidth}
    \begingroup
    \comparisonstyle
    \newlength{\samplingpredw}
    \setlength{\samplingpredw}{\dimexpr(\linewidth-0.13\linewidth-0.02\linewidth-\comparisonsep)/2\relax}
    \begin{tabular}{@{}Ic@{\hspace{0.02\linewidth}}c@{}}
        \comparisoninputheading & \small Without static boundary supervision & \small With static boundary supervision \\[3pt]
    \begin{minipage}[c]{0.13\linewidth}
        \offinterlineskip
        \hbox{\includegraphics[width=\linewidth]{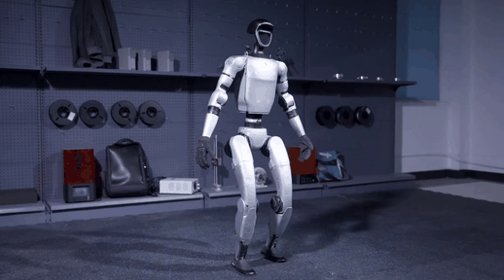}}\kern2pt
        \hbox{\includegraphics[width=\linewidth]{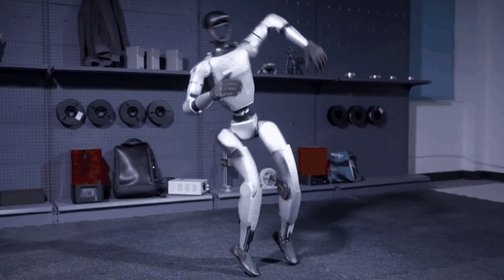}}\kern2pt
        \hbox{\includegraphics[width=\linewidth]{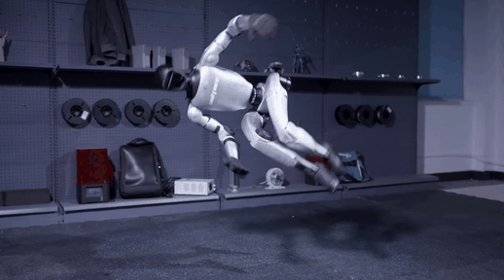}}\kern4pt
        \hbox to \linewidth{\hfil\ldots\hfil}
    \end{minipage} &
    \begin{minipage}[c]{\samplingpredw}
        \centering
        \includegraphics[width=\linewidth]{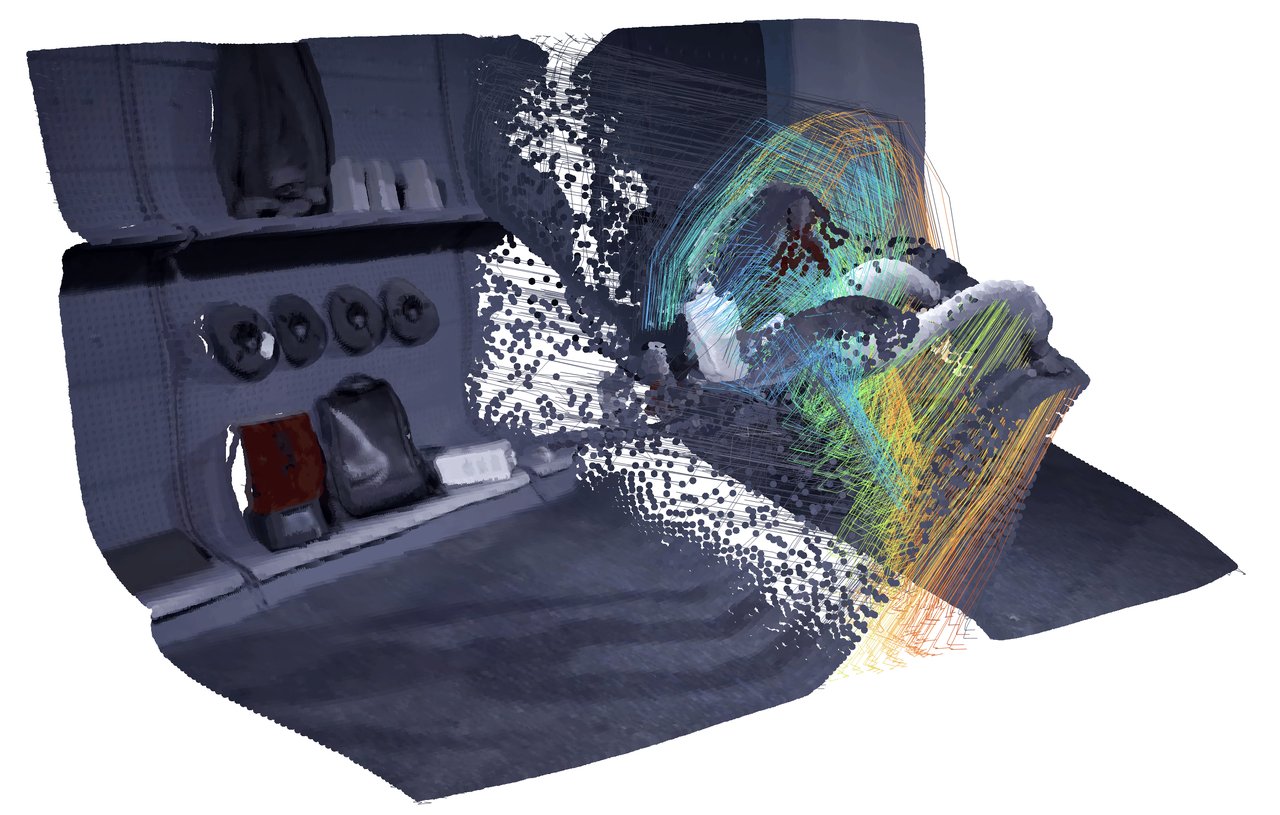}
    \end{minipage} &
    \begin{minipage}[c]{\samplingpredw}
        \centering
        \includegraphics[width=\linewidth]{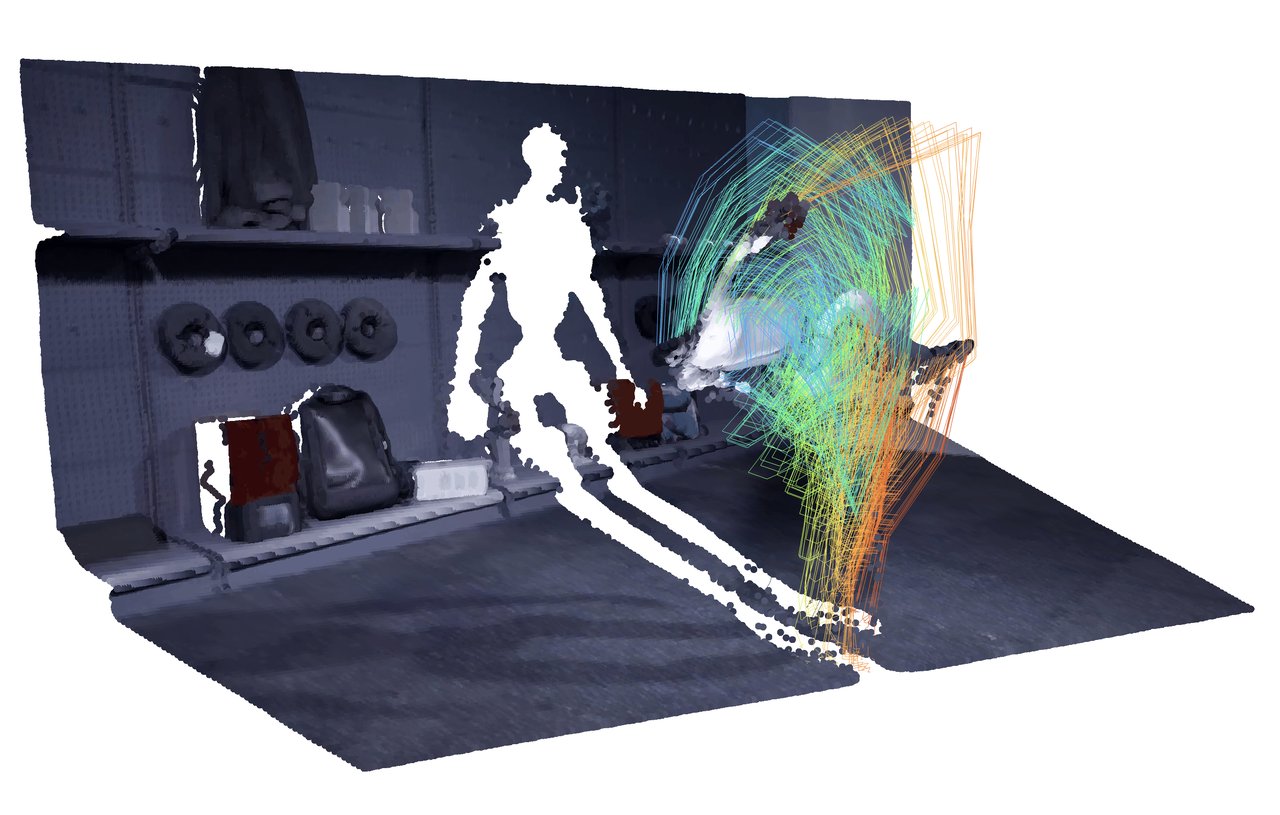}
    \end{minipage}
    \end{tabular}
    \endgroup
    \end{minipage}
    \end{subfigure}
    \caption{
        \textbf{Motion boundary oversampling.}
        (a) Oversampling near motion boundaries encourages the model to distinguish static and moving points. Blue marks indicate queries sampled from 3D tracks, and orange points indicate queries oversampled near motion boundaries.
        (b) Qualitative comparison of the impact of oversampled motion boundaries. Left: RGB input; middle: tracks without motion boundary oversampling; right: tracks with motion boundary oversampling.
The fraction of queries constructed this way varies by dataset (see~\cref{tab:sampling-rules}).
    }
    \label{fig:sampling-motion-boundaries}
\end{figure}

\paragraph{Motion and boundary allocation.}
After reserving a dataset-specific track share, we divide ordinary pixel
queries among static-side motion boundaries, dynamic regions, and background
geometry. The boundary partition covers a nine-pixel band, with Gaussian
weights centered $4.5$ pixels into the static side and standard deviation
$2$ pixels. Dynamic sampling covers the object interior with additional
silhouette weight. Background geometry uses the edge distribution above or
uniform sampling for flat-background profiles. Sampling the static side is
motivated by a qualitative failure mode: supervising predominantly the moving
side can allow the nearby static background to deform with it.
See \cref{fig:sampling-motion-boundaries} for a visualization of boundary sampling and the described failure mode.

\paragraph{Frame roles and valid supervision.}
Static scenes allocate $40\%$ of queries to
$i_{\mathrm{src}}=i_{\mathrm{tgt}}=i_{\mathrm{ref}}$; the rest sample a target
observation and a different reference observation where available.
Dynamic-scene queries use $i_{\mathrm{ref}}=i_{\mathrm{tgt}}$ for $40\%$ of
queries and another valid reference for the rest. This coordinate choice is
independent of whether the point changes state.

Track queries use valid, visible source states and valid target states,
including occluded ones. Ordinary dynamic-region queries remain same-time,
$i_{\mathrm{tgt}}=i_{\mathrm{src}}$. Background queries can connect observations
when their static status is trusted. All ordinary queries remain same-time
for nuScenes, ParallelDomain-4D, and PointOdyssey's moving-background profile;
their available tracks retain correspondence-based targets. Waymo additionally
requests cross-time queries on semantic classes identified as static.
\Cref{tab:sampling-rules} reports requested allocations: candidate
shortages can increase the geometry share, and remaining non-track shortfalls
can use unselected dynamic pixels.

\section{Evaluation Details}
\label{sup:samplingevaluation}

\subsection{Evaluation Protocols}
\label{sup:protocols}

\paragraph{WorldTrack benchmark details.}
\label{sup:worldtrack}
Following WorldTrack~\citep{feng2025st4rtrack}, we initialize queries at
visible positions in the first frame and evaluate trajectories over the first
$64$ frames. We use the global median scale alignment protocol.
Ground-truth trajectories are expressed in the first camera's coordinate system,
and predictions are multiplied by
\begin{equation}
    s=\frac{\operatorname{median}_{i\in\gV}\|\vx_i^{\mathrm{gt}}\|_2}
    {\operatorname{median}_{i\in\gV}\|\hat{\vx}_i\|_2},
    \label{eq:worldtrack_scale}
\end{equation}
where $\gV$ indexes valid track--time correspondences with finite ground truth
and predictions. This fits a single scale without adjusting rotation or
translation. APD averages the percentage of valid positions with Euclidean
error at most $\{0.1,0.3,0.5,1.0\}$ meters; EPE is the mean Euclidean error.
Dynamic-only evaluation retains tracks whose cumulative displacement over
consecutive valid ground-truth positions exceeds $1$\,cm, and estimates scale
separately on this subset. Results are averaged across sequences.

\paragraph{Multi-view benchmark adaptation.}
\label{sup:samplingmvtracker}
We provide the sampling and scoring details for the RGB-only adaptation in
\cref{sec:multiview_tracking}. For Multi-View Kubric and DexYCB we retain the
$24$ timesteps and the released query subsets.
For Panoptic Studio, we subsample every second frame from the first $48$ frames
($0,2,\ldots,46$) for each of the camera-rig configurations.
To maintain sufficient query coverage in this shorter clip, we reapply
MVTracker's query-sampling procedure within the retained window.
The source view is sampled uniformly among views in which
the query is visible; ground-truth calibration is used only to project the
query into that image and construct evaluation targets.

\paragraph{Multi-view coordinates and scoring.}
Each query is predicted at all retained timesteps, both before and after its
source time. We score one 3D position per physical timestep in the first
input camera's coordinate system, taking the first camera's target slot.
Ground-truth visibility is the union over rig cameras, using physical
visibility rather than the causal mask used for query selection; predicted
visibility likewise uses the maximum over views.
APD and EPE evaluate valid, ground-truth-visible positions,
whereas OA measures visibility agreement over all valid positions, including occluded states.

We fit one scale per scene and rig using \cref{eq:worldtrack_scale} over
valid, visible correspondences with finite predictions.
For \cref{tab:mvtrack-paper-ready}, Kubric EPE is converted to meters
using approximately $0.13$\,m per native unit.
We pool track--time pairs within each scene, average scenes and then Panoptic Studio rigs equally, and weight the three
datasets equally in the overall average.
These protocol changes, including the different input assumptions, preclude
direct numerical comparison with the official MVTracker results.

\subsection{Baseline adaptation for multi-view tracking}
\label{sup:baselines}

\paragraph{Trackers with external geometry.}
For the multi-view comparison, we replace ground-truth depth and calibration
with predicted geometry and lift pixel queries using the predicted source-view
depth. MVTracker uses VGGT or DA3 reconstruction independently at each
timestep, or VGGT-$\Omega$ reconstruction jointly over the clip.
TAPIP3D~\citep{zhang2025tapip3d} tracks each query in its source-view video
using joint VGGT-$\Omega$ geometry, while LAPA~\citep{galoaa2026lapa} uses
this geometry with CoTracker3 observations.
MVTracker and LAPA combine forward and reversed-time passes at each query's
source time; TAPIP3D uses its bidirectional inference mode.
MVTracker uses a $4\times4$ global support grid per view at the start of each
pass, and TAPIP3D uses a $40\times40$ support grid.
The VGGT-$\Omega$ comparisons use the $416$-resolution checkpoint released in August 2026.

\paragraph{Joint geometry and tracking baselines.}
For 4RC, we place the reference image first, sample queries from the dense
track field of their source image, and restore the original input order.
OmniX~\citep{jiang2026omnix} receives camera-grouped videos with synchronized
timestamps. Given OmniX's high memory requirements, we batch trajectory-head
and deformable-attention computations to avoid GPU memory exhaustion while
preserving full temporal context. Both methods are scored using the
same adapted protocol; AJ and OA are omitted because neither provides a
native visibility prediction.

\subsection{Camera Pose Recovery}
\label{sup:camera_recovery}

Following D4RT~\citep{zhang2026d4rt}, \cref{code:camera_recovery}
recovers the pose of each camera $i$ relative to reference camera $j$.
The point matrices $\mX^{(i)}$ and $\mX^{(j)}$, with columns
$\vx_k^{(i)}$ and $\vx_k^{(j)}$, describe the same physical points at
source state $i$, expressed in cameras $i$ and $j$; moving objects therefore
also provide valid correspondences. Both query sets use the same enumeration
$\gG_i=\{(u_k,v_k)\}_{k=1}^{M}$ of the source grid; decoding preserves $k$ in
the point-matrix columns and confidence and mask entries.
We choose the middle input as reference
and assign it the identity pose. The decoding step constructs validity masks
$\vm^{(i)},\vm^{(j)}$ that retain
finite 3D predictions with positive finite confidence, in-bounds predicted UV,
and positive visibility logits. The final SVD solve fits a proper rotation
and translation without rescaling. We omit prediction hats in this algorithm;
$\mR_{i\to j}$ is a rotation matrix and $\vt_{i\to j}$ is a translation vector.

\begin{algorithm}[t]
    \caption{Derivation of camera pose $i\to j$.}
    \label{code:camera_recovery}
    \small
    \begin{algorithmic}[1]
        \Require Encoded context $\mF$, images $I_i$, $I_j$.
        \Statex \textcolor{black!60}{\it\# Query construction and decoding}
        \State $\gG_i \gets \operatorname{UniformGrid}(I_i,64\times64)$ \Comment{Grid for source frame $i$}
        \State $\gQ_i \gets \{\vq_{i,k}\}_{k=1}^{M},\quad \vq_{i,k}=(u_k,v_k,i,i,i)$ \Comment{Coordinates in camera $i$}
        \State $\gQ_j \gets \{\vq_{j,k}\}_{k=1}^{M},\quad \vq_{j,k}=(u_k,v_k,i,i,j)$ \Comment{Coordinates in camera $j$}
        \State $(\mX^{(s)},\vc^{(s)},\vm^{(s)}) \gets \operatorname{Decode}(\gQ_s,\mF),\quad s\in\{i,j\}$
        \Statex
        \Statex \textcolor{black!60}{\it\# Confidence-based selection and weighting}
        \State $\vc \gets \vc^{(i)} \odot \vc^{(j)}$ \Comment{Joint confidence}
        \State $\gK \gets \{k\in\{1,\ldots,M\}\mid \evm_k^{(i)}=\evm_k^{(j)}=1\}$ \Comment{Valid in both frames}
        \State $\gK \gets \{k \in \gK \mid \evc_k \geq \operatorname{median}(\vc_{\gK})\}$ \Comment{Keep the most confident half}
\State $\evw_k \gets \evc_k^2/\sum_{\ell\in\gK}\evc_\ell^2,\quad k\in\gK$ \Comment{Normalize squared confidence}
        \Statex
        \Statex \textcolor{black!60}{\it\# Solve via SVD: align camera $i$ to camera $j$}
        \State $(\mR_{i\to j},\vt_{i\to j}) \gets \argmin_{\mR\in\mathrm{SO}(3),\;\vt\in\R^3}\;\sum_{k\in\gK}\evw_k\left\|\vx_k^{(j)}-\left(\mR\vx_k^{(i)}+\vt\right)\right\|_2^2$ \Comment{Solve by SVD}
        \State \Return $(\mR_{i\to j},\vt_{i\to j})$ \Comment{$\vx^{(j)}=\mR_{i\to j}\vx^{(i)}+\vt_{i\to j}$}
    \end{algorithmic}
\end{algorithm}

\section{Additional Results}
\label{sup:additional_results}

\subsection{Further Ablations}
\label{sup:additional_ablations}
\paragraph{Design ablations.}
We evaluate design alternatives on $64$-frame WorldTrack sequences using
the final configuration in \cref{tab:ablation-paper-main} as the baseline.
Each variant follows the same ablation training protocol, except for the indicated
change.
\Cref{tab:ablation-paper-design} reports absolute baseline scores and
differences for each variant, for all-point tracking.
Freezing the encoder and training only the decoder reduces mean APD by
$8.71$ percentage points overall.
This illustrates that while the encoder representation can provide a
good baseline for tracking, fine-tuning the encoder is important for optimal performance.

Multi-level decoding uses DA3-Large layers $11,15,19,23$, matching the
feature levels of its DPT head. Successive decoder layers attend to these
levels and then reuse the deepest level. Gains are small on average and
vary across datasets; we therefore retain a shared final-layer context
for simplicity and memory efficiency.
We also evaluate a common alternative for query initialization: bilinearly
sampling encoder patch features at the queried UV coordinates, replacing
the RGB crop and Fourier-coordinate encoding \citep{chen2026uniquery4r,huang2026fastspatial,yu2026infinidepth}. Although this variant
converges faster initially, it yields lower final tracking accuracy and
predictions with a more patch-like spatial structure. These observations
favor initializing the query's appearance and spatial components
independently of encoder patch features.
Finally, we maintain an exponential moving average (EMA) of the model
parameters. EMA slightly improves average all-point APD and EPE,
so we retain it for the final model.

\begin{table}[t]
    \centering
    \scriptsize
    \setlength{\tabcolsep}{2pt}
    \renewcommand{\arraystretch}{1.2}
    \sisetup{mode=text,reset-text-series=false,retain-explicit-plus=true}
    \caption{\textbf{Design-choice ablations on WorldTrack.}
        All rows use the $100\mathrm{k}$-step DA3-Large ablation setup of \cref{tab:ablation-paper-main}.
        Entries report APD (\%) and EPE (m) over all points.
S5 reports absolute scores; other rows report signed differences from S5.
        APD differences are in percentage points.
        The first three variants are trained separately with the indicated change.
        The EMA row evaluates exponential-moving-average weights from the S5 training run.
    }
    \label{tab:ablation-paper-design}
    \begin{tabularx}{\linewidth}{@{}>{\raggedright\arraybackslash}X*{5}{S[table-format=-2.2]S[table-format=+1.4]}@{}}
        \toprule
        \multirow{2}{*}[-0.5\dimexpr\aboverulesep+\cmidrulewidth+\belowrulesep\relax]{\textbf{Configuration}} & \multicolumn{2}{c}{\textbf{Aria Digital Twin}} & \multicolumn{2}{c}{\textbf{Dynamic Replica}} & \multicolumn{2}{c}{\textbf{PointOdyssey}} & \multicolumn{2}{c}{\textbf{Panoptic Studio}} & \multicolumn{2}{c}{\textbf{Avg.}} \\
        \cmidrule(lr){2-3}
        \cmidrule(lr){4-5}
        \cmidrule(lr){6-7}
        \cmidrule(lr){8-9}
        \cmidrule(lr){10-11}
        & {\mbox{APD \ensuremath{\uparrow}}} & {\mbox{EPE \ensuremath{\downarrow}}} & {\mbox{APD \ensuremath{\uparrow}}} & {\mbox{EPE \ensuremath{\downarrow}}} & {\mbox{APD \ensuremath{\uparrow}}} & {\mbox{EPE \ensuremath{\downarrow}}} & {\mbox{APD \ensuremath{\uparrow}}} & {\mbox{EPE \ensuremath{\downarrow}}} & {\mbox{APD \ensuremath{\uparrow}}} & {\mbox{EPE \ensuremath{\downarrow}}} \\
        \midrule
        Frozen encoder & \cellcolor[HTML]{f0bfc4}-6.85 & \cellcolor[HTML]{f0bfc4}+0.0612 & \cellcolor[HTML]{f0bfc4}-8.20 & \cellcolor[HTML]{f0bfc4}+0.0706 & \cellcolor[HTML]{f0bfc4}-4.39 & \cellcolor[HTML]{f0bfc4}+0.0851 & \cellcolor[HTML]{f0bfc4}-15.41 & \cellcolor[HTML]{f0bfc4}+0.1487 & \cellcolor[HTML]{f0bfc4}-8.71 & \cellcolor[HTML]{f0bfc4}+0.0914 \\
        Multi-level context & \cellcolor[HTML]{f5f9e8}\rankunderline{+0.33} & \cellcolor[HTML]{f9fcea}\rankunderline{-0.0013} & \cellcolor[HTML]{fdf7e8}-0.33 & \cellcolor[HTML]{f8fbea}\bfseries -0.0018 & \cellcolor[HTML]{fdf6e7}-0.21 & \cellcolor[HTML]{fdf6e7}+0.0042 & \cellcolor[HTML]{f1f7e5}\bfseries +1.32 & \cellcolor[HTML]{f4f8e7}\bfseries -0.0088 & \cellcolor[HTML]{f7fbe9}\rankunderline{+0.28} & \cellcolor[HTML]{f9fcea}\bfseries -0.0019 \\
        Patch-feature query initialization & \cellcolor[HTML]{fcf2e5}-0.59 & \cellcolor[HTML]{fbefe3}+0.0072 & \cellcolor[HTML]{f8e1da}-2.59 & \cellcolor[HTML]{f6d9d5}+0.0311 & \cellcolor[HTML]{fcf3e5}-0.33 & \cellcolor[HTML]{f8e1da}+0.0270 & \cellcolor[HTML]{fbeee2}-2.06 & \cellcolor[HTML]{fbf0e3}+0.0166 & \cellcolor[HTML]{faece1}-1.39 & \cellcolor[HTML]{f9e7dd}+0.0205 \\
        \textbf{Stage~S5 (\cref{tab:ablation-paper-main})} & 87.52 & 0.1495 & \bfseries 81.01 & \rankunderline{0.2026} & \rankunderline{74.36} & \rankunderline{0.2519} & \rankunderline{77.13} & \rankunderline{0.1894} & 80.01 & 0.1983 \\
        S5 + EMA & \cellcolor[HTML]{e6f1e0}\bfseries +1.37 & \cellcolor[HTML]{e8f2e1}\bfseries -0.0110 & \cellcolor[HTML]{fef9e9}\rankunderline{-0.23} & \cellcolor[HTML]{fcf4e6}+0.0046 & \cellcolor[HTML]{f4f8e7}\bfseries +0.26 & \cellcolor[HTML]{f4f9e7}\bfseries -0.0049 & \cellcolor[HTML]{fefdec}-0.09 & \cellcolor[HTML]{fdf7e8}+0.0060 & \cellcolor[HTML]{f7fae8}\bfseries +0.33 & \cellcolor[HTML]{fbfceb}\rankunderline{-0.0013} \\
        \bottomrule
    \end{tabularx}
\end{table}

\paragraph{Short-context tracking.}
To assess whether our more general formulation incurs a cost in
 short-context tracking accuracy, we evaluate all cumulative stages on
 $24$-frame WorldTrack sequences, keeping within the $2$--$24$-image training range
 (\cref{tab:ablation-paper-short-sequence}). All variants therefore use
 full-context inference. Stage~S5 achieves the best average all-point APD and EPE,
 although earlier stages lead on
 some individual datasets. These results demonstrate that the final design's
 broader input support comes without an average accuracy penalty at context
 lengths seen during training.

\begin{table}[t]
    \renewcommand{\cmark}{\textcolor{black!50}{\ding{51}}}
    \newcommand{\dartquery}{\raisebox{-0.25em}{\includegraphics[height=1em]{fig/query_icons/d4rt-dart.pdf}}}
    \newcommand{\arrowquery}{\raisebox{-0.25em}{\includegraphics[height=1em]{fig/query_icons/wooden-arrow-6.pdf}}}
    \centering
    \tiny
    \setlength{\tabcolsep}{3pt}
    \renewcommand{\arraystretch}{1.2}
    \sisetup{mode=text,reset-text-series=false,retain-explicit-plus=true,table-format=-2.2}
    \renewcommand{\tabularxcolumn}[1]{S[table-column-width=#1]}
    \setlength{\bigstrutjot}{\dimexpr\aboverulesep+\cmidrulewidth+\belowrulesep\relax}
    \newcommand{\ablationheader}[1]{\multirow{3}[b1]{*}{\rotatebox{90}{\bfseries\shortstack{#1}}}}
    \caption{\textbf{Short-context ablations on WorldTrack.}
        We use $24$-frame full-context inference for all stages.
        Stages use the $100\mathrm{k}$-step DA3-Large setup of \cref{tab:ablation-paper-main}.
        \dartquery{}: order-dependent querying (D4RT); \arrowquery{}: order-invariant querying (\method{}).
        Entries report APD (\%) and EPE (m) over all points.
S5 reports absolute scores; other rows report signed differences from S5.
        APD differences are in percentage points.
    }
    \label{tab:ablation-paper-short-sequence}
    \begin{tabularx}{\linewidth}{@{}*{6}{c}*{5}{X>{\sisetup{table-format=+1.4}}X}@{}}
        \toprule
        \ablationheader{Stage} & \ablationheader{Query\\Approach} & \ablationheader{PE} & \ablationheader{Sequential\\Sampling} & \ablationheader{Data} & \ablationheader{Eval} & \multicolumn{10}{c}{} \\
        & & & & & & \multicolumn{2}{c}{\textbf{Aria Digital Twin}} & \multicolumn{2}{c}{\textbf{Dynamic Replica}} & \multicolumn{2}{c}{\textbf{PointOdyssey}} & \multicolumn{2}{c}{\textbf{Panoptic Studio}} & \multicolumn{2}{c}{\textbf{Avg.}} \\
        \cmidrule(lr){7-8}
        \cmidrule(lr){9-10}
        \cmidrule(lr){11-12}
        \cmidrule(lr){13-14}
        \cmidrule(lr){15-16}
        & & & & & & {\mbox{APD \ensuremath{\uparrow}}} & {\mbox{EPE \ensuremath{\downarrow}}} & {\mbox{APD \ensuremath{\uparrow}}} & {\mbox{EPE \ensuremath{\downarrow}}} & {\mbox{APD \ensuremath{\uparrow}}} & {\mbox{EPE \ensuremath{\downarrow}}} & {\mbox{APD \ensuremath{\uparrow}}} & {\mbox{EPE \ensuremath{\downarrow}}} & {\mbox{APD \ensuremath{\uparrow}}} & {\mbox{EPE \ensuremath{\downarrow}}} \\
        \midrule
        S1 & \dartquery & \cmark & \cmark & video & full & \cellcolor[HTML]{f0bfc4}-9.09 & \cellcolor[HTML]{f0bfc4}+0.0795 & \cellcolor[HTML]{fbfdeb}\rankunderline{+0.04} & \cellcolor[HTML]{fcfdeb}\rankunderline{-0.0001} & \cellcolor[HTML]{fdf6e7}-0.19 & \cellcolor[HTML]{f9e4dc}+0.0043 & \cellcolor[HTML]{f6dad5}\rankunderline{-2.50} & \cellcolor[HTML]{f7ddd7}\rankunderline{+0.0178} & \cellcolor[HTML]{f3cbcb}-2.94 & \cellcolor[HTML]{f0bfc4}+0.0254 \\
        S2 & \arrowquery & \cmark & \cmark & video & full & \cellcolor[HTML]{f6d8d4}-4.29 & \cellcolor[HTML]{f6d9d5}+0.0351 & \cellcolor[HTML]{dcebda}\bfseries +1.31 & \cellcolor[HTML]{b8d6c7}\bfseries -0.0113 & \cellcolor[HTML]{dbeada}\bfseries +1.27 & \cellcolor[HTML]{d3e6d5}\rankunderline{-0.0078} & \cellcolor[HTML]{f0c0c5}-5.51 & \cellcolor[HTML]{f0bfc4}+0.0469 & \cellcolor[HTML]{f6d9d5}-1.81 & \cellcolor[HTML]{f4d0cf}+0.0157 \\
        S3 & \arrowquery & \xmark & \cmark & video & full & \cellcolor[HTML]{f9e5dc}\rankunderline{-2.30} & \cellcolor[HTML]{f9e7de}\rankunderline{+0.0174} & \cellcolor[HTML]{fcfdeb}+0.03 & \cellcolor[HTML]{f8e1da}+0.0035 & \cellcolor[HTML]{e2eedd}\rankunderline{+0.95} & \cellcolor[HTML]{cde2d2}\bfseries -0.0096 & \cellcolor[HTML]{f0bfc4}-5.71 & \cellcolor[HTML]{f0c0c5}+0.0455 & \cellcolor[HTML]{f6dad5}\rankunderline{-1.75} & \cellcolor[HTML]{f5d3d1}\rankunderline{+0.0142} \\
        S4 & \arrowquery & \xmark & \xmark & video & full & \cellcolor[HTML]{f7ddd7}-3.46 & \cellcolor[HTML]{f8dfd9}+0.0272 & \cellcolor[HTML]{fbece1}-0.61 & \cellcolor[HTML]{f4d1cf}+0.0069 & \cellcolor[HTML]{f4d1cf}-2.24 & \cellcolor[HTML]{f0bfc4}+0.0163 & \cellcolor[HTML]{f4cfce}-3.65 & \cellcolor[HTML]{f3cdcd}+0.0316 & \cellcolor[HTML]{f4d0cf}-2.49 & \cellcolor[HTML]{f2c7c9}+0.0205 \\
        \midrule
        \textbf{S5} & \arrowquery & \xmark & \xmark & all & full & \bfseries 87.19 & \bfseries 0.1560 & 79.14 & 0.2211 & 73.68 & 0.2599 & \bfseries 80.10 & \bfseries 0.1668 & \bfseries 80.03 & \bfseries 0.2010 \\
        \bottomrule
    \end{tabularx}
\end{table}

\subsection{Long-Range Tracking}
\label{sup:longtracking}

\begin{figure}[t]
    \centering
    \includegraphics[width=\linewidth]{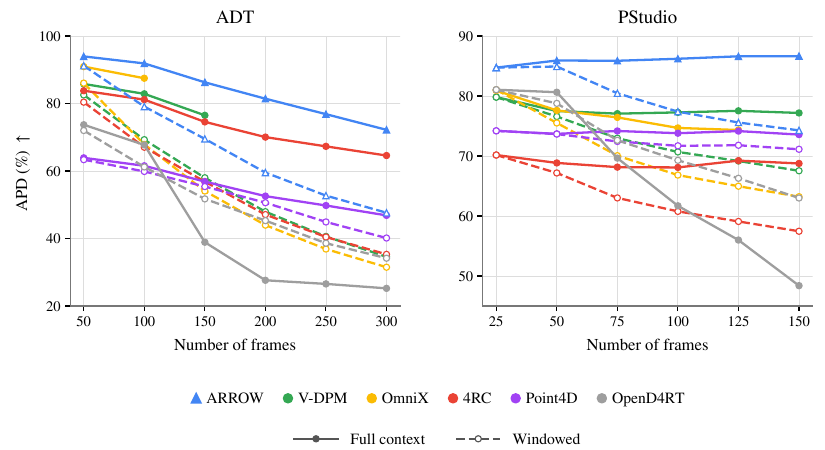}
    \caption{
        \textbf{Long tracking performance.}
        All-point APD over the indicated sequence prefix with full-context inference (solid) or $48$-frame windows with $8$-frame overlap (dashed).
        Full-context inference performs better than windowed inference for all methods except OpenD4RT.
        \method{} outperforms other methods under matched scene size and inference method.
        Evaluations run on NVIDIA GH200 GPUs.
        V-DPM evaluation on 200 frames did not finish within 80 GPU-hours.
        Other missing points ran out of memory.
    }
    \label{fig:longtracking}
\end{figure}

\Cref{fig:longtracking} compares full-context inference with $48$-frame
windows and $8$-frame overlap on Aria Digital Twin and Panoptic Studio. Each point evaluates an
entire sequence prefix using WorldTrack's APD thresholds and one global
median-distance scale. Windowed inference transfers predicted 2D query
locations between windows, except Point4D, which propagates 3D query positions. For \method{}, full context yields higher APD at
every evaluated prefix longer than one window, with a widening gap as the
sequence grows. This comparison supports retaining full context, although it
does not isolate the contributions of handoff drift and occlusion recovery.

\section{Limitations}
\label{sec:limitations}

While \method{} shows strong tracking performance as reported in the main paper, removing explicit time encodings may limit the model's ability to interpolate between points in time or to extrapolate to future points in time.
To be more precise, \method{} only supports outputs at those points in time that correspond to the observations that it received as input.

\method{} enables efficient sparse decoding and can produce dense, high-resolution outputs while batching queries to alleviate memory bottlenecks.
However, dense decoding incurs higher computational costs. Improving the efficiency of query-based approaches for dense prediction remains an important direction for future work.

Besides, while queries share image context during decoding, they do not interact directly.
We hypothesize that this lack of interaction might lead to weaker performance in far-away/low-confidence regions.

Furthermore, in multi-view tasks, the model exhibits performance degradation when in-distribution views are combined with out-of-distribution inputs, such as egocentric views. 
Addressing such domain gaps is an important direction for future work.

Finally, while the pose and $(u,v)$ estimates are globally accurate, fine-grained precision remains sensitive to local variations.
Future work may improve the precision of such estimates, for example via feature matching or local optimization.

\section{Data Preprocessing}
\label{sup:samplingpre-processing}

\begin{figure*}[t]
    \centering
    \begingroup
    \setlength{\tabcolsep}{0pt}
    \renewcommand{\arraystretch}{0}

\newcommand{\pomain}[2][0.32\linewidth]{\makebox[0.32\linewidth][c]{\raisebox{-0.5\height}{\includegraphics[
                    width=#1,
                    totalheight=0.18\linewidth,
                    keepaspectratio
                ]{#2/view}}}}

\newcommand{\posmall}[1]{\raisebox{-\height}{\includegraphics[
                width=0.1568\linewidth
            ]{#1}}}

\newcommand{\popair}[2]{\hbox to 0.32\linewidth{\posmall{#1/image_#2}\hfil
            \posmall{#1/depth_#2}}}

    \begin{tabular}{@{}c@{\hspace{0.02\linewidth}}c@{\hspace{0.02\linewidth}}c@{}}
\small\normalfont Floating objects &
        \small\normalfont Fake background &
        \small\normalfont Annotation noise
        \\[4pt]

\hspace{0.02\linewidth}\pomain[0.36\linewidth]{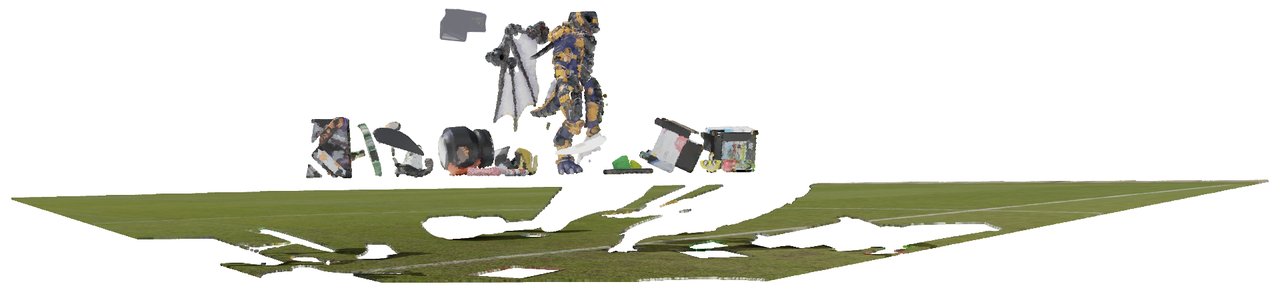}\hspace{-0.02\linewidth} &
        \pomain[0.4\linewidth]{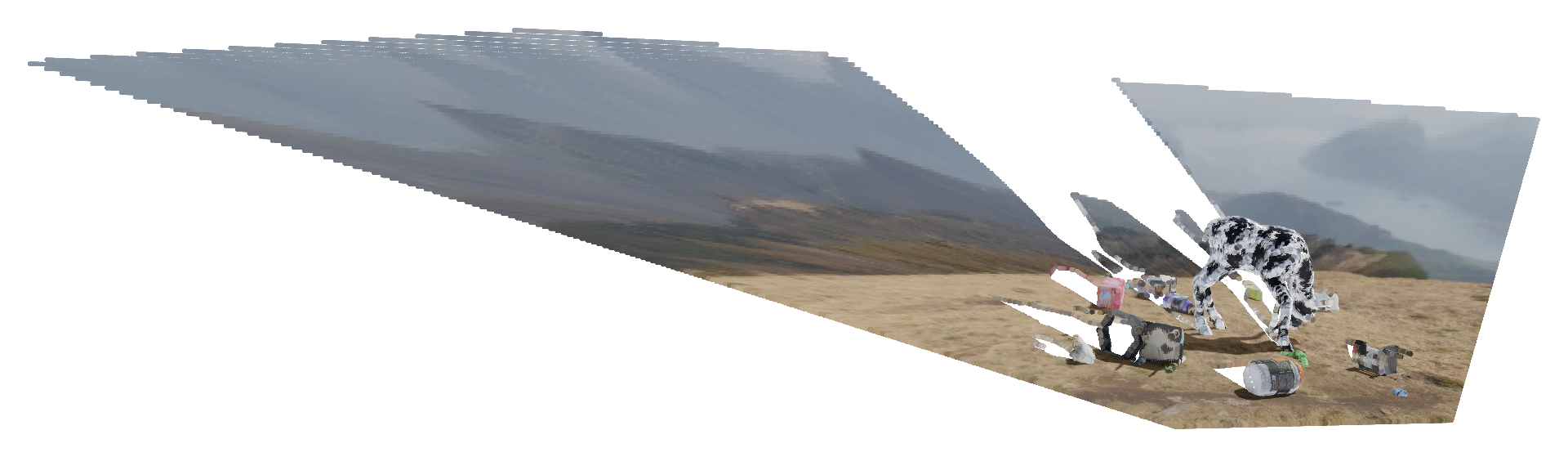} &
        \pomain{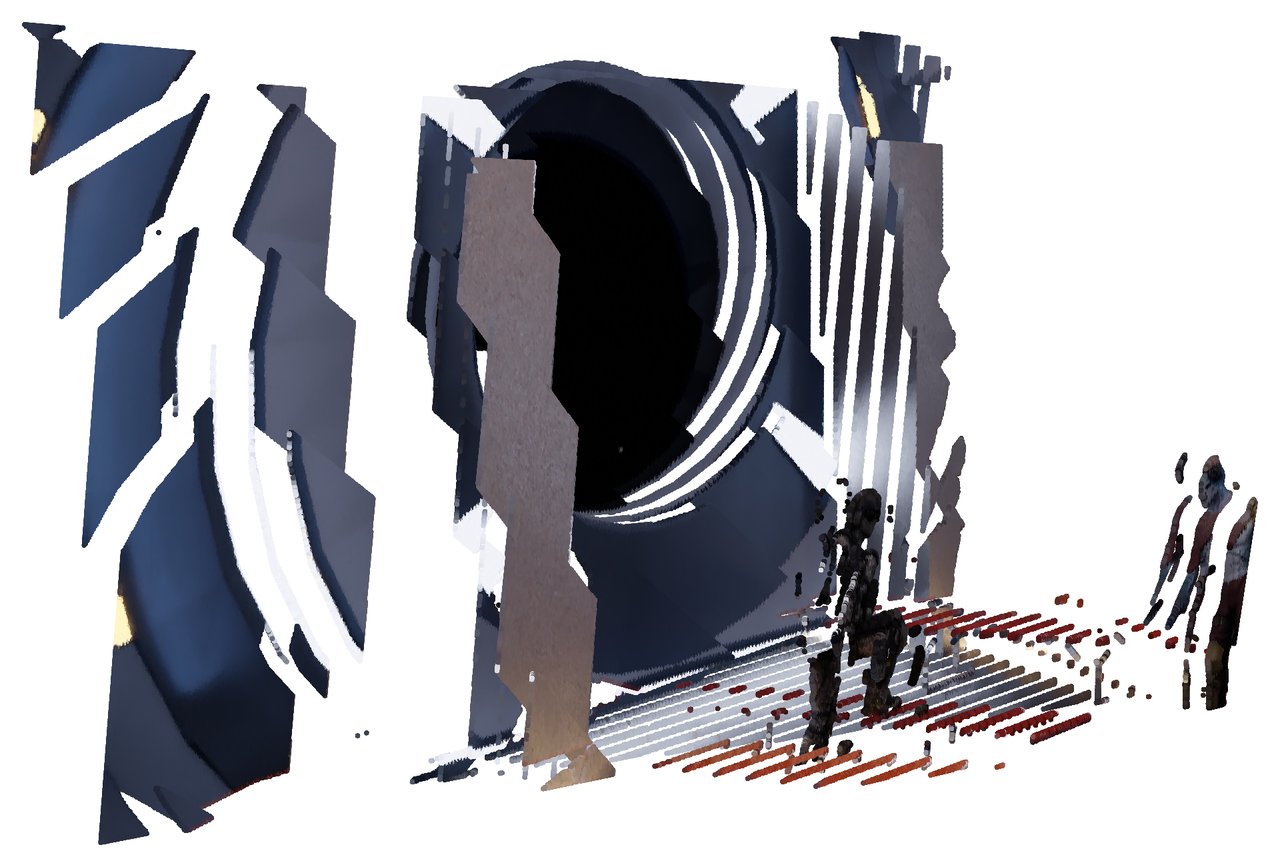}
        \\
        \noalign{\vskip 4pt}

\popair{fig/supp/pointodyssey/pointodyssey-character0_-moving}{0033} &
        \popair{fig/supp/pointodyssey/pointodyssey-ani12_new_-flat}{0932} &
        \popair{fig/supp/pointodyssey/pointodyssey-cab_h_bench_3rd-jigged}{1794}
    \end{tabular}\par

    \endgroup
    \caption{
        \textbf{Geometry and annotation issues in PointOdyssey.}
        From left to right: objects floating above the ground instead of resting on it, a background image mapped onto a flat plane rather than modeled scene geometry, and depth annotation artifacts that produce fragmented geometry.
        Each example shows a 3D visualization and the corresponding RGB image (bottom left) and depth map (bottom right).
        These corrupted cases motivate our manual scene review, scene-specific sampling rules, and exclusion of unreliable supervision.
    }
    \label{fig:pointodyssey}
\end{figure*}

\paragraph{PointOdyssey.}
Moving backgrounds, flattened background geometry and noisy annotations complicate geometric and motion supervision in PointOdyssey, as illustrated in \cref{fig:pointodyssey}.
Instance labels include static environment objects and do not directly specify motion.
We therefore review scenes in our training index using an interactive viewer, label instances as static, dynamic, or unreliable, and record background geometry and annotation issues.
Mask refinement repairs local label dropouts, treats larger depth-valid unlabeled regions separately, and combines manual decisions with track-based checks for unresolved regions.
Unreliable regions receive a dedicated ignore label.
The resulting scene metadata determines the sampling profile: standard, flat-background, or moving-background.
The flat-background profile emphasizes tracks and moving instances, while the moving-background profile restricts ordinary queries to the source point state.
We preserve per-frame intrinsics for zoom sequences and transform the released normals into our camera coordinate convention.

\paragraph{Hypersim.}
Some supplied camera orientation matrices do not have determinant $+1$.
We correct these using SVD, flipping the singular-vector direction associated with the smallest singular value when necessary to obtain a proper rotation.
This avoids treating a reflection as a camera rotation.
We also reconcile normal-map axes and orient valid normals toward the camera.

\paragraph{Track and visibility corrections.}
We decode Syn4D's compact surface correspondences into shared world-space trajectories and map Dynamic Replica's frame-local instance labels to persistent identities.
ParallelDomain-4D trajectories are derived from flow and depth and validated using forward--backward, cross-view depth, flow, and instance-consistency checks.

\paragraph{Other data preparation.}
We remove simulated vignetting and rectify ASE cameras, filter unreliable BlendedMVS geometry, and retain WildRGB-D depth only in reliable foreground regions.
For each scene, we construct a covisibility graph from pairwise overlap estimated through depth reprojection consistency, which guides image sampling during training.
Dynamic identities remain defined by the dataset's track correspondences.

\clearpage
\section{Additional Qualitative Results}
\label{sup:qualitative}
\suppressfloats[t]

\begin{figure}[htbp]
    \centering
    \begingroup
    \definecolor{depthfocus}{HTML}{28647A}
    \definecolor{depthtrees}{HTML}{A63D40}
    \definecolor{depthbench}{HTML}{315EAA}
    \newlength{\depthunit}
    \setlength{\depthunit}{\dimexpr\linewidth/1000\relax}
    \newcommand{\depthimage}[5]{\node[anchor=north west, inner sep=0pt, outer sep=0pt] at (#1,#2)
            {\includegraphics[width=#3\depthunit,height=#4\depthunit]{fig/depth/#5}};
    }
\newcommand{\depthcrop}[8]{\pgfmathsetmacro{\cropheight}{77*#7/#6}
        \pgfmathsetlengthmacro{\cropwidth}{137/#6*\depthunit}
        \pgfmathsetlengthmacro{\cropimageheight}{77/#6*\depthunit}
        \begin{scope}
            \clip (#1,#2) rectangle ++(137,\cropheight);
            \node[anchor=north west, inner sep=0pt, outer sep=0pt]
                at ({#1-137*#4/#6},{#2-77*#5/#6})
                {\includegraphics[width=\cropwidth,height=\cropimageheight]{fig/depth/#3}};
        \end{scope}
        \draw[#8, line width=0.5pt] (#1,#2) rectangle ++(137,\cropheight);
    }
    \newcommand{\depthcolumn}[3]{\node[anchor=south, font=\scriptsize] at ({#1+68.5},111) {#2};
        \depthimage{#1}{116}{137}{77}{#3}
        \foreach \left/\top/\wide/\high/\col in {
            .415/.17/.26/.36/depthtrees,.66/.49/.34/.38/depthbench}{
            \draw[white, densely dashed, line width=1.2pt]
                ({#1+137*\left},{116+77*\top}) rectangle ++({137*\wide},{77*\high});
            \draw[\col, densely dashed, line width=0.5pt]
                ({#1+137*\left},{116+77*\top}) rectangle ++({137*\wide},{77*\high});
        }
        \depthcrop{#1}{197}{#3}{.415}{.17}{.26}{.36}{depthtrees}
        \depthcrop{#1}{307.6}{#3}{.66}{.49}{.34}{.38}{depthbench}
    }
    \begin{tikzpicture}[x=\depthunit,y=-\depthunit,
        every node/.style={font=\scriptsize,inner sep=0pt}]
        \node[anchor=west, font=\small\bfseries] at (0,9) {Video frames};
        \foreach \idx/\file in {0/00,1/01,2/02,3/03,4/04,5/05,6/06_selected,7/07,8/08}{
            \depthimage{\idx*1000/9}{29}{111.111}{62}{rgb/frame_\file}
        }
        \draw[depthfocus, line width=1pt] (666.667,29) rectangle (777.778,91);
        \draw[depthfocus, line width=0.6pt] (722.222,19) -- (722.222,28);
        \node[text=depthfocus, font=\scriptsize\bfseries]
            at (722.222,9) {Selected frame};

        \depthcolumn{0}{RGB}{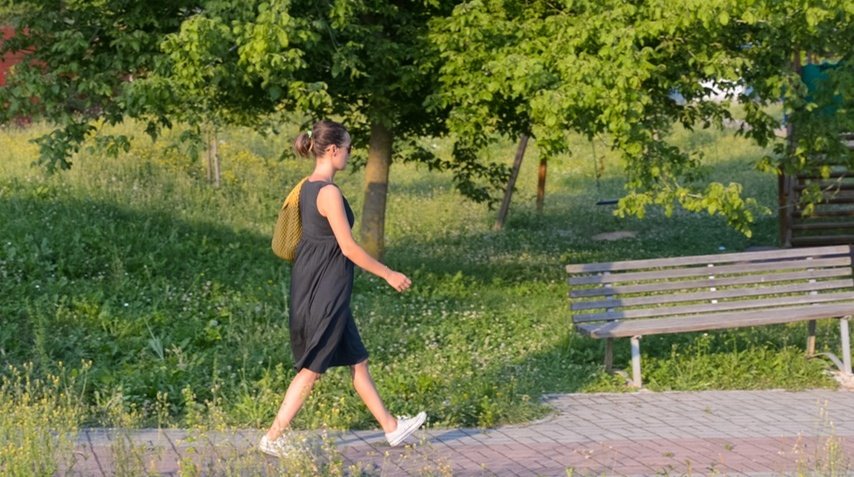}
        \depthcolumn{143.833}{$\pi^3$}{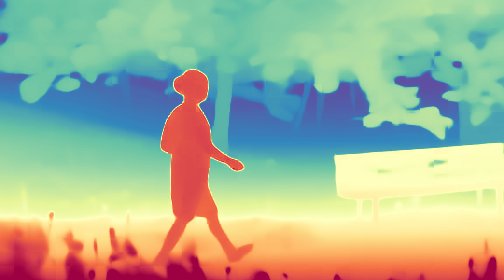}
        \depthcolumn{287.667}{VGGT-$\Omega$}{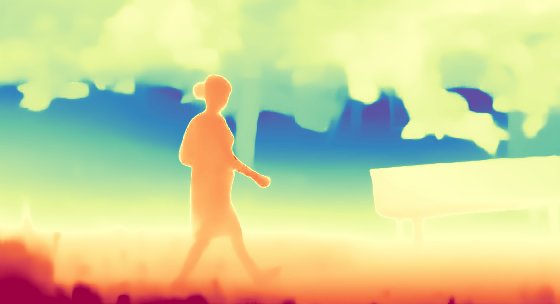}
        \depthcolumn{431.5}{V-DPM}{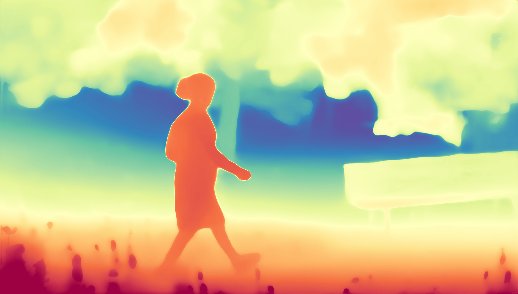}
        \depthcolumn{575.333}{4RC}{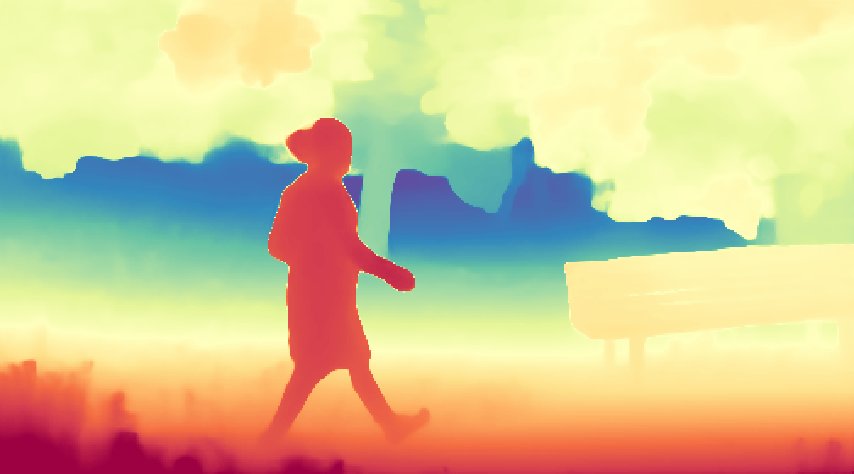}
        \depthcolumn{719.167}{OpenD4RT}{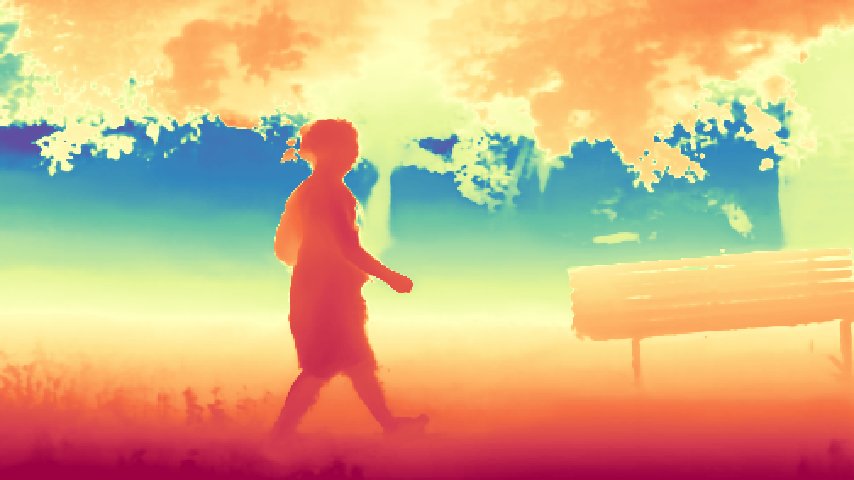}
        \depthcolumn{863}{\textbf{\method{} (ours)}}{predictions/arrow}

        \node[anchor=south west, text=depthtrees, fill=white, fill opacity=0.85,
            text opacity=1, inner sep=1pt, font=\fontsize{6}{7}\selectfont\bfseries]
            at (3,300.6) {A\enspace Tree trunk};
        \node[anchor=south west, text=depthbench, fill=white, fill opacity=0.85,
            text opacity=1, inner sep=1pt, font=\fontsize{6}{7}\selectfont\bfseries]
            at (3,390.6) {B\enspace Bench};
\draw[depthtrees, densely dashed, line width=0.45pt]
            (56.855,156.81) -- (0,197);
        \draw[depthtrees, densely dashed, line width=0.45pt]
            (92.475,156.81) -- (137,197);
        \draw[depthbench, densely dashed, line width=0.45pt]
            (90.42,182.99) -- (140.4,191) -- (140.4,305.6) -- (0,307.6);
        \draw[depthbench, densely dashed, line width=0.45pt]
            (137,182.99) -- (140.4,188) -- (140.4,305.6) -- (137,307.6);
        \pgfresetboundingbox
        \path[use as bounding box] (0,-4) rectangle (1000,395);
    \end{tikzpicture}
    \endgroup
    \caption{\textbf{Video depth comparison.} Top: RGB frames in temporal order,
        with the selected frame outlined. Middle: the selected RGB frame and
        depth predictions. Bottom: matching crops of the tree trunk (A)
        and the bench (B), marked by dashed boxes in the full views.
        \method{} preserves fine structural details in both highlighted regions.}
    \label{fig:video_depth}
\end{figure}

\begin{figure}[htbp]
    \centering
    \includegraphics[width=\linewidth]{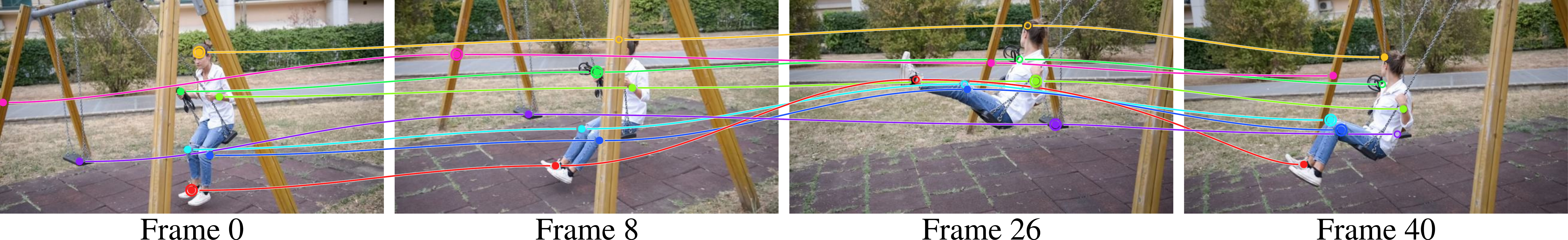}
    \par\vspace{4pt}
    \includegraphics[width=\linewidth]{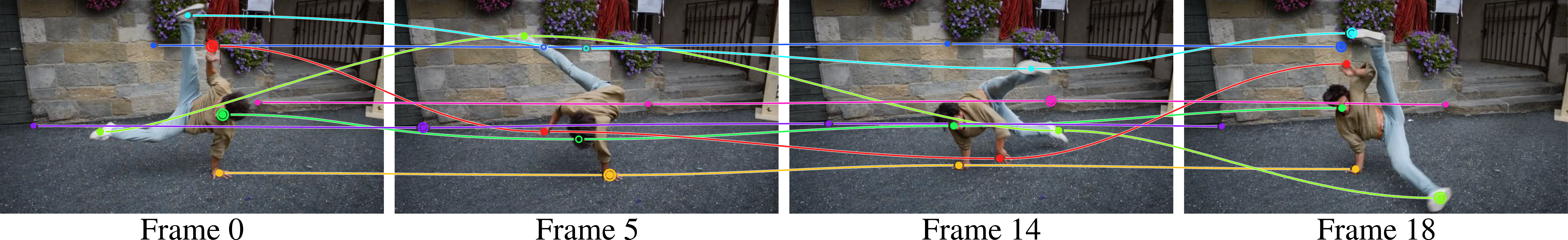}
    \caption{\textbf{Qualitative UV tracking.}
        \method{} predicts image coordinates $(\hat u,\hat v)$ and visibility
        for the swing (top) and breakdance (bottom) sequences.
        A circled dot ($\odot$) marks each source query in its source frame;
        all other markers are predictions. Filled dots ($\bullet$) indicate
        visible points and hollow circles ($\circ$) indicate occluded points.
        Colors identify corresponding points, and curves connect
        their locations across the displayed frames. These examples illustrate
        occlusion handling, tracking through challenging articulated motion,
        and consistent tracking of static scene points under camera motion.}
    \label{fig:uv_tracking}
\end{figure}

\begin{figure*}[p]
    \centering
    \begingroup
    \comparisonstyle

    \newlength{\reconrgbw}
    \setlength{\reconrgbw}{0.07\linewidth}
    \newlength{\reconrgbh}
    \setlength{\reconrgbh}{0.04\linewidth}
    \newlength{\reconh}
    \setlength{\reconh}{0.18\linewidth}
    \newlength{\reconw}
    \setlength{\reconw}{\dimexpr(\linewidth-\reconrgbw-4pt-\comparisonsep)/3\relax
    }

    \newcommand{\reconrgb}[2][]{\includegraphics[width=\reconrgbw,#1]{fig/supp/qualitative/#2}}

    \newcommand{\reconpanel}[3][]{\raisebox{\dimexpr(\depth-\height)/2\relax}{\makebox[\reconw]{\includegraphics[width=\reconw,#1]{fig/supp/qualitative/#2/#3}}}}

    \newcommand{\mvreconrow}[5]{\raisebox{\dimexpr(\depth-\height)/2\relax}{\vbox to \reconh{\vfil\offinterlineskip
                \hbox{\reconrgb[height=\reconrgbh]{#1/arrow-2_image_#3}}\kern1pt
                \hbox{\reconrgb[height=\reconrgbh]{#1/arrow-2_image_#4}}\kern1pt
                \hbox{\reconrgb[height=\reconrgbh]{#1/arrow-2_image_#5}}\kern2pt
                \hbox to \reconrgbw{\hfil\ldots\hfil}\vfil}}&
        \reconpanel{#1}{vggt-omega-new-1_view_t#2} &
        \reconpanel{#1}{da3-3_view_t#2} &
        \reconpanel{#1}{arrow-2_view_t#2}}

    \begin{tabular}{@{}Ic@{\hspace{2pt}}c@{\hspace{2pt}}c@{}}
        \comparisoninputheading &
        \small VGGT-\ensuremath{\Omega} &
        \small Depth Anything 3 &
        \comparisonours \\[3pt]

        \mvreconrow{mv_recon_nrgbd_kitchen}{0000}{0000}{0300}{0600}
        \comparisonrowsep
        \mvreconrow{mv_recon_ai-034-002}{0000}{0000}{0012}{0024}
        \comparisonrowsep
        \raisebox{-0.5\height}{\reconrgb{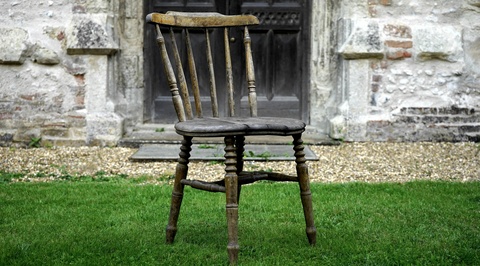}} &
        \reconpanel{mono_chair}{vggt-omega-new-5_view_t0000} &
        \reconpanel{mono_chair}{da3-4_view_t0000} &
        \reconpanel{mono_chair}{arrow-3_view_t0000}\comparisonrowsep
        \raisebox{-0.5\height}{\reconrgb{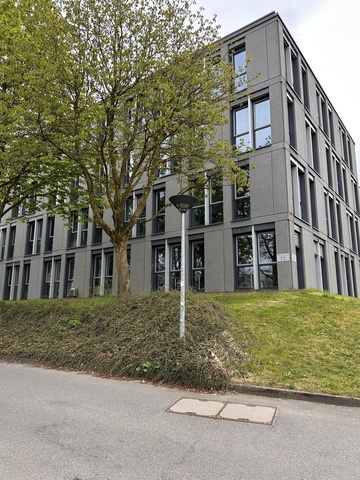}} &
        \reconpanel{mono_umic}{vggt-omega-new-3_view_t0005} &
        \reconpanel{mono_umic}{da3-2_view_t0005} &
        \reconpanel{mono_umic}{arrow-1_view_t0005}\end{tabular}\par
    \endgroup

    \caption{
        \textbf{3D reconstruction.}
        Multi-view (top two rows) and monocular (bottom two rows) 3D reconstruction results compared with state-of-the-art methods.
    }
    \label{fig:qual_reconstruction}
\end{figure*}

\begin{figure*}[p]
\centering
\begingroup
\comparisonstyle

\newlength{\mvgridrgbw}
\setlength{\mvgridrgbw}{0.07\linewidth}
\newlength{\mvgridinputw}
\setlength{\mvgridinputw}{\dimexpr3\mvgridrgbw+4pt\relax}

\newcommand{\mvgridrgb}[3]{\includegraphics[width=\mvgridrgbw]{fig/supp/qualitative/#1_image_v#2_#3}}

\newcommand{\mvgridvideo}[4]{\vbox{\offinterlineskip
    \hbox to \mvgridrgbw{\hfil\scriptsize View~#3\hfil}\kern3pt
    \foreach \frame in {#4} {\hbox{\mvgridrgb{#1}{#2}{\frame}}\kern1pt
    }\kern1pt
    \hbox to \mvgridrgbw{\hfil\ldots\hfil}}}

\newlength{\mvgridpanelh}
\setlength{\mvgridpanelh}{0.25\linewidth}
\newlength{\mvgridpanelw}
\setlength{\mvgridpanelw}{\dimexpr(\linewidth-\mvgridinputw-4pt-\comparisonsep)/3\relax
}

\newcommand{\mvgridpanel}[2][1.15]{\parbox[c][\mvgridpanelh][c]{\mvgridpanelw}{\centering
    \makebox[\mvgridpanelw][c]{\includegraphics[
        width=#1\mvgridpanelw,
        totalheight=#1\mvgridpanelh,
        keepaspectratio
      ]{fig/supp/qualitative/#2}}}}

\begin{tabular}{@{}Ic@{\hspace{2pt}}c@{\hspace{2pt}}c@{}}
  \comparisoninputheading &
  \small OmniX &
  \small\shortstack{MVTracker \; + \emph{DA3}} &
  \comparisonours \\[3pt]

  \parbox[c][\mvgridpanelh][c]{\mvgridinputw}{\hbox{\mvgridvideo{mv_track_pstudio_tennis/arrow-2}{00}{1}{0000,0010,0020}\kern2pt
      \mvgridvideo{mv_track_pstudio_tennis/arrow-2}{03}{2}{0000,0010,0020}\kern2pt
      \mvgridvideo{mv_track_pstudio_tennis/arrow-2}{08}{3}{0000,0010,0020}}} &
  \mvgridpanel{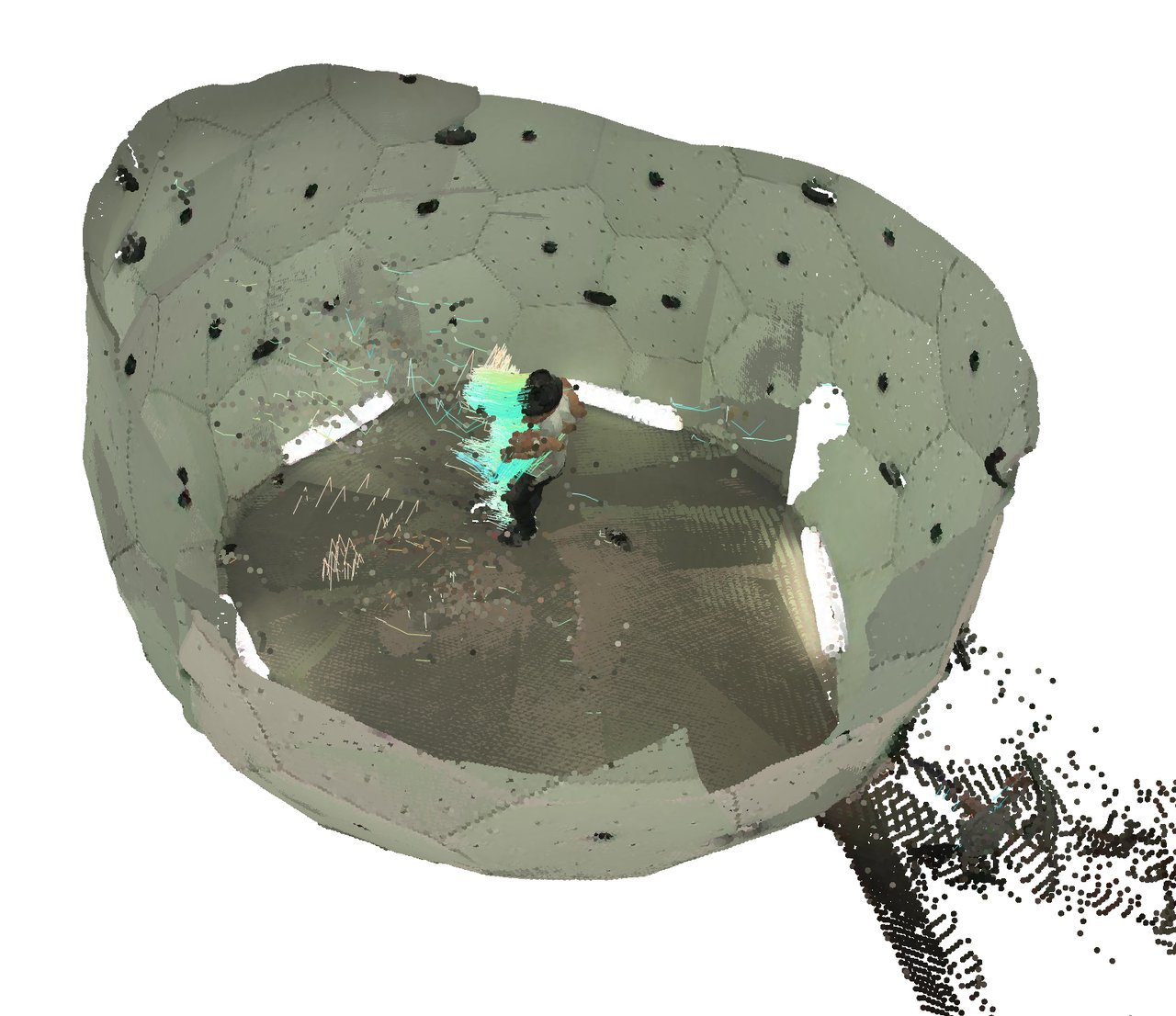} &
  \mvgridpanel{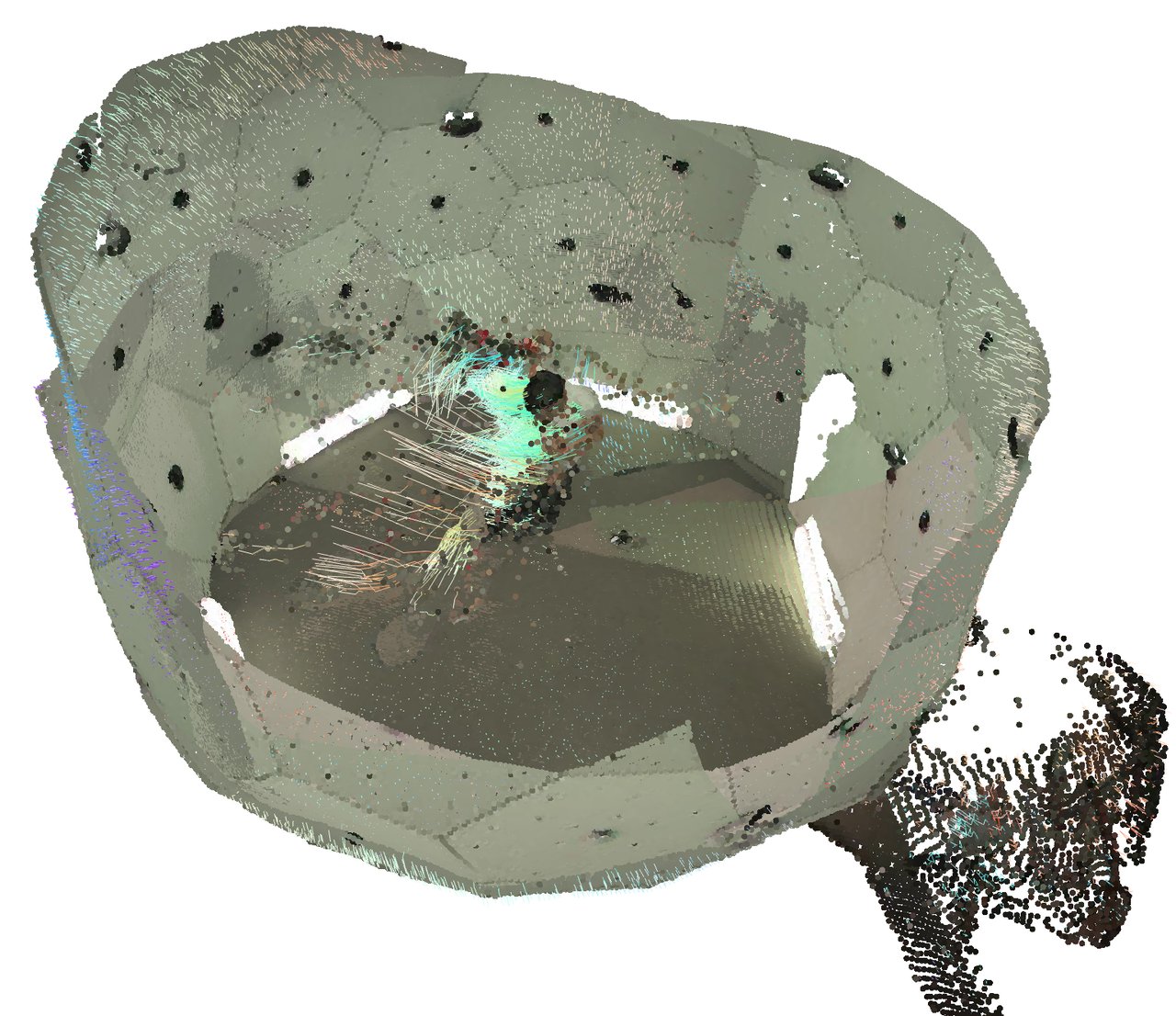} &
  \mvgridpanel{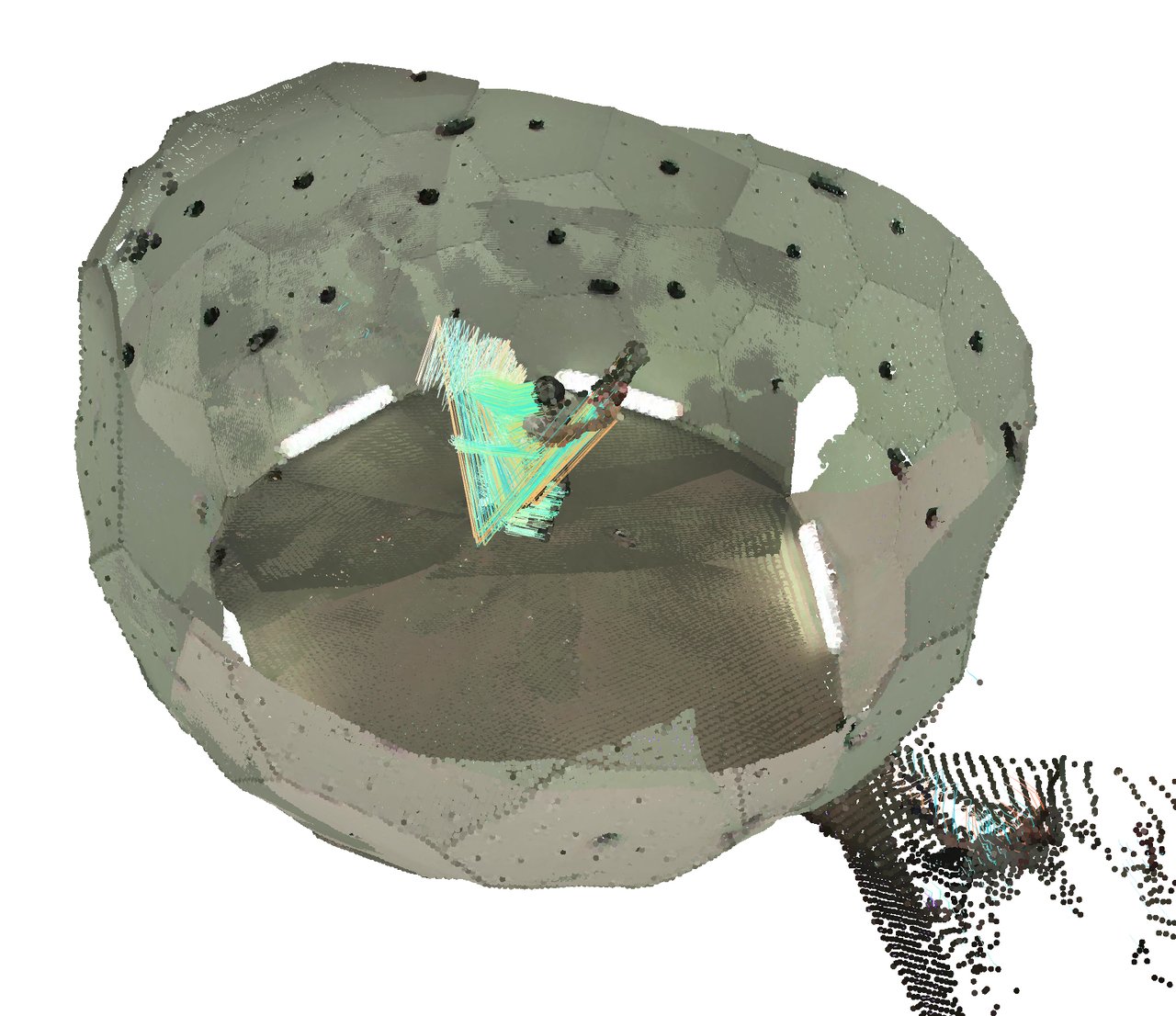}\comparisonrowsep
  \parbox[c][\mvgridpanelh][c]{\mvgridinputw}{\hbox{\mvgridvideo{mv_track_kubric_20/arrow-3}{00}{1}{0000,0006,0012}\kern2pt
      \mvgridvideo{mv_track_kubric_20/arrow-3}{01}{2}{0000,0006,0012}\kern2pt
      \mvgridvideo{mv_track_kubric_20/arrow-3}{02}{3}{0000,0006,0012}}} &
  \mvgridpanel[1]{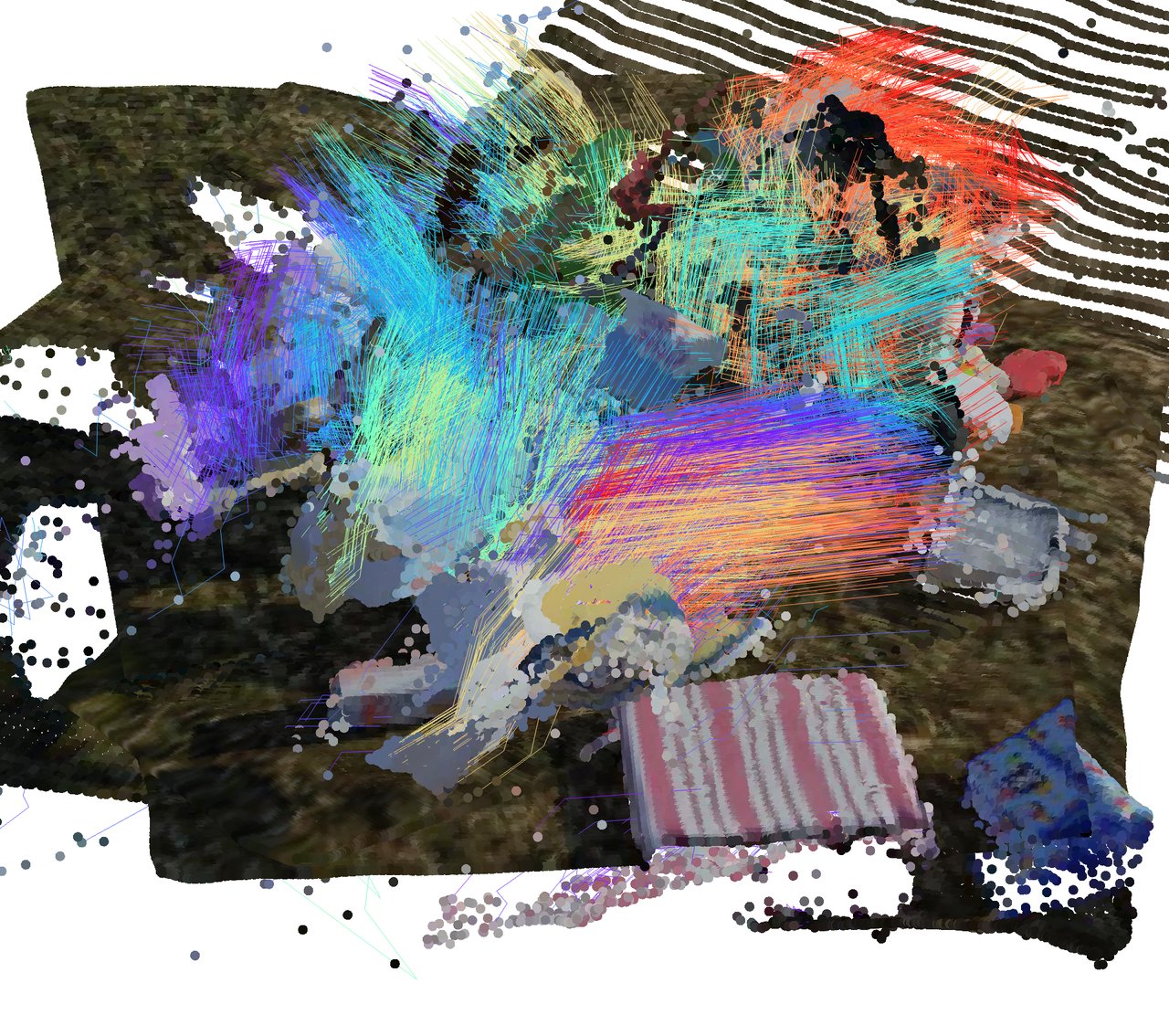} &
  \mvgridpanel[1]{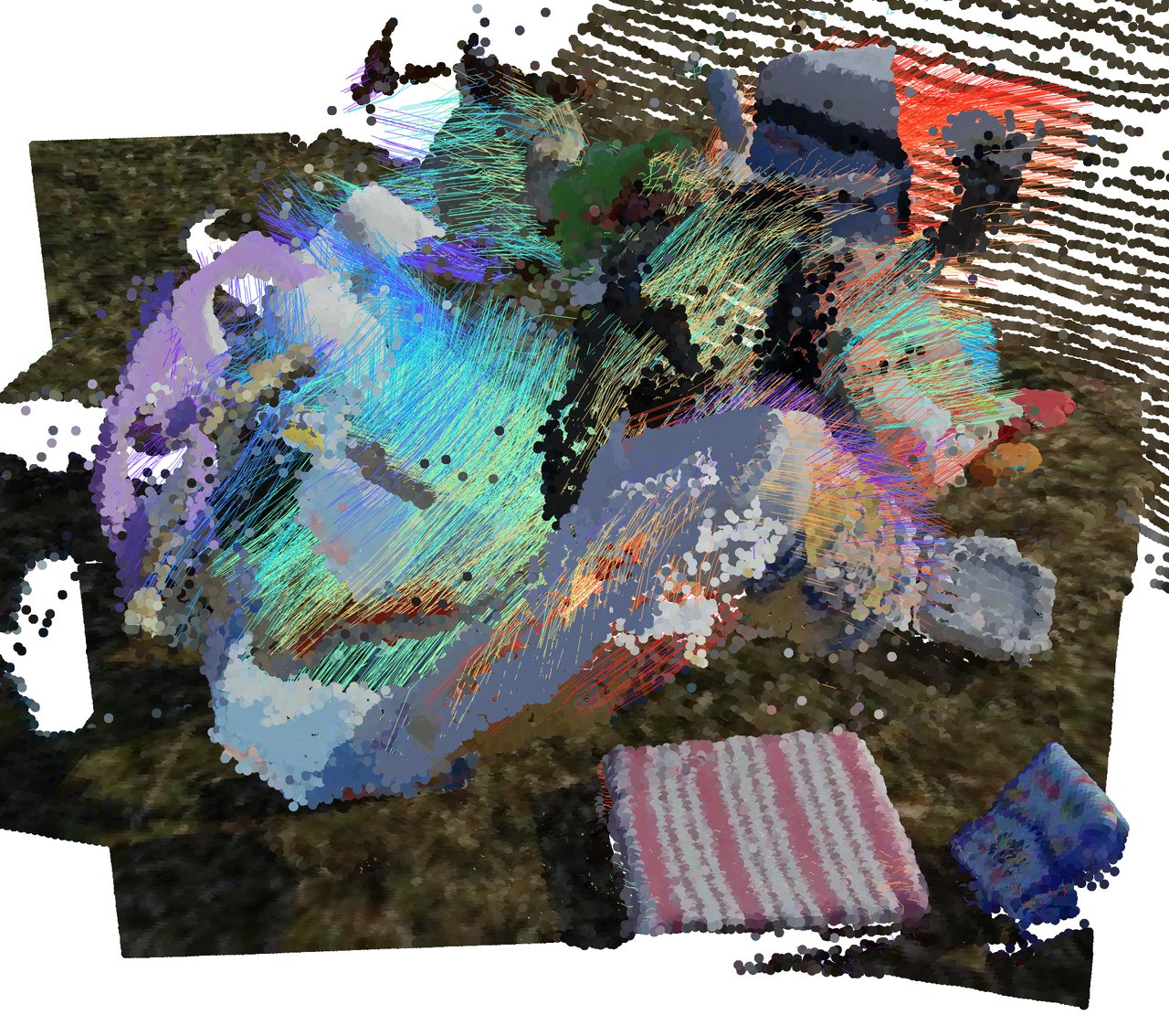} &
  \mvgridpanel[1]{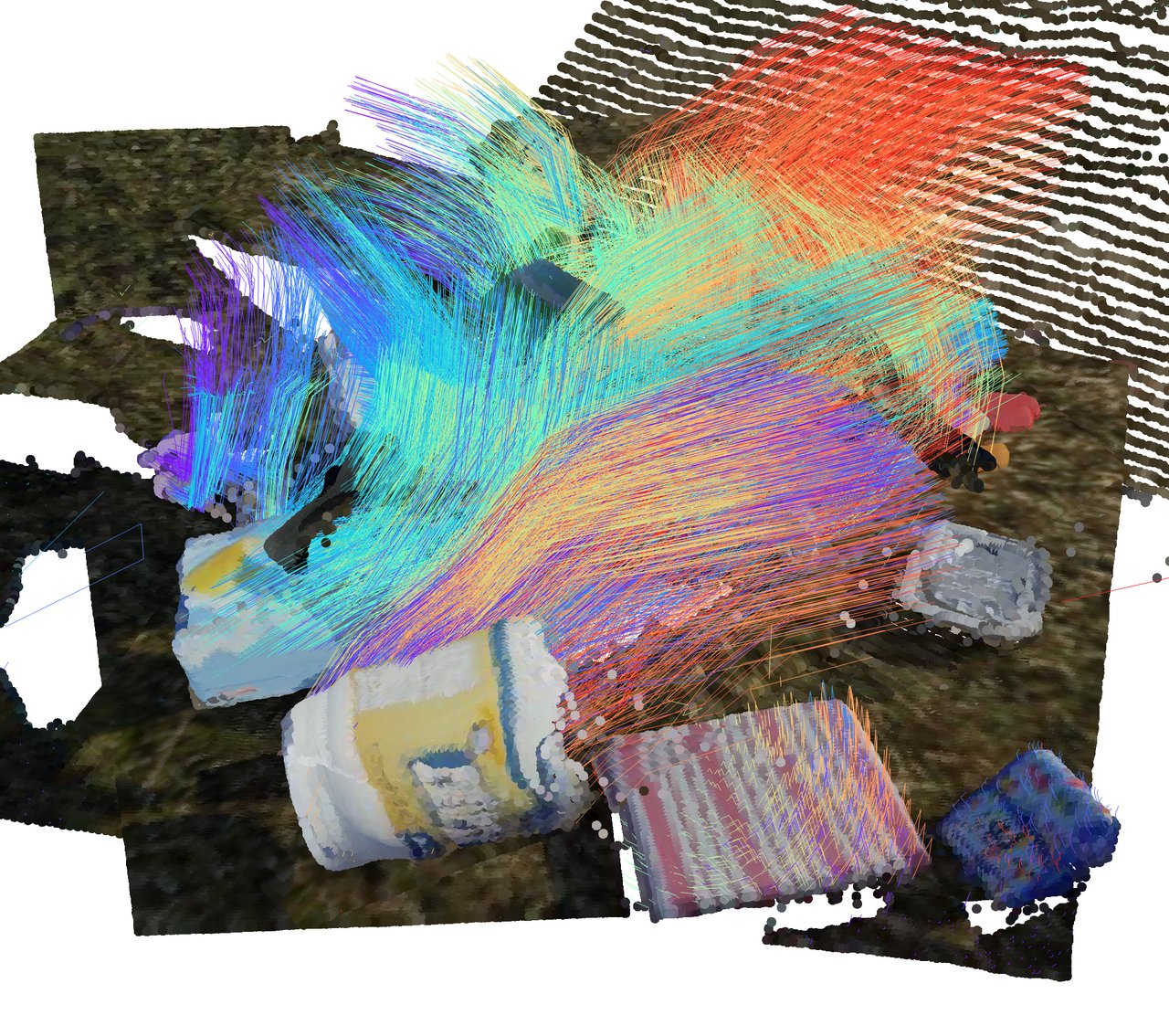}\end{tabular}\par
\endgroup

\caption{\textbf{Multi-view 3D tracking.}
    Examples from Panoptic Studio (top) and Multi-View Kubric (bottom).
    The three input columns on the left show successive frames from three camera views.
}
\label{fig:qual_multiview_tracking}
\end{figure*}

\begin{figure}[p]
    \centering
    \begingroup
    \newlength{\gallerywidth}\setlength{\gallerywidth}{0.485\linewidth}
    \newlength{\galleryinputwidth}\setlength{\galleryinputwidth}{0.07\linewidth}
    \newlength{\galleryinputheight}\setlength{\galleryinputheight}{0.052\linewidth}
    \newlength{\gallerygap}\setlength{\gallerygap}{8pt}
    \newlength{\gallerypanelwidth}
    \setlength{\gallerypanelwidth}{\dimexpr\gallerywidth-\galleryinputwidth-\gallerygap\relax}
    \newcommand{\galleryscene}[3]{\hbox to \gallerywidth{\raisebox{\dimexpr(\depth-\height)/2\relax}{\vbox{\offinterlineskip
                    \foreach \frame in {#2} {\hbox to \galleryinputwidth{\hfil
                            \includegraphics[width=\galleryinputwidth,height=\galleryinputheight,keepaspectratio]{fig/#1/rgb\frame}\hfil}\kern1pt
                    }#3}}\hspace{\gallerygap}\raisebox{\dimexpr(\depth-\height)/2\relax}{\includegraphics[width=\gallerypanelwidth]{fig/#1/arrow}}}}
    \newcommand{\galleryellipsis}{\kern1pt\hbox to \galleryinputwidth{\hfil\ldots\hfil}}
    \begingroup
    \setlength{\gallerywidth}{\linewidth}
    \setlength{\galleryinputheight}{0.10\linewidth}
    \setlength{\gallerypanelwidth}{\dimexpr\gallerywidth-\galleryinputwidth-\gallerygap\relax}
    \galleryscene{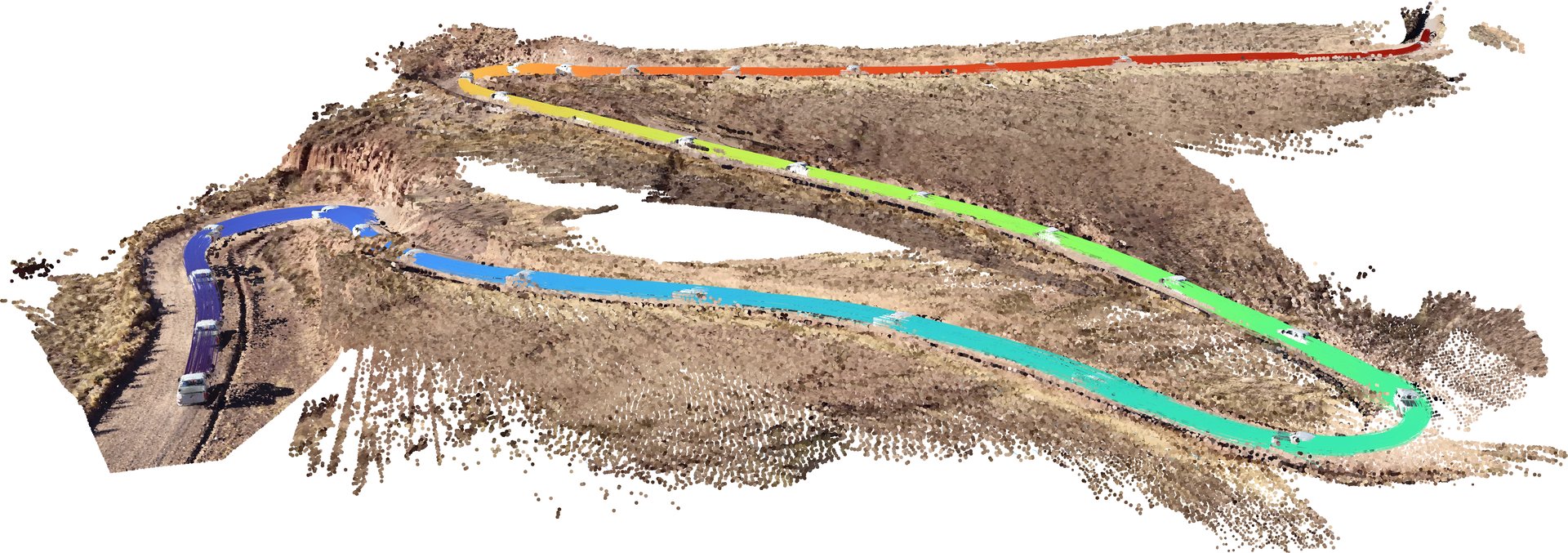}{1,4,7}{\galleryellipsis}
    \par\endgroup
    \vspace{12pt}
    \hbox to \linewidth{\galleryscene{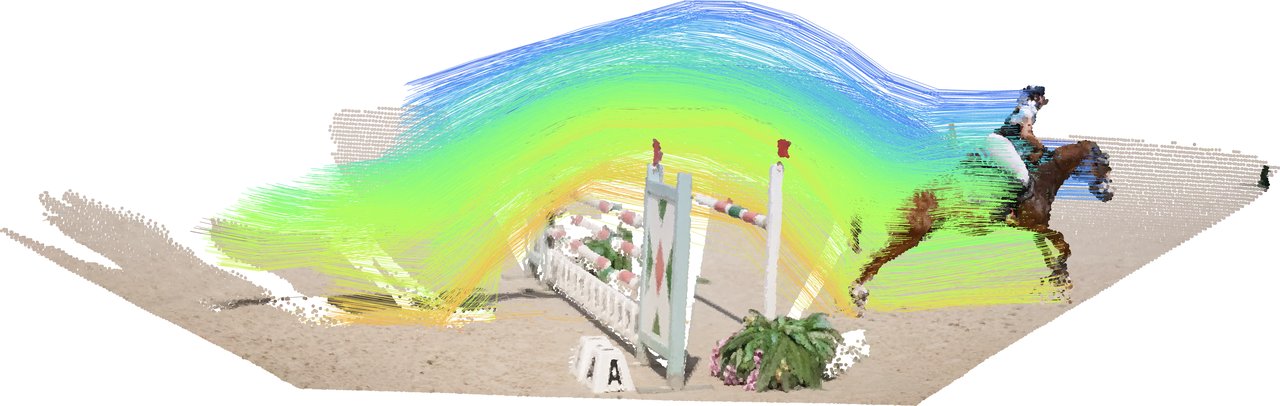}{1,2,4}{\galleryellipsis}\hfil
        \galleryscene{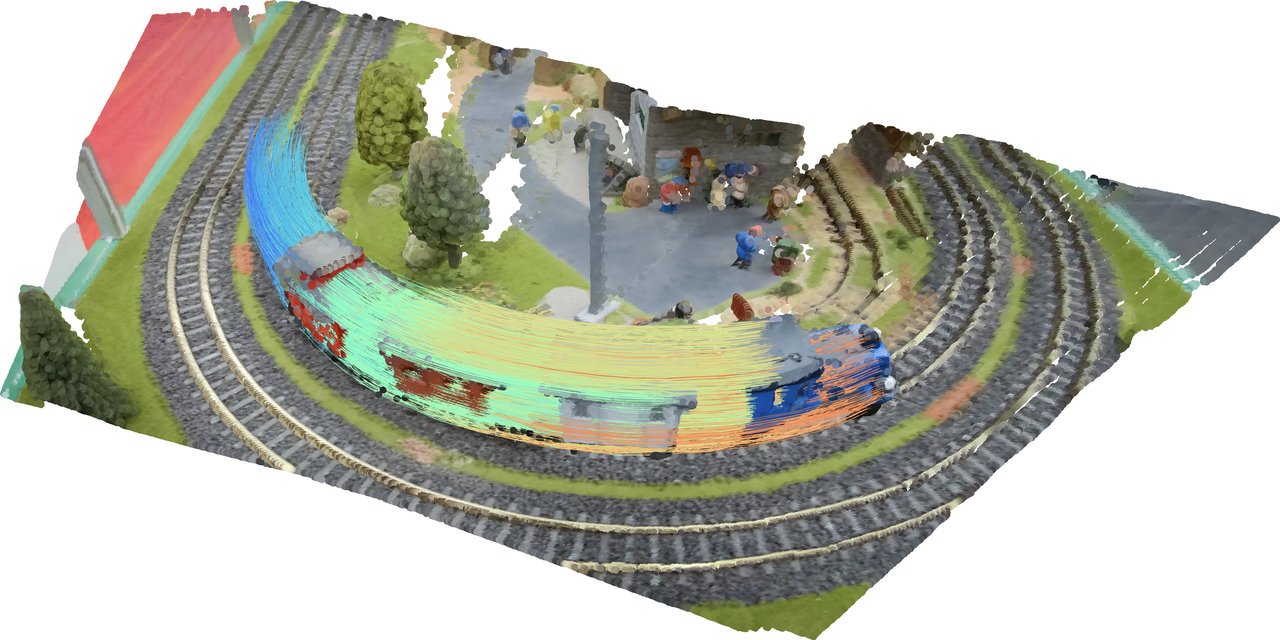}{1,2,4}{\galleryellipsis}}
    \par\vspace{12pt}
    \hbox to \linewidth{\galleryscene{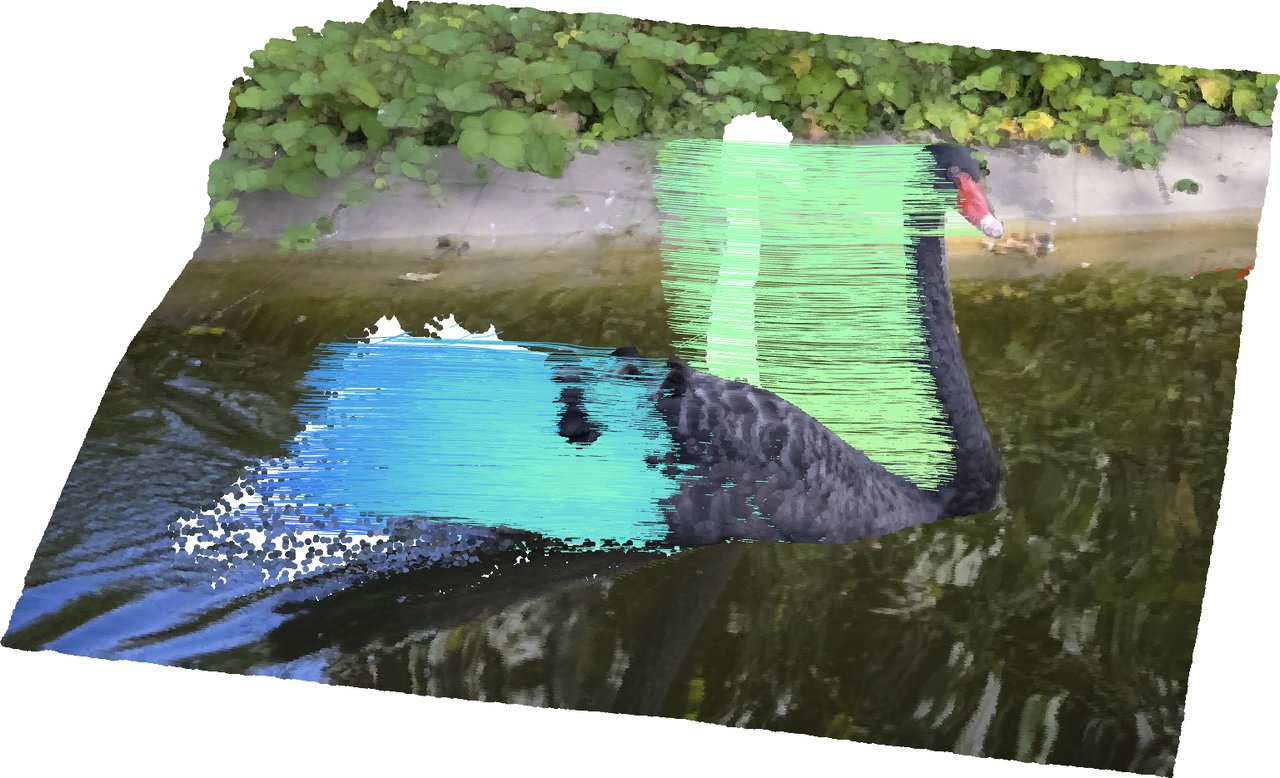}{1,2,3}{\galleryellipsis}\hfil
        \galleryscene{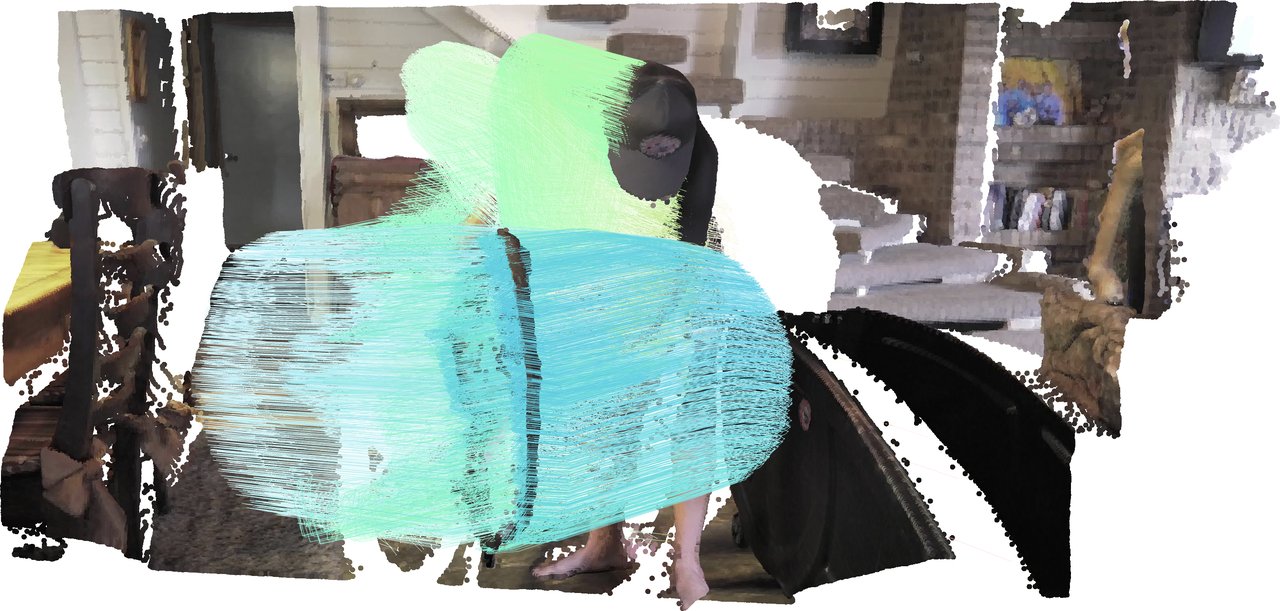}{1,2,4}{\galleryellipsis}}
    \par\vspace{12pt}
    \hbox to \linewidth{\galleryscene{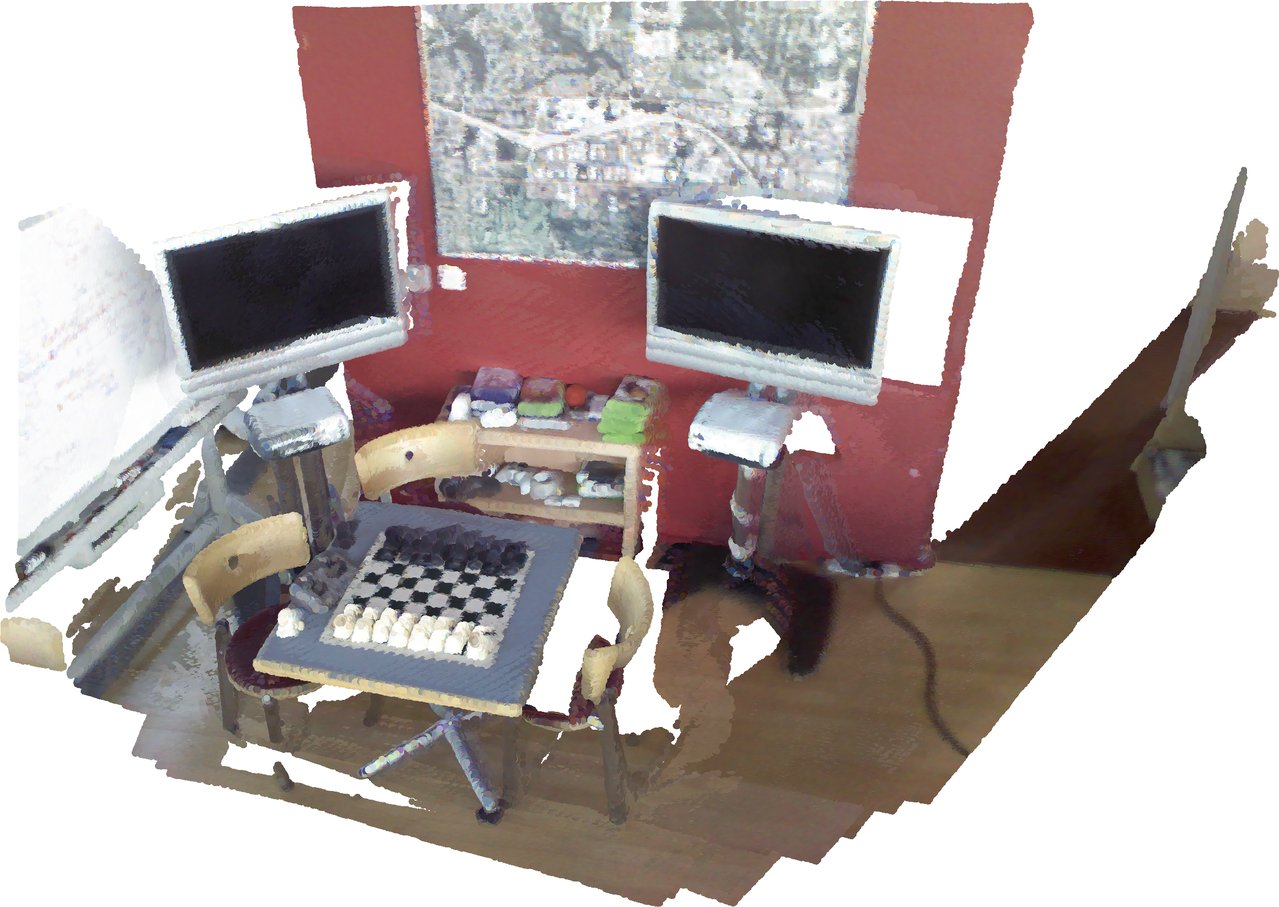}{1,2,3}{\galleryellipsis}\hfil
        \galleryscene{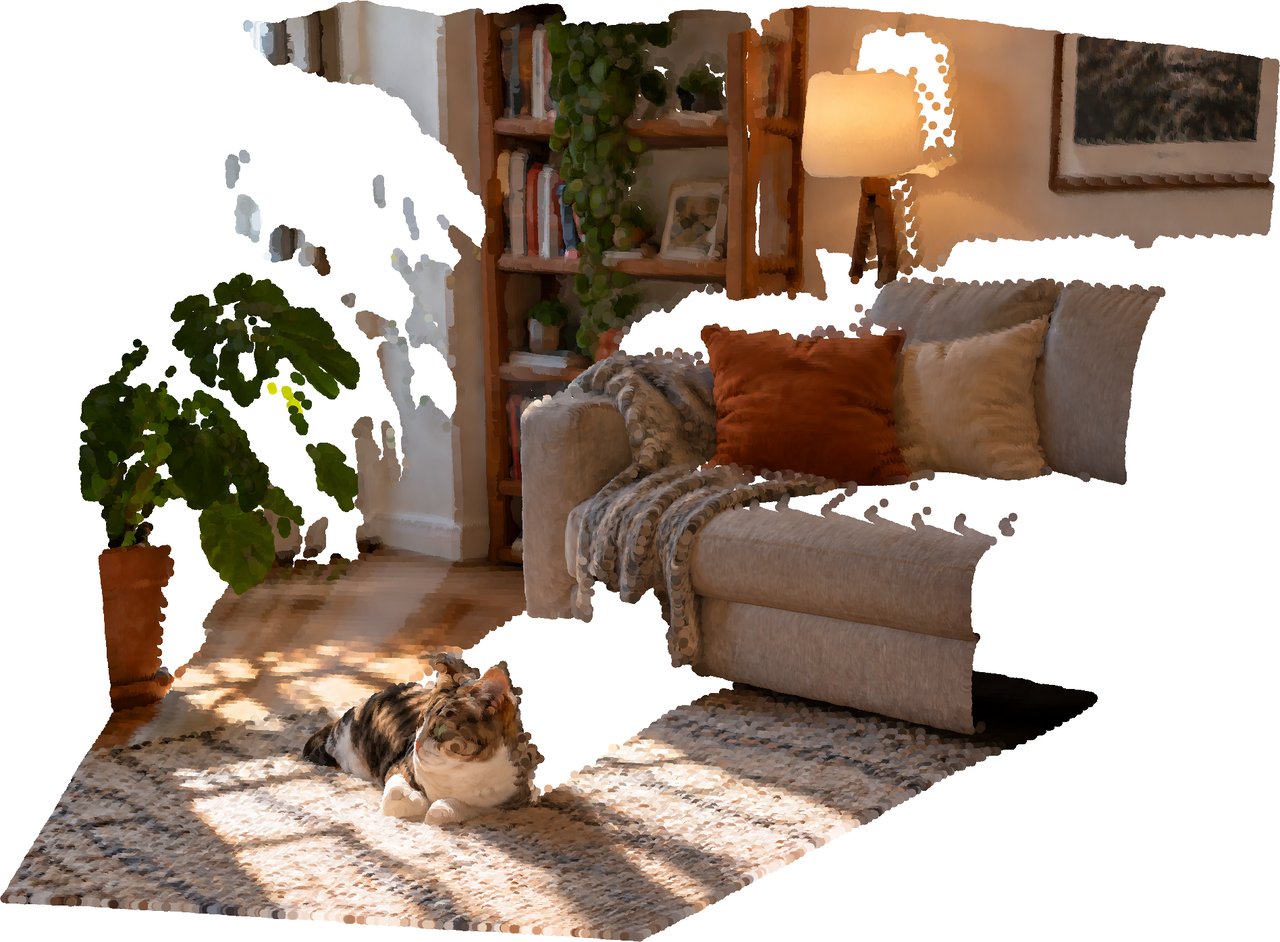}{1}{}}
    \par\vspace{12pt}
    \hbox to \linewidth{\galleryscene{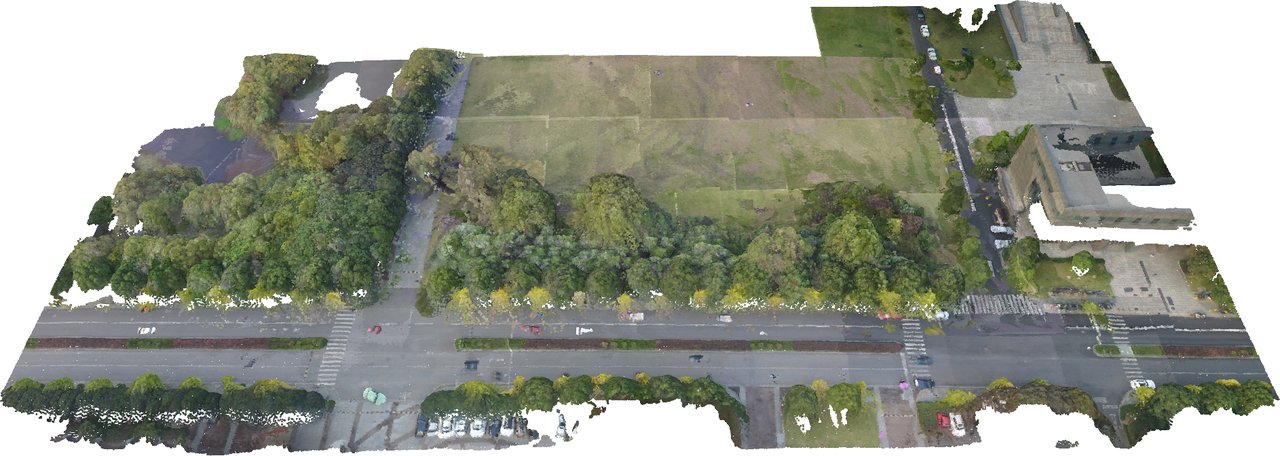}{1,2,3}{\galleryellipsis}\hfil
        \galleryscene{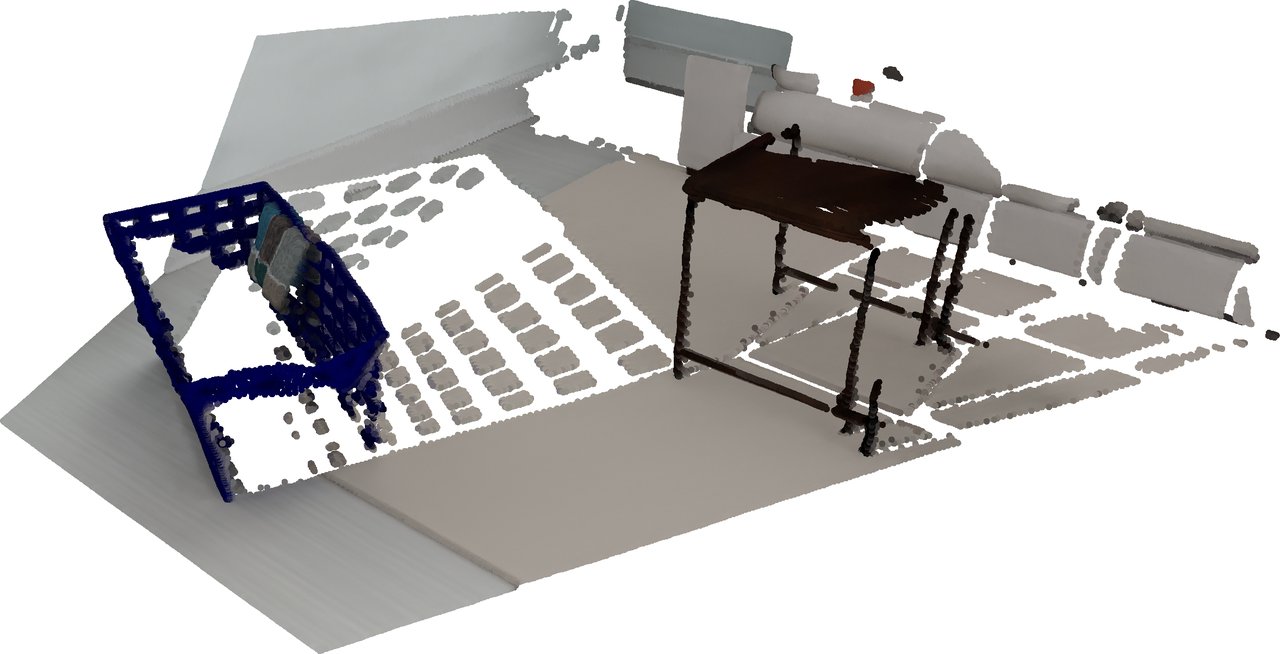}{1}{}}
    \par
    \endgroup
    \caption{\textbf{Additional qualitative results.}
        \method{} works well across dynamic scenes (top three rows), static scenes (bottom two left) and monocular inputs (bottom two right).}
    \label{fig:additional_qualitative}
\end{figure}

\Cref{fig:video_depth} compares depth predictions for a selected video frame.
\Cref{fig:uv_tracking} shows UV tracks and visibility for queries from different
source frames, with correct occlusion predictions, consistent tracks through
articulated motion, and static background points remaining attached to the same
scene locations despite camera motion.
\Cref{fig:qual_reconstruction,fig:qual_multiview_tracking} compare reconstruction
and multi-view 3D tracking, respectively, while
\cref{fig:additional_qualitative} shows further qualitative examples.

\clearpage

\end{document}